\documentclass[10pt,twocolumn,a4paper]{article}   

\usepackage[datasets,pagenumbers]{wacv}

\makeatletter
\patchcmd{\@maketitle}{\vskip .375in}{}{}{\PackageWarning{arxiv}{title-space patch failed}}
\apptocmd{\@maketitle}{%
  \vspace{2pt}%
  \begin{center}%
    \includegraphics[width=\textwidth]{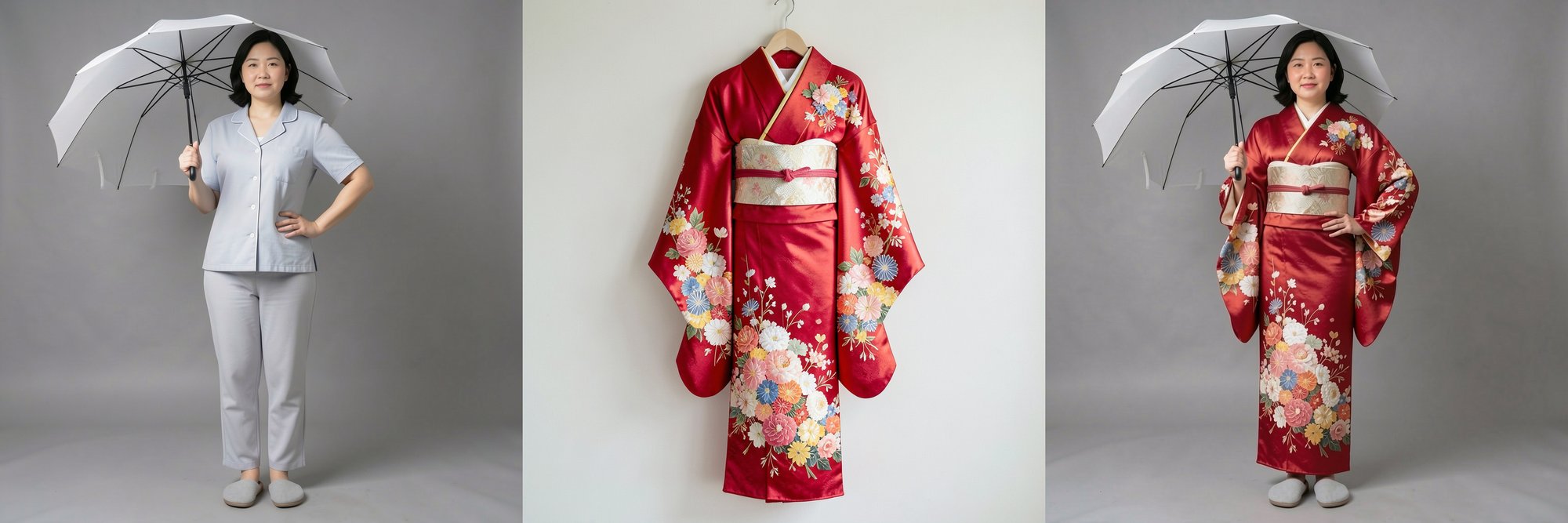}%
  \end{center}%
  \vspace{-8pt}%
  \captionof{figure}{CURVTON-205K triplets: source person, in-shop garment, and synthesized try-on, all at $1024\times1024$ resolution, generated without cloth segmentation masks.}%
  \label{fig:teaser}%
  \vspace{6pt}%
}{}{\PackageWarning{arxiv}{teaser-injection patch failed}}
\makeatother

\usepackage[T1]{fontenc}
\usepackage{microtype}

\usepackage{multirow}

\usepackage{wrapfig}

\usepackage{placeins}

\usepackage{multicol}

\usepackage{listings}

\lstdefinestyle{systemprompt}{
    basicstyle=\ttfamily\scriptsize,
    breaklines=true,
    breakatwhitespace=false,
    columns=fullflexible,
    keepspaces=true,
    frame=single,
    showstringspaces=false,
    tabsize=2,
    xleftmargin=4pt,
    xrightmargin=4pt,
    aboveskip=6pt,
    belowskip=8pt
}

\lstdefinestyle{dictionarycode}{
    language=Python,
    basicstyle=\ttfamily\scriptsize,
    breaklines=true,
    breakatwhitespace=false,
    columns=fullflexible,
    keepspaces=true,
    frame=single,
    showstringspaces=false,
    upquote=true,
    tabsize=2,
    xleftmargin=4pt,
    xrightmargin=4pt,
    aboveskip=6pt,
    belowskip=8pt,
    literate=
      {=}{{=}}1
      {ä}{{\"a}}1
      {é}{{\'e}}1
      {í}{{\'i}}1
      {á}{{\'a}}1
      {ó}{{\'o}}1
      {ö}{{\"o}}1
      {ü}{{\"u}}1
      {ñ}{{\~n}}1
      {ç}{{\c{c}}}1
      {–}{{--}}1
      {—}{{---}}1
      {─}{{-}}1
      {═}{{=}}1
}

\usepackage[hyphens]{url}
\DeclareUrlCommand{\modelid}{\urlstyle{tt}}

\newcommand{\contfig}[1]{\par\vspace{6pt}{\small Figure~\ref{#1} (continued).}}

\newcommand{\projecturl}{\url{https://curvton2.github.io/curvton/}}
\newcommand{\codeurl}{\url{https://github.com/curvton2/CURVTON_205K}}
\newcommand{\dataseturl}{\url{https://huggingface.co/datasets/curvton2/curvton2}}

\usepackage{float}

\definecolor{wacvblue}{rgb}{0.21,0.49,0.74}
\usepackage[
    breaklinks,
    colorlinks,
    allcolors=wacvblue
]{hyperref}
\hypersetup{
  pdftitle={CURVTON-205K: A Mask-Free Difficulty-Stratified Synthetic Dataset for Robust and Scalable Virtual Try-On},
  pdfauthor={Dipan Bartaula, Anuj Shrestha, Ankit Belbase, Anish Bishwakarma, Saroj Poudel, Samip Ghimire, Shashank Bhatta, Eric Arazo, Binod Bhattarai},
}

\def\wacvPaperID{242}
\def\confName{WACV}
\def\confYear{2027}

\title{CURVTON-205K: A Mask-Free Difficulty-Stratified Synthetic Dataset
for Robust and Scalable Virtual Try-On}

\author{%
Dipan Bartaula\textsuperscript{1,3,$\dagger$,*}\quad
Anuj Shrestha\textsuperscript{1,$\dagger$}\quad
Ankit Belbase\textsuperscript{1}\quad
Anish Bishwakarma\textsuperscript{1}\quad
Saroj Poudel\textsuperscript{1}\\
Samip Ghimire\textsuperscript{1}\quad
Shashank Bhatta\textsuperscript{1}\quad
Eric Arazo\textsuperscript{2}\quad
Binod Bhattarai\textsuperscript{2,4}\\
\small\textsuperscript{1}IOE Pulchowk Campus, Tribhuvan University, Nepal\quad
\textsuperscript{2}Fogsphere (Redev.AI Ltd, UK)\quad
\textsuperscript{3}NAAMII, Nepal\\
\small\textsuperscript{4}University of Aberdeen, UK\\
\small\textsuperscript{$\dagger$}Equal contribution\quad
\small\textsuperscript{*}Corresponding author: \texttt{078bct040.dipan@pcampus.edu.np}
}

\begin{document}

\maketitle


\begin{abstract}
Recent advances in 2D virtual try-on (VTON) have largely focused on architectural innovation, yet robust performance in diverse real-world conditions remains challenging. We argue that failure modes in VTON, such as inaccurate occlusion handling, unconstrained poses, and poor coverage of under-represented garments and appearance factors, are tightly coupled to dataset design choices and long-tail data regimes. 
We present a scalable data generation pipeline that coordinates specialized editing and generation models with vision-language model (VLM) guidance to synthesize triplets without requiring parsing masks or dense-pose annotations. Using this pipeline, we create CURVTON-205K, a curated synthetic dataset of 205K triplets: source person, target garment, and corresponding try-on image. CURVTON-205K is explicitly designed to broaden coverage across garments, person appearance, poses, occlusions, and visual conditions. Additionally, CURVTON-205K is organized into three difficulty tiers, enabling controlled evaluation under varying levels of difficulty and supporting difficulty-aware experimentation. We analyze the dataset diversity and demonstrate its potential to improve try-on fidelity across four VTON architectures and its cross-dataset generalization. Generated at $1024\times1024$ resolution, CURVTON-205K and its code are publicly released for reproducible research. Project page: \projecturl.

\end{abstract}

\section{Introduction}
Virtual try-on (VTON) aims to synthesize a realistic image of a person wearing a target garment while preserving both the person’s appearance and the garment’s visual characteristics. Beyond its applications in online retail, VTON presents a challenging conditional image-generation problem involving geometric alignment, fine-grained appearance preservation, and plausible synthesis under pose changes and occlusions. Despite substantial progress, modern VTON systems remain brittle under diverse and unconstrained conditions, raising the question of whether current limitations arise solely from model architectures or also from the coverage and structure of existing datasets.

Architectural advances have long driven 2D virtual try-on, evolving from multi-stage warping and adversarial synthesis (e.g., VITON~\cite{Han2018VITON}, CP-VTON~\cite{Wang2018CPVTON}) toward higher-fidelity and more robust pipelines that reduce reliance on human parsing cues (e.g., CP-VTON+~\cite{Minar2020CPVTONPLUS}, VITON-HD~\cite{Choi2021VITONHD}, PF-AFN~\cite{Ge2021PFAFN}). Recent diffusion-based methods~\cite{Kim2024StableVITON,Morelli2023LaDIVTON,Xu2024OOTDiffusion} further improve photorealism and fine-grained garment preservation, with extensions targeting efficiency and consistency (e.g.,~\cite{Choi2024IDMVTON,Chen2024MVVTON}).

Despite these architectural advances, robust virtual try-on in unconstrained settings remains elusive. Our empirical analysis of representative VTON pipelines suggests that several persistent limitations (e.g., poor handling of occlusions, limited pose extrapolation) are not solely architectural. Instead, they are strongly influenced by data coverage and dataset design. Existing datasets, such as DeepFashion~\cite{Liu2016DeepFashion}, DressCode~\cite{Morelli2022DressCode}, and specialized benchmarks like People Snapshot~\cite{Alldieck2018PeopleSnapshot}, MoVi~\cite{Ghorbani2021MoVi}, StreetTryOn~\cite{Yang2023StreetTryOn}, and MPV~\cite{Guo2023MPV}, have enabled substantial progress, but they also exhibit coverage gaps that can hinder generalization. In particular, pose and viewpoint variation, garment diversity, severe occlusions, and diverse body and appearance attributes remain difficult to cover comprehensively. Consequently, evaluation is often dominated by aggregate metrics that can obscure how models behave under increasingly challenging visual conditions. By contrast, curriculum-aware dataset structure enables evaluation across coarse but controlled tiers that jointly capture harder combinations of pose variation, occlusion, and garment-person interaction, while also making it possible to assess whether difficulty-aware training offers benefits as visual complexity increases. Finally, many VTON pipelines rely on parsing masks, dense pose, or other auxiliary annotations, which increase annotation and preprocessing costs, and can propagate upstream errors into the try-on model, making it difficult to isolate the source of failure.


In this work, we question the assumption that architectural advances alone are sufficient for robust VTON performance. Instead, we adopt a data-centric approach that targets coverage, diversity, and controllable difficulty. We propose a scalable data generation pipeline that coordinates specialized image editing and generation models with vision-language models (VLMs) to synthesize mask-free image triplets: a source person image, a target garment image, and the corresponding try-on image. The resulting triplet format supports VTON models that do not require explicit parsing masks or dense-pose conditioning, thereby reducing reliance on auxiliary preprocessing systems and enabling simpler training and inference pipelines. Using this pipeline, we build CURVTON-205K, a curated synthetic dataset explicitly designed to broaden coverage across garment types and styles, person appearance attributes, and capture conditions. Beyond scale, the goal of CURVTON-205K is to provide a structured resource for measuring performance across controlled sources of difficulty, rather than evaluating VTON models only through aggregate paired-test performance. To improve reliability at scale, we annotate a subset of generated samples and train a lightweight quality-filtering model that filters out triplets predicted to be corrupted or inconsistent. CURVTON-205K explicitly includes occlusion-rich examples and is stratified into three tiers of increasing difficulty, enabling controlled evaluation under progressively harder conditions and supporting curriculum-aware experimentation.

Our contributions can be summarized as follows: 
(1) We design a scalable multi-model data generation and curation pipeline that synthesizes diverse virtual try-on triplets without requiring parsing-mask annotations or dense-pose estimation.
(2) We release CURVTON-205K\footnote{Code: \codeurl; dataset: \dataseturl.}, a large-scale dataset containing 205,000 curated mask-free virtual try-on triplets for training and 4,500 triplets for testing, designed to broaden coverage across garment types, styles, person appearance, poses, occlusions, and capture conditions.
(3) We introduce a difficulty-tiered organization of the dataset, enabling evaluation across increasingly challenging try-on scenarios and supporting curriculum-aware experimentation.
(4) We conduct comprehensive experiments across four VTON architectures, analyzing try-on quality, cross-dataset generalization, and performance under different difficulty levels, thereby examining data-centric design as a complement to architectural advances.

\section{Related Work}

\noindent \textbf{VITON Approaches:}
VTON methods have evolved primarily through architectural advances in alignment and generation. Early GAN-based pipelines combined explicit geometric warping with adversarial rendering, from VITON~\cite{Han2018VITON} and CP-VTON~\cite{Wang2018CPVTON} to texture-preserving and high-resolution extensions such as CP-VTON+~\cite{Minar2020CPVTONPLUS} and VITON-HD~\cite{Choi2021VITONHD}. Later methods introduced stronger conditioning signals to improve robustness to pose variation and occlusion, including semantic/layout guidance, flow-based deformation, contextual reasoning, and pose-aware synthesis~\cite{Han2019ClothFlow,Yang2020ACGPN,Fele2021DCTON}.
Parser-free approaches, such as PF-AFN~\cite{Ge2021PFAFN}, reduce dependence on auxiliary parsing signals, while recent diffusion-based methods substantially improve photorealism and garment preservation~\cite{Kim2024StableVITON, Xu2024OOTDiffusion, Choi2024IDMVTON}.
Despite these advances, VTON performance remains strongly shaped by the coverage and biases of the underlying training data, motivating datasets that explicitly target diversity and controllable difficulty.

\noindent \textbf{VITON datasets:}
Dataset design strongly shapes VTON performance and failure modes. DeepFashion~\cite{Liu2016DeepFashion} is a large-scale fashion benchmark with rich attribute supervision, but it was not designed for try-on and does not provide explicit try-on triplets. Most VTON research instead relies on shop-to-model paired benchmarks collected under controlled conditions, notably VITON-HD~\cite{Choi2021VITONHD}, which provides high-resolution frontal-view person--garment pairs for upper-body try-on, and DressCode~\cite{Morelli2022DressCode}, which broadens category coverage (upper-body, lower-body, dresses). More recent datasets incorporate more diverse capture conditions, including variations in pose, background, lighting, and occlusions, as in StreetTryOn~\cite{Yang2023StreetTryOn} and MPV~\cite{Guo2023MPV}. Recent advancements such as OpenVTON-Bench~\cite{li2026openvtonbenchlargescalehighresolutionbenchmark} incorporate increased poses and garment diversity, and address shortcomings of evaluation metrics. Dress-ED~\cite{dress_ed} augments DressCode person--garment pairs with instruction-conditioned edited counterparts, substantially broadening the task space to unified VTON, VTOFF, and text-guided garment editing. However, existing datasets remain constrained by coverage gaps: occlusions, garment--person interactions, and under-represented appearance attributes. Many pipelines also depend on costly and error-prone annotations such as masks or parsing maps. These limitations motivate synthetic datasets with broader coverage, mask-free triplet supervision, and controlled difficulty structure. Rather than expanding the task space, CURVTON-205K targets robustness within conventional VTON, organizing supervision across increasingly difficult pose, occlusion, viewpoint, background, and garment--person interaction conditions, which enables controlled analysis of how performance degrades across difficulty regimes rather than only in aggregate.

\noindent \textbf{Synthetic Data and Curriculum Learning:}
Synthetic data has long served as a practical way to scale supervision in vision when real-world data is costly or difficult to obtain. Classic examples include synthetic driving benchmarks (SYNTHIA~\cite{RosSMVL16_Synthia}), or game-engine data with pixel-accurate labels~\cite{RichterVRK16_PlayingForData}. Pico-Banana-400K~\cite{Qian2025PicoBanana} further illustrates the use of large-scale image-transformation pairs for training and benchmarking. In VTON, however, synthetic generation has been explored primarily to support limited supervision~\cite{LiSP25_SyntheticPairsVTON}, rather than to construct a curated large-scale dataset explicitly designed for broad coverage mask-free supervision, and structured difficulty. Curriculum learning formalizes the idea of ordering training examples from easy to hard to facilitate optimization and improve generalization~\cite{BengioLCW09_Curriculum}. Our use of difficulty tiers is motivated by this principle but the primary goal is broader: CURVTON-205K organizes samples into levels of increasing visual complexity reflecting the main challenges in VTON (e.g. occlusions, clutter, pose). This structure supports curriculum-aware training while providing a controlled difficulty-aware lens for evaluating robustness beyond aggregated performance.

\section{CURVTON-205K Dataset Construction}
\label{sec:dataset_construction}

\begin{figure*}[!t]
    \centering
    \includegraphics[width=\textwidth, height=0.8\textheight,keepaspectratio]{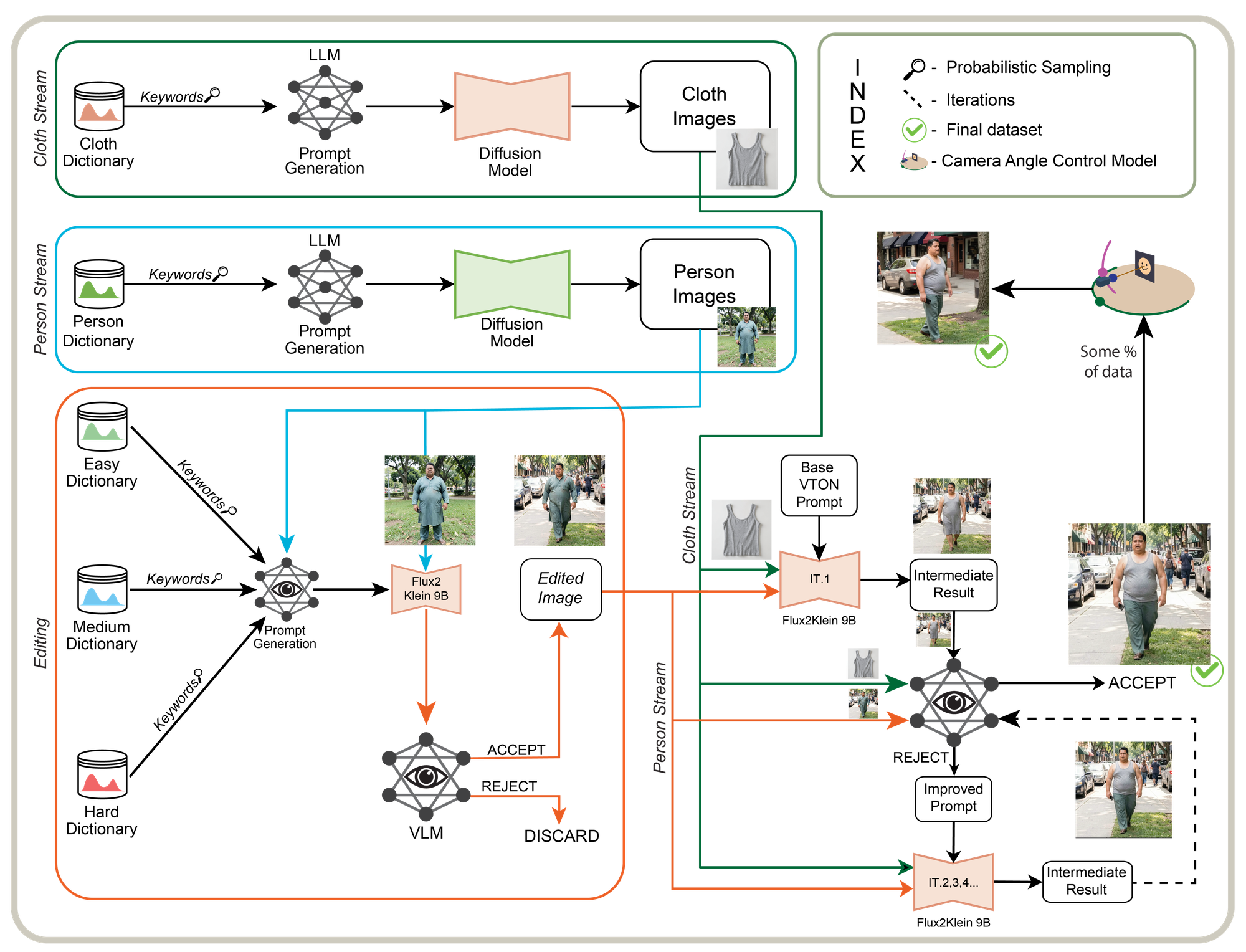}
    \vspace{-10pt}
    \caption{Data generation pipeline with clothing (orange), person (green), editing (blue), and merging (red) streams producing mask-free triplets.}
    \label{fig:stage1}
\end{figure*}

CURVTON-205K is built through the multi-stage synthesis and curation pipeline illustrated in Fig.~\ref{fig:stage1}. It contains four stages: garment generation, person generation, difficulty-aware person editing, and try-on synthesis, followed by quality screening. This pipeline generates mask-free triplets $(I_p, I_c, I_t)$, where $I_p$ is a source person image, $I_c$ is a target garment image, and $I_t$ is the corresponding try-on image.

\subsection{Data Synthesis Pipeline}\label{subsec:data_synth}

\noindent \textbf{Garment Stream:}
For each garment type $c_i \in \mathcal{D}_c$, we sample multiple prompt variants using an LLM conditioned on a clothing-specific system prompt. Each prompt variant captures different visual attributes, such as color, material, pattern, cut, and style. We then generate multiple clothing images per prompt with the text-to-image generator FLUX.2 [klein] 9B~\cite{Flux2Klein9B}, denoted $G_{\text{Flux9B}}$. Overall, this yields a clothing image set $\mathcal{I}_c = \{I_c^1, \ldots, I_c^{N_c}\}$ of size $N_c = 35{,}000$ for the dataset released with this paper.

\noindent\textbf{Person Stream:}
We define a probabilistic attribute dictionary $\mathcal{D}_p$ spanning (i) body-shape categories, (ii) geographic, cultural, and ethnic style cues,\footnote{We use non-identifying appearance groupings to encourage representation diversity; see Suppl. for details.} (iii) accessories, e.g., spectacles, (iv) visible physical disabilities and assistive devices, e.g., wheelchair, (v) background context, and (vi) photographic style. For photographic style, we use a 50\% studio/professional and 50\% in-the-wild split. We sample attributes from a factorized distribution $\mathbb{P}_{\mathcal{D}_p}$ to control coverage and avoid long-tail collapse.

For each person image, we sample an attribute tuple from $\mathbb{P}_{\mathcal{D}_p}$ and use an LLM with a person-specific system prompt to generate a detailed, realistic text prompt. We synthesize $N_p=41{,}000$ person images at $1024\times1024$ resolution using $G_{\text{Flux9B}}$, yielding $\mathcal{I}_p = \{I_p^1, \ldots, I_p^{N_p}\}$.

\noindent\textbf{Sample Difficulty Stream:}
To introduce controllable difficulty, we define a collection of editing dictionaries $\mathcal{E}=\{\mathcal{E}_{\text{easy}},\allowbreak \mathcal{E}_{\text{medium}},\allowbreak \mathcal{E}_{\text{hard}}\}$, with three tiers corresponding to increasing geometric and visual complexity through pose deviation, occlusion, clutter, and viewpoint changes. For each person image $I_p^j$ and each tier in $\mathcal{E}$, we sample editing keywords to generate multiple edited variants. This produces $|\mathcal{I}_{p,\text{edit}}| = 300{,}000$ edited person images, i.e., $100{,}000$ per difficulty tier.

\noindent\textbf{Merging Stream:}
We pair person images, both the original set $\mathcal{I}_p$ and the edited set $\mathcal{I}_{p,\text{edit}}$, with corresponding target garment images to synthesize try-on outputs, yielding candidate triplets that are subsequently curated into CURVTON-205K (Subsection~\ref{subsec:quality_screening_pipeline}). Given $\mathcal{I}_{p,\text{total}} = \mathcal{I}_p \cup \mathcal{I}_{p,\text{edit}}$, where $|\mathcal{I}_{p,\text{total}}|=341{,}000$, and the garment pool $\mathcal{I}_c$, we sample person-garment pairs according to a pairing policy $\mathbb{P}_{\text{pair}}$ that balances garment categories and difficulty tiers. For each sampled pair, we generate the corresponding try-on image using $G_{\text{Flux9B}}$ and form a candidate triplet $x_n=(I_{p, n},I_{c, n},I_{t, n})$.

To improve robustness at scale, we include an extensive validation loop, depicted in Fig.~\ref{fig:stage1}, that identifies inconsistent or low-quality generations and triggers regeneration when needed. Concretely, for each candidate triplet $x_n=(I_{p, n},I_{c, n},I_{t, n})$, 
comprising a source person image $I_{p, n}$, a target garment image $I_{c, n}$, and a synthesized try-on image $I_{t, n}$, 
we use Qwen3-VL 32B Instruct~\cite{Qwen3VL} as a VLM-based assessor.
Through a dedicated system prompt, the assessor evaluates the consistency of the synthesized try-on image $I_{t, n}$ with the source person image $I_{p, n}$ and target garment image $I_{c, n}$. The assessment focuses on criteria such as garment faithfulness, visual artifacts, placement plausibility, and semantic consistency. If the VLM-based assessor detects an inconsistency, the sample is flagged for regeneration, the associated prompts are refined, and the image is generated again. Unlike rejection sampling, the refinement loop uses the detected failure reason to revise the generation prompt and produce a new candidate, increasing the likelihood of obtaining valid samples for challenging configurations. This process is repeated up to $T=4$ times. Once $I_{t, n}$ is deemed consistent, the triplet is accepted into $\mathcal{D}_{\text{cand}}$ for downstream curation (see 
Suppl.\ for the specific improvements achieved through refinement iterations). This automated validation stage results in the candidate set $\mathcal{D}_{\text{cand}} = \{x_n\}_{n=1}^{N}$ with $N \ge 209{,}500$, which is then filtered by the quality screening pipeline to obtain the final CURVTON-205K, comprising $205{,}000$ training and $4{,}500$ test samples.

\subsection{Quality Screening Pipeline}
\label{subsec:quality_screening_pipeline}

To promote semantic consistency and visual quality at scale, we apply a two-stage screening process that combines limited manual annotation with automated filtering.

\textbf{Stage 1: Manual Labeling and Filter Model Training.}
After refinement through the feedback loop in the merging stream, the resulting candidates may still contain generation artifacts. Thus, we annotate a subset of $5{,}000$ generated triplets, sampled to cover all garment categories and difficulty tiers, with a binary quality label. Annotation follows a fixed 10-question artifact rubric that is available in the source code repository of the dataset (\codeurl).
Each triplet is labeled \emph{positive} if (i) the target garment identity, including type, color, and pattern, is preserved, (ii) the garment is plausibly worn with correct placement and deformation, and (iii) the output is free of major artifacts; otherwise it is labeled \emph{negative}. This labeled set is used to train a lightweight quality filter based on ViT-Small~\cite{Dosovitskiy2021ViT}, denoted $f_{\text{ViT}}$. The filter receives the three triplet images concatenated channel-wise, denoted by $\phi(I_{p,n}, I_{c,n}, I_{t,n})$, and is trained with binary cross-entropy to predict whether the sample is positive or negative. The model prediction is used in Stage 2 as a proxy for sample quality.

\textbf{Stage 2: Automated Screening.}
We deploy the trained ViT filter $f_\text{ViT}$ to score the pool of generated candidate triplets $\mathcal{D}_{\text{cand}}$ as $s_n = f_\text{ViT}(\phi(I_{p,n}, I_{c,n}, I_{t,n}))$ and retain samples satisfying $s_n \ge \tau$; we use $\tau{=}0.85$ unless otherwise stated. To assess potential distribution shifts introduced by filtering, we monitor acceptance rates across difficulty tiers, garment categories, and appearance and capture-condition attributes, including possible collapse in long-tail regimes. This two-stage procedure, based on an annotated subset of samples, scales curation to large volumes while preserving defined quality criteria, resulting in the final CURVTON-205K dataset consisting of 205,000 mask-free triplets for training and 4,500 triplets for testing.

\subsection{Dataset Overview}
\label{subsec:dataset_overview}

CURVTON-205K is a curated synthetic dataset of mask-free virtual try-on triplets designed to support model training and difficulty-stratified evaluation. Each sample contains a source person image, a target garment image, and the corresponding synthesized try-on image. CURVTON-205K provides direct triplet supervision without parsing masks or dense-pose annotations, thereby supporting pipelines that reduce dependence on auxiliary preprocessing models. The training and test partitions are disjoint by construction, $\mathcal{D}_{\text{train}} \cap \mathcal{D}_{\text{test}} = \emptyset$, so no triplet used for evaluation occurs anywhere in the training data.

The dataset is constructed to broaden coverage across garments, subjects, and capture conditions. Garments are drawn from a dictionary of 200 types, including common fashion categories as well as long-tail, traditional, and culturally specific clothing. Person prompts are sampled from a factorized attribute dictionary spanning body shape, geographic and cultural style cues, accessories, visible assistive devices, background context, and photographic style. In addition, CURVTON-205K is stratified into three difficulty tiers (easy, medium, and hard) introduced through editing operations designed to increase pose variation, occlusion, clutter, viewpoint change, and garment--person interaction complexity. This tiered structure enables difficulty-specific analysis and supports curriculum-aware training experiments.

Table~\ref{tab:dataset_comparison} contextualizes CURVTON-205K with respect to widely used VTON datasets. Prior benchmarks are predominantly real, shop-to-model paired datasets and are commonly used with auxiliary signals such as masks, parsing maps, or pose estimates. They also typically lack explicit stratification by difficulty. In contrast, CURVTON-205K provides large-scale mask-free triplets with explicit diversity targets and difficulty tiers, including targeted coverage of long-tail garment styles that may be under-represented in standard benchmarks.

\begin{table*}[t]
\centering
\caption{Comparison of virtual try-on datasets across key characteristics. Diversity coverage summarizes whether a dataset explicitly accounts for both garment diversity and person-appearance diversity: Limited indicates neither, Moderate indicates one of the two, and Broad indicates both. All listed datasets are publicly available except WildVTON~\cite{WildVTON}.}
\label{tab:dataset_comparison}
\scriptsize
\begin{tabular}{lcccccccc}
\toprule
\textbf{Dataset}
  & \textbf{\shortstack{Train}}
  & \textbf{\shortstack{Test}}
  & \textbf{\shortstack{Type}}
  & \textbf{\shortstack{Mask-\\Free}}
  & \textbf{\shortstack{Diff.\\tiers}}
  & \textbf{\shortstack{In-the-\\Wild}}
  & \textbf{\shortstack{Cloth\\Pairs}}
  & \textbf{\shortstack{Diversity\\coverage}} \\
\midrule
VITON~\cite{Han2018VITON}
  & 14K
  & 2K
  & Real
  & $\times$
  & $\times$
  & $\times$
  & $\checkmark$
  & Limited \\
VITON-HD
  & 12.3K
  & 1.4K
  & Real
  & $\times$
  & $\times$
  & $\times$
  & $\checkmark$
  & Limited \\
DressCode
  & 50K
  & 3K
  & Real
  & $\times$
  & $\times$
  & $\times$
  & $\checkmark$
  & Limited \\
StreetTryOn
  & 12.5K
  & 1.9K
  & Real
  & $\times$
  & $\times$
  & $\checkmark$
  & $\times$
  & Moderate \\
WildVTON
  & N/A
  & 1.2K
  & Hybrid
  & $\checkmark$
  & $\times$
  & $\checkmark$
  & $\checkmark$
  & Limited \\
OpenVTON-Bench
  & $\sim$90K
  & $\sim$10K
  & Real
  & $\checkmark$
  & $\times$
  & $\checkmark$
  & $\checkmark$
  & Moderate \\
\textbf{CURVTON-205K}
  & \textbf{205K}
  & \textbf{4.5K}
  & \textbf{Synth.}
  & $\checkmark$
  & $\checkmark$
  & $\checkmark$
  & $\checkmark$
  & Broad \\
\bottomrule
\end{tabular}
\end{table*}


\section{Experiments}\label{sec:experiments}

\begin{figure}[t]
  \centering
  \includegraphics[width=0.6\linewidth]{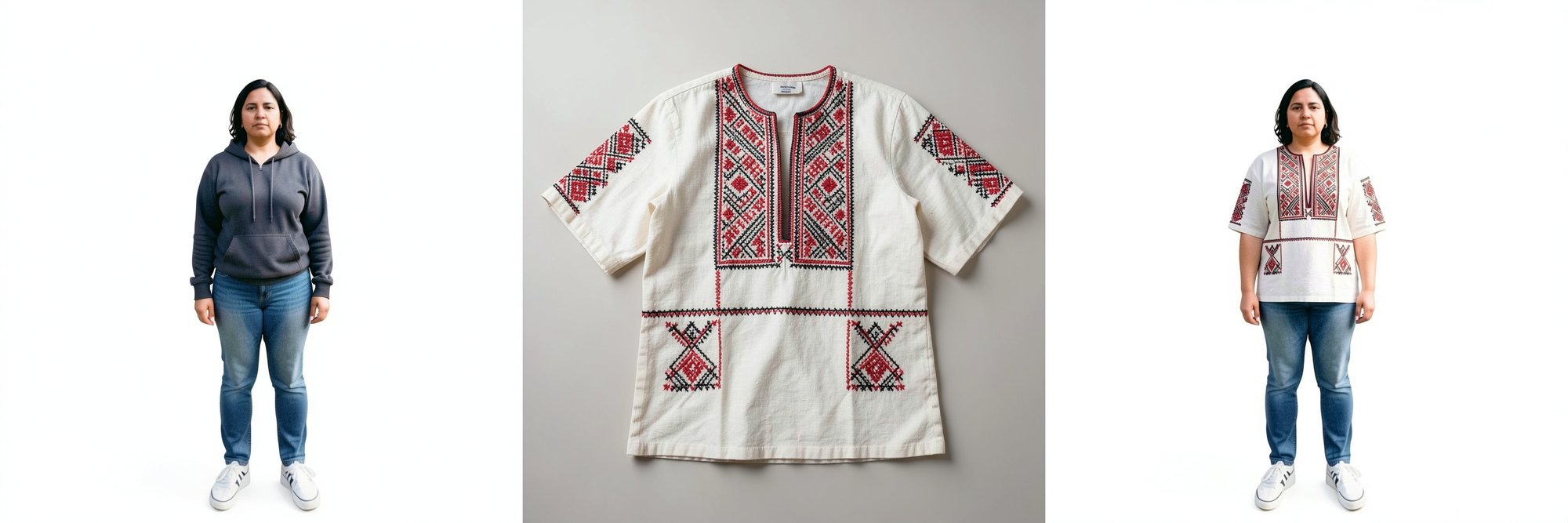}\\[-4pt]
  \includegraphics[width=0.6\linewidth]{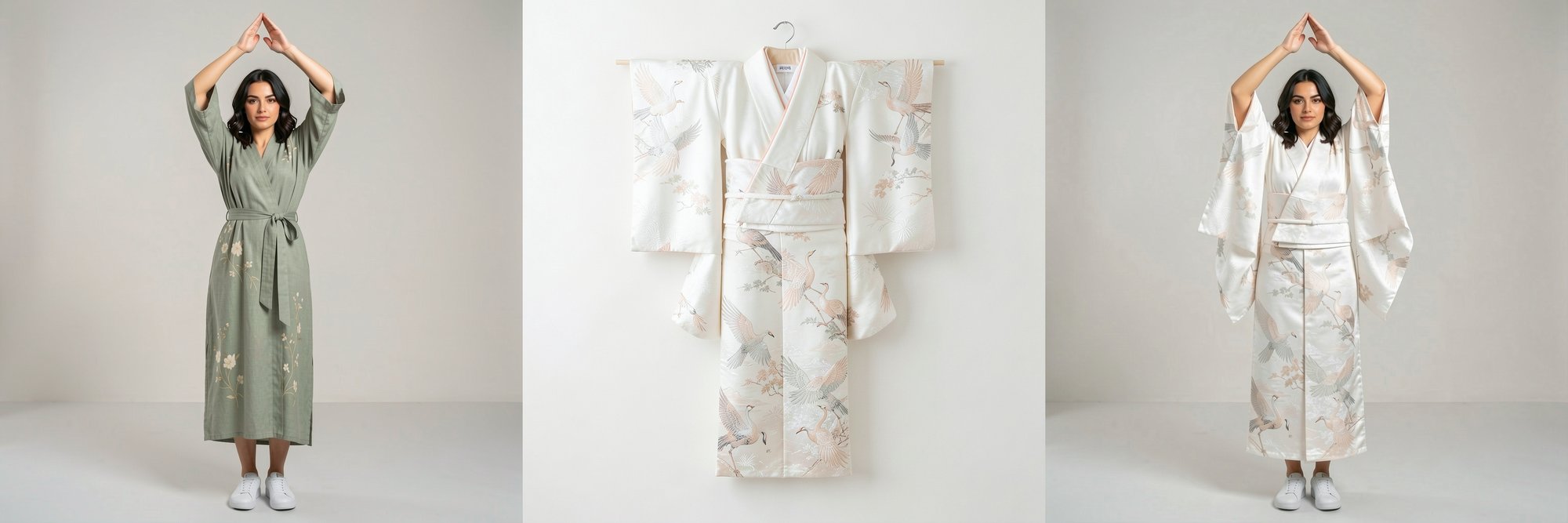}\\[-4pt]
  \includegraphics[width=0.6\linewidth]{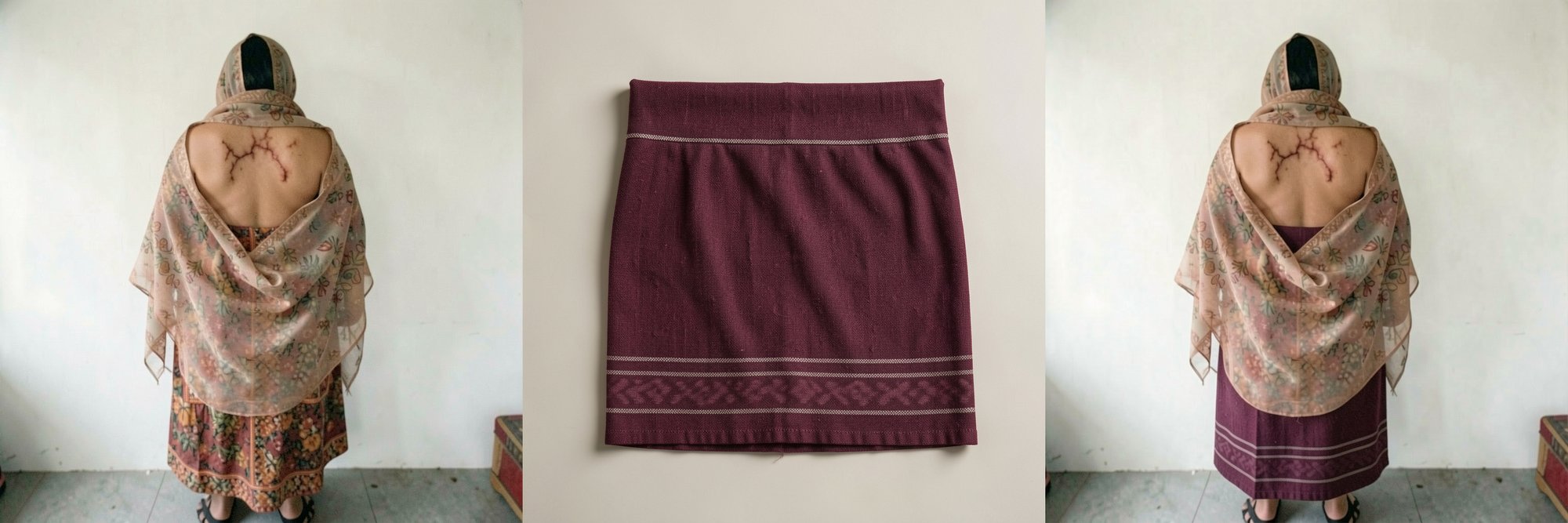}
  \caption{Try-on examples across difficulty tiers. Rows show \textbf{Easy}, \textbf{Medium}, and \textbf{Hard} samples, with increasing pose variation, occlusion, background complexity, and garment structure. Each triplet shows the source person, target garment, and synthesized try-on result.}
  \label{fig:qual_composite}
\end{figure}

First, we evaluate representative VTON approaches on CURVTON-205K to establish the dataset as a benchmark and analyze performance across difficulty tiers. We then assess cross-dataset generalization by training models with different data sources and evaluating them on standard paired test splits from VITON-HD~\cite{Choi2021VITONHD} and DressCode~\cite{Morelli2022DressCode}. Second, we analyze CURVTON-205K from a data-centric perspective comparing against broader fashion corpora and in-the-wild benchmarks, e.g., StreetTryOn~\cite{Yang2023StreetTryOn}. Third, curriculum-aware training ablations using the proposed difficulty tiers are reported in the Suppl. Paired perceptual metrics (SSIM~\cite{Wang2004SSIM} and LPIPS~\cite{Zhang2018LPIPS}) are computed on the corresponding test splits and distribution-level metrics (FID~\cite{Heusel2017FID} and KID~\cite{Binkowski2018KID}) are computed over the full test partition. Qualitative examples across difficulty tiers are shown in Fig.~\ref{fig:qual_composite}.

\subsection{VTON-Based Evaluations}

\begin{table*}[t]
  \caption{Stratified FID performance (lower is better) of all models trained and evaluated on CURVTON-205K. Results are reported overall and by difficulty tier and cloth type: upper (UB) and lower body (LB); garment (G); and traditional (Tr) and non-traditional (NTr). Full results with additional metrics are provided in the Suppl.}
  \label{tab:stratified_perf}
  \centering
  \scriptsize
  \setlength{\tabcolsep}{2pt}
  \renewcommand{\arraystretch}{0.95}
  \begin{tabular}{ll c ccc ccccc}
    \toprule
      \multicolumn{2}{c}{ }
      & {Overall}
      & \multicolumn{3}{c}{Difficulty Tier}
      & \multicolumn{5}{c}{Cloth Type} \\
    \cmidrule(lr){4-6} \cmidrule(lr){7-11}
      & &
      & Easy & Med. & Hard
      & UB & LB & G & Tr & NTr \\
    \midrule
    \multirow{2}{*}{OOTDiffusion~\cite{Xu2024OOTDiffusion}}
      & Zero Shot
      & 18.90 & 25.20 & 24.40 & 25.50 & 19.50 & 27.15 & 29.70 & 29.95 & 17.15 \\
      & Training
      & 8.75 & 9.88 & 11.54 & 14.12 & 11.42 & 11.83 & 13.79 & 14.22 & 13.83 \\
    \midrule
    \multirow{2}{*}{CatVTON~\cite{Chong2025CatVTON}}
      & Zero Shot
      & 23.85 & 22.55 & 24.40 & 27.55 & 22.60 & 23.55 & 28.90 & 31.35 & 20.16 \\
      & Training
      & 9.14 & 9.14 & 9.23 & 12.88 & 10.62 & 11.34 & 14.94 & 15.57 & 12.70\\
    \midrule
    \multirow{2}{*}{IDM-VTON~\cite{Choi2024IDMVTON}}
      & Zero Shot
      & 17.05 & 16.20 & 17.30 & 19.30 & 16.70 & 17.25 & 22.00 & 25.55 & 15.15 \\
      & Training
      & 8.44 & 8.28 & 8.85 & 10.29 & 10.21 & 10.57 & 13.54 & 15.75 & 7.98 \\
    \midrule
    \multirow{1}{*}{\textbf{FLUX.2 [klein] 9B (Generator)}}
      & Zero Shot
      & 14.50 & 13.10 & 15.35 & 17.95 & 14.75 & 16.05 & 17.50 & 23.27 & 20.23 \\
    \bottomrule
  \end{tabular}
\end{table*}

\begin{table*}[t]
\centering
\caption{
Cross-dataset generalization on VITON-HD (VHD), DressCode (DC),
and StreetTryOn. For VHD and DC, SSIM and LPIPS measure paired
reconstruction quality, while FID and KID measure distribution-level
quality. StreetTryOn is evaluated in the unpaired in-the-wild setting
using FID and KID.
}
\label{tab:cross_dataset_gen}

\renewcommand{\arraystretch}{0.9}
\scriptsize
\begin{tabular}{ll cccc cccc cc}
\toprule

\multirow{2}{*}{Model}
& \multirow{2}{*}{Train Data}
& \multicolumn{4}{c}{VITON-HD}
& \multicolumn{4}{c}{DressCode}
& \multicolumn{2}{c}{StreetTryOn} \\

\cmidrule(lr){3-6}
\cmidrule(lr){7-10}
\cmidrule(lr){11-12}

& &
SSIM$\uparrow$ &
LPIPS$\downarrow$ &
FID$\downarrow$ &
KID$\downarrow$ &
SSIM$\uparrow$ &
LPIPS$\downarrow$ &
FID$\downarrow$ &
KID$\downarrow$ &
FID$\downarrow$ &
KID$\downarrow$ \\

\midrule


\multirow{3}{*}{CatVTON~\cite{Chong2025CatVTON}}

& Real baseline
& 0.87 & 0.11 & 9.01 & 1.09
& 0.89 & 0.10 & 6.14 & 1.40
& 25.31 & 6.95 \\

& CURVTON-205K
& 0.78 & 0.13 & 11.74 & 1.41
& 0.80 & 0.12 & 7.98 & 1.79
& 21.36 & 5.42 \\

& Real + CURVTON
& 0.89 & 0.10 & 8.72 & 1.04
& 0.90 & 0.10 & 5.92 & 1.36
& 20.45 & 5.18 \\

\midrule


\multirow{3}{*}{IDM-VTON~\cite{Choi2024IDMVTON}}

& Real baseline
& 0.85 & 0.12 & 9.26 & 1.27
& 0.89 & 0.12 & 14.13 & 6.03
& 23.62 & 6.18 \\

& CURVTON-205K
& 0.80 & 0.13 & 11.02 & 1.46
& 0.83 & 0.13 & 16.94 & 6.88
& 19.08 & 4.96 \\

& Real + CURVTON
& 0.87 & 0.11 & 8.98 & 1.22
& 0.90 & 0.11 & 13.68 & 5.86
& 18.55 & 4.82 \\

\midrule


\multirow{3}{*}{StableVITON~\cite{Kim2024StableVITON}}

& Real baseline
& 0.85 & 0.15 & 8.93 & 2.54
& 0.88 & 0.11 & 13.60 & 6.05
& 23.15 & 4.63 \\

& CURVTON-205K
& 0.77 & 0.18 & 11.01 & 3.08
& 0.80 & 0.13 & 17.08 & 7.11
& 18.44 & 3.71 \\

& Real + CURVTON
& 0.86 & 0.14 & 8.71 & 2.46
& 0.89 & 0.11 & 13.05 & 5.72
& 17.95 & 3.60 \\

\midrule


\multirow{3}{*}{OOTDiffusion~\cite{Xu2024OOTDiffusion}}

& Real baseline
& 0.83 & 0.14 & 11.73 & 3.15
& 0.85 & 0.12 & 9.21 & 2.66
& 27.62 & 8.08 \\

& CURVTON-205K
& 0.75 & 0.16 & 12.21 & 4.90
& 0.77 & 0.15 & 17.73 & 3.45
& 23.31 & 7.25 \\

& Real + CURVTON
& 0.84 & 0.13 & 11.62 & 3.19
& 0.87 & 0.11 & 9.08 & 2.61
& 18.70 & 4.58 \\

\bottomrule
\end{tabular}

\end{table*}

\begin{table*}[!htbp]
  \centering
  \caption{Task-free dataset diversity comparison across seven dimensions, each split into Complexity (C) and Diversity (D) sub-metrics. Bold: best; underline: second best.}
  \label{tab:diversity_comparison}
  \begingroup
  \setlength{\tabcolsep}{4pt}
  \renewcommand{\arraystretch}{0.9}
\scriptsize
\begin{tabular}{l *{13}{c}}
    \toprule
    Dataset
      & \multicolumn{1}{c}{Pose}
      & \multicolumn{2}{c}{Occlusion}
      & \multicolumn{3}{c}{Background}
      & \multicolumn{2}{c}{Illumination}
      & \multicolumn{1}{c}{Body Shape}
      & \multicolumn{2}{c}{Garment}
      & \multicolumn{2}{c}{Camera} \\
    \cmidrule(lr){2-2}
    \cmidrule(lr){3-4}
    \cmidrule(lr){5-7}
    \cmidrule(lr){8-9}
    \cmidrule(lr){10-10}
    \cmidrule(lr){11-12}
    \cmidrule(lr){13-14}
      & $\sigma^2_{\mathrm{pose}}\uparrow$
      & \shortstack{Occl.\\Cmplx.$\uparrow$}
      & \shortstack{Person\\Vis.$\downarrow$}
      & \shortstack{BGE\\(C)$\uparrow$}
      & \shortstack{BSE\\(D)$\uparrow$}
      & \shortstack{BOD\\(D)$\uparrow$}
      & IGM
      & \shortstack{LVM\\(D)$\uparrow$}
      & $\sigma^2_{\mathrm{shape}}\uparrow$
      & \shortstack{Gmt\\Var.$\uparrow$}
      & \shortstack{Gmt\\Div.$\uparrow$}
      & $\hat{\sigma}_\phi\uparrow$
      & $\hat{\sigma}_\theta\uparrow$ \\
    \midrule
    VITON-HD
      & 145.55
      & 0.52 & 0.31
      & 2.93 & 0.01 & 0.29
      & 0.60 & 0.06
      & 0.54
      & 475.81 & 0.04
      & 0.070 & 0.090 \\
    DressCode
      & 127.79
      & 0.45 & 0.45
      & 1.60 & 0.07 & 0.56
      & 0.73 & 0.10
      & 0.76
      & 669.59 & 0.26
      & 0.140 & 0.070 \\
    StreetTryOn
      & \textbf{204.48}
      & \underline{0.62} & \textbf{0.23}
      & 4.57 & \underline{0.54} & 3.15
      & 0.45 & 0.10
      & 0.47
      & 621.65 & 0.15
      & 0.120 & 0.140 \\
    Dress-ED
      & 129.92
      & 0.43 & 0.43
      & 1.63 & 0.07 & 0.60
      & 0.80 & 0.09
      & 0.77
      & 725.11 & 0.32
      & 0.170 & 0.113 \\
    \midrule
    CURVTON Easy
      & 66.68
      & 0.53 & 0.33
      & 4.00 & 0.43 & 3.45
      & 0.50 & \textbf{0.16}
      & 0.62
      & \textbf{831.32} & \textbf{0.52}
      & 0.022 & 0.066 \\
    CURVTON Medium
      & 178.45
      & \underline{0.62} & 0.27
      & \underline{4.78} & \textbf{0.57} & \textbf{5.17}
      & 0.37 & 0.09
      & 0.74
      & 821.99 & 0.49
      & 0.347 & 0.292 \\
    CURVTON Hard
      & \underline{201.88}
      & \textbf{0.66} & \underline{0.24}
      & \textbf{4.81} & 0.50 & \underline{5.09}
      & 0.33 & 0.09
      & \textbf{1.02}
      & \underline{827.66} & \underline{0.51}
      & \textbf{1.197} & \textbf{0.422} \\
    CURVTON-205K (All)
      & 182.22
      & 0.61 & 0.28
      & 4.54 & 0.50 & 4.61
      & 0.40 & \underline{0.12}
      & \underline{0.89}
      & 823.06 & 0.50
      & \underline{0.683} & \underline{0.295} \\
    \bottomrule
  \end{tabular}
  \endgroup
\end{table*}

CURVTON-205K defines a stratified evaluation benchmark with explicit difficulty tiers (\emph{easy/medium/hard}) and garment-type metadata. This enables reporting performance by tier and clothing category, providing targeted insight into failure modes beyond aggregate scores. Table~\ref{tab:stratified_perf} reports the performance of representative VTON approaches trained and evaluated on CURVTON-205K. The reported results show model FID across tiers of difficulty and cloth types; full results, including SSIM and LPIPS and an additional DCI-VTON~\cite{Gou2023DCIVTON} baseline, are provided in the Suppl. For each method, we report the original pretrained model (\emph{Zero Shot}) and the corresponding model trained or fine-tuned on CURVTON-205K. Before CURVTON adaptation, most models obtain progressively higher FID on harder tiers and exhibit substantial performance differences across garment categories. Training on CURVTON-205K substantially improves in-domain performance within every reported difficulty tier and garment-category subset, while the persistent easy-to-hard degradation suggests that the tier structure remains informative after adaptation. Because all garment-category subsets contain the same number of samples, their FID scores can be compared directly. For all three models, CURVTON training reduces the range between the lowest and highest category-specific FID—from 12.80 to 2.80 for OOTDiffusion, from 11.19 to 4.95 for CatVTON~\cite{Chong2025CatVTON}, and from 10.40 to 7.77 for IDM-VTON. This consistent reduction indicates more uniform performance across garment types, supporting the claim that CURVTON provides broad and relatively balanced category coverage rather than benefiting only a narrow subset.

Table~\ref{tab:cross_dataset_gen} evaluates cross-benchmark transfer under three training settings: the real VTON datasets used in the corresponding baselines, CURVTON-205K alone, and their combination, while keeping the test protocols fixed. We compare CatVTON, IDM-VTON, StableVITON, and OOTDiffusion trained on the individual VITON-HD and DressCode train sets.
For VITON-HD and DressCode, we synthesize source-person inputs while retaining the original real target images and garment references, and train all baselines with unmasked person inputs (see Suppl.).
The \emph{Real + CURVTON} setting augments each corresponding real training set with CURVTON-205K. The results for StreetTryOn, as in the original paper, use a combination of unpaired samples from different real datasets as the real-data baseline, to which CURVTON-205K is added in the combined setting. On the paired VITON-HD and DressCode tests, in-domain real-data training outperforms training on CURVTON alone on all reported metrics, revealing the remaining synthetic-to-real gap, although CURVTON-trained models remain competitive. Crucially, combining real and CURVTON data numerically improves most of the reported measurements over real-only training across all four architectures and both datasets, with two ties and only one minor regression. SSIM increases by 0.01--0.02 and FID decreases by 0.11--0.55 across all comparisons; LPIPS improves in six comparisons and remains unchanged in two, while KID improves in seven of eight comparisons. On StreetTryOn, CURVTON-trained models consistently achieve lower FID/KID than the real-data baselines across all architectures. The combined setting achieves the best FID and KID for every architecture, reducing FID by 4.86--8.92 and KID by 1.03--3.50 relative to real-only training. These results suggest that CURVTON-205K does not replace in-domain real supervision but effectively complements it with transferable synthetic supervision. In particular, improvements over real-only training demonstrate that its utility is not confined to reproducing the generator’s synthetic distribution, supporting the practical use of CURVTON to extend existing real datasets with broader coverage, particularly under out-of-benchmark, in-the-wild evaluation.

Table~\ref{tab:wild_grouped} reports ground-truth-free, in-the-wild evaluation of OOTDiffusion trained on different data sources, alongside pretrained FLUX.2 [klein] 9B and IDM-VTON references. Across the automatic criteria, training on CURVTON-205K yields the strongest results, achieving the highest garment-image alignment (CLIP-I) and visual consistency assessed by the VLM evaluator. Human evaluation further supports these trends through blind pairwise comparisons aggregated over 20 annotators who select the preferred images across models, marked with the same 10-question rubric (Section~\ref{app:rubric} of the Suppl.). Human Preference scores report the percentage of annotator preferences assigned to each model, normalized to sum to 100\% within each evaluation criterion. The model trained on CURVTON-205K achieves the highest preference for both try-on quality (35\%) and realism (28\%). Together, these results provide complementary evidence that CURVTON training improves perceptual quality and visual plausibility under less controlled, in-the-wild conditions.


%

\subsection{Task-Free Dataset Analysis}
\begin{figure}[tb]
  \centering
  \includegraphics[width=0.45\linewidth]{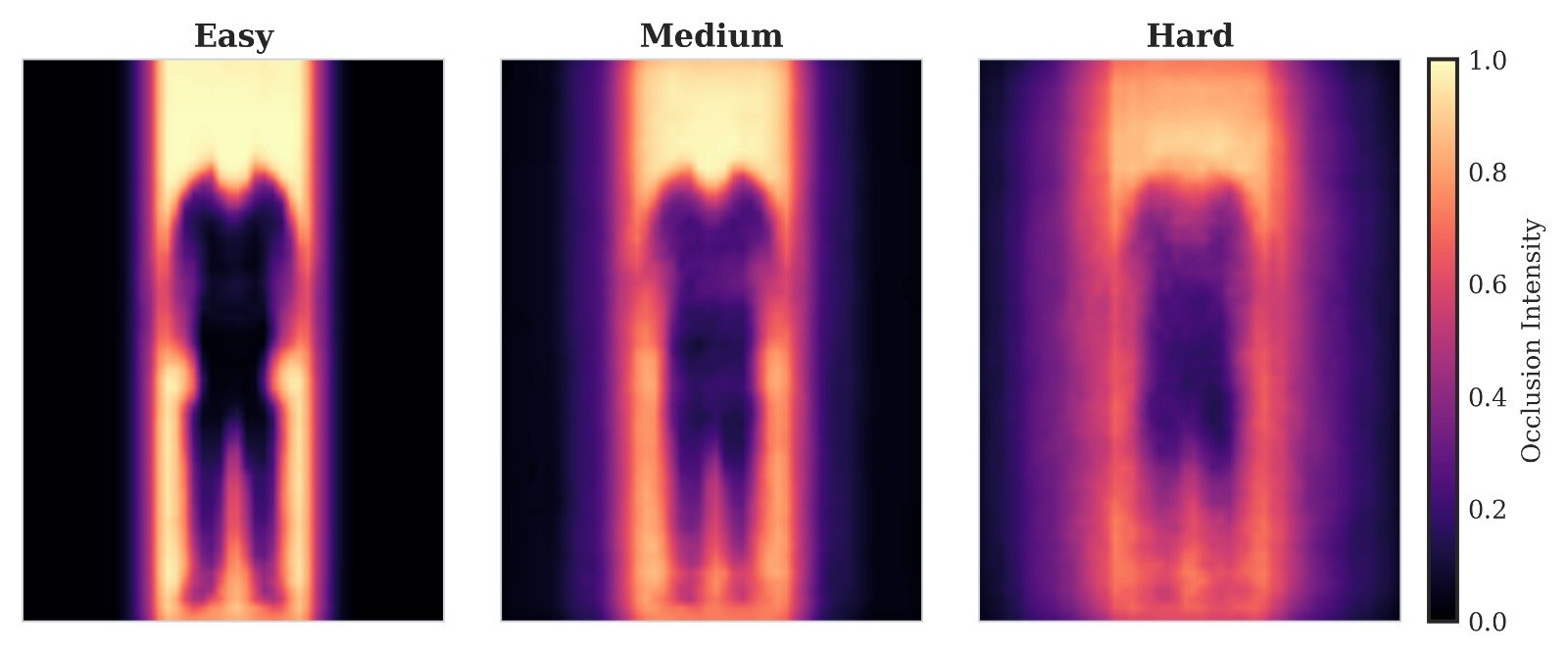}\hspace{4pt}
  \includegraphics[width=0.48\linewidth]{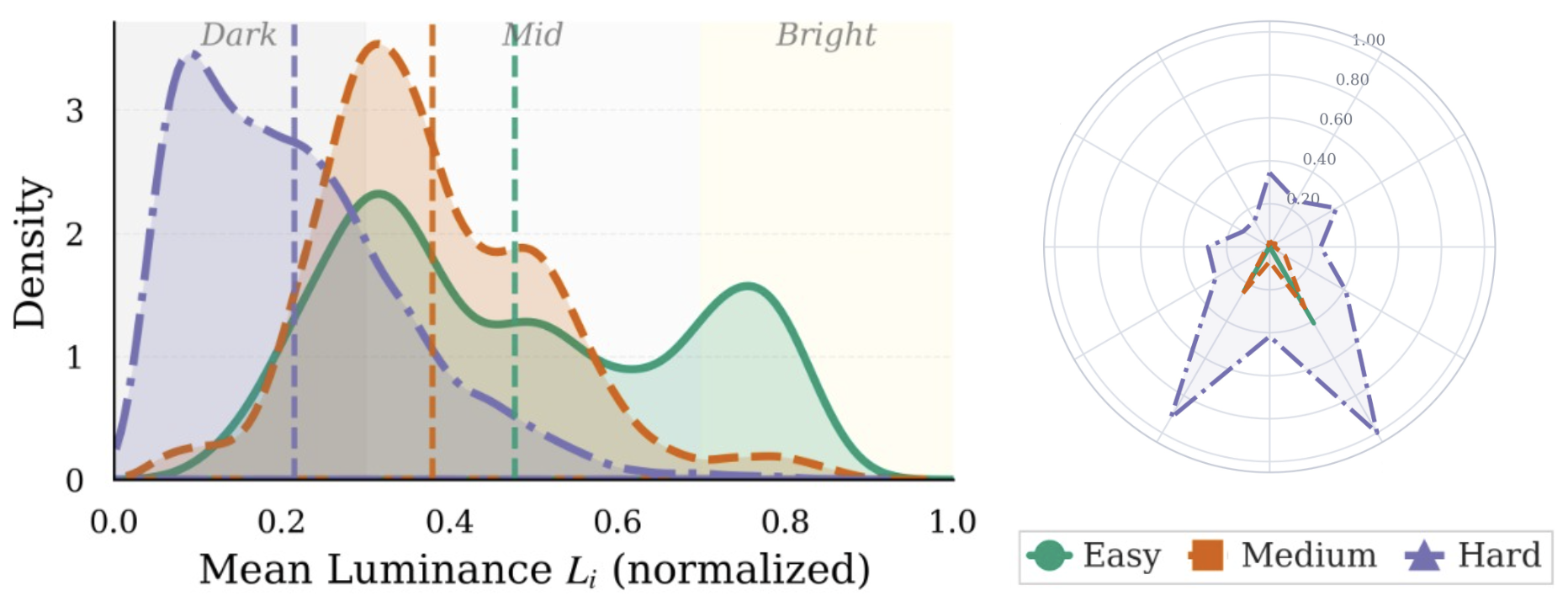}
\caption{Deployment realism of CURVTON-205K data via occlusions (left), illumination (center), and pose (right) distributions for the three difficulty tiers: easy, medium, and hard.}   
\label{fig:deployment_realism_occlusions}
\end{figure}

We provide task-free analyses to characterize the diversity and coverage of CURVTON-205K independently of any downstream VTON model. All results are computed under a fixed protocol using the same pretrained feature extractors and dataset-level descriptors, enabling consistent comparisons across datasets without confounding effects from downstream architecture or training choices. 
Table~\ref{tab:diversity_comparison} compares CURVTON-205K with existing VTON and fashion datasets across pose, occlusion, background, illumination, body-shape, garment, and camera dimensions; the reported metrics are defined in 
  Suppl.
Fig.~\ref{fig:deployment_realism_occlusions} complements this comparison by summarizing visual variation relevant to in-the-wild settings. The left panel shows normalized spatial occlusion heatmaps, the middle panel reports luminance distributions, and the right panel summarizes pose distributions across the three difficulty tiers. These visualizations characterize variation in occlusion patterns and illumination conditions while exposing differences in pose coverage across tiers.
Additional t-SNE~\cite{vanderMaaten2008tSNE} projections of garment and facial features are provided in 
Suppl. to visualize variation in the corresponding embedding spaces. 


Among the full-dataset comparisons, CURVTON-205K achieves the highest measured variation in background objects, luminance, body shape, garments, and camera viewpoint, whereas StreetTryOn retains the highest pose variation and strongest occlusion measures. Within CURVTON, the progression from easy to hard is accompanied by increasing pose variance, occlusion complexity, body-shape variation, and camera-viewpoint variation. These results support the intended difficulty structure and show that CURVTON provides broad, complementary coverage across several measured dimensions, rather than uniformly dominating existing datasets on every criterion.

\begin{figure}[t!]
  \centering
  \includegraphics[width=0.18\linewidth]{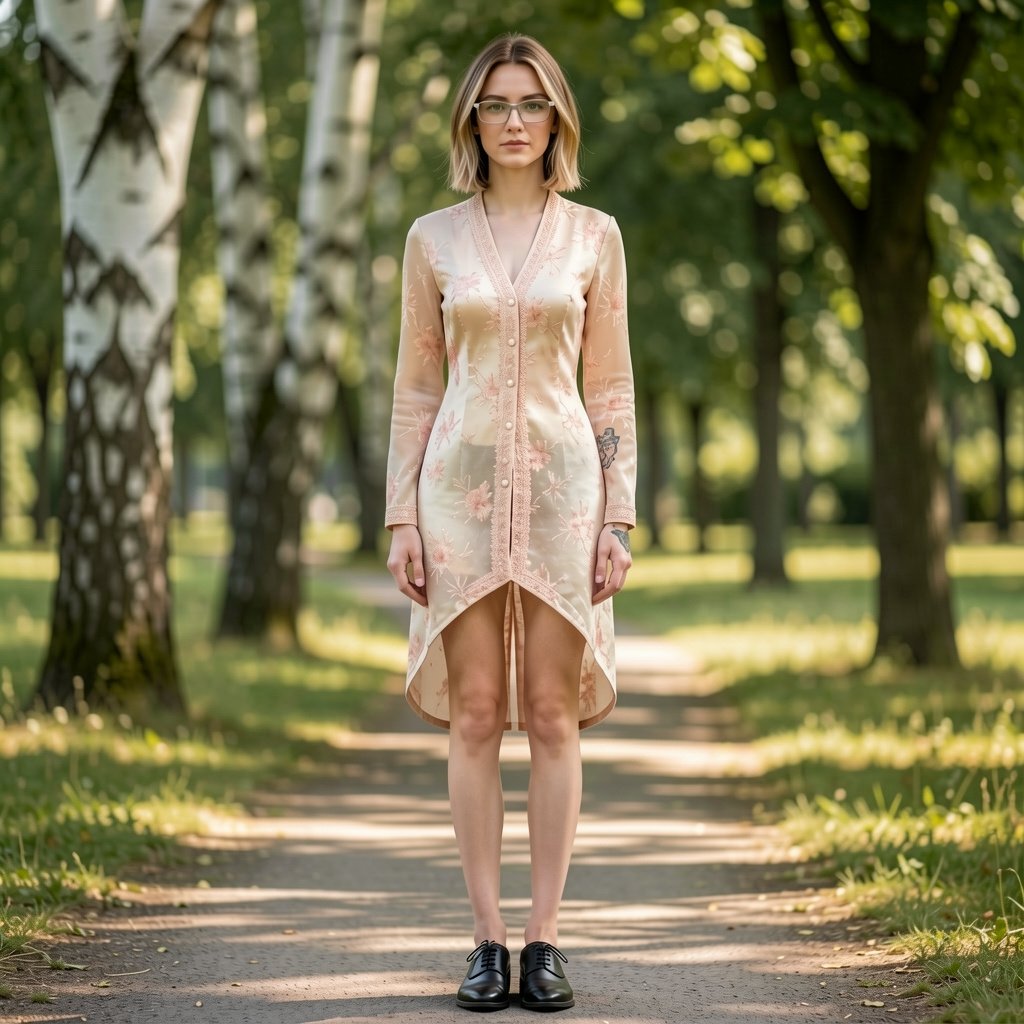}\hspace{-2.5pt}
  \includegraphics[width=0.18\linewidth]{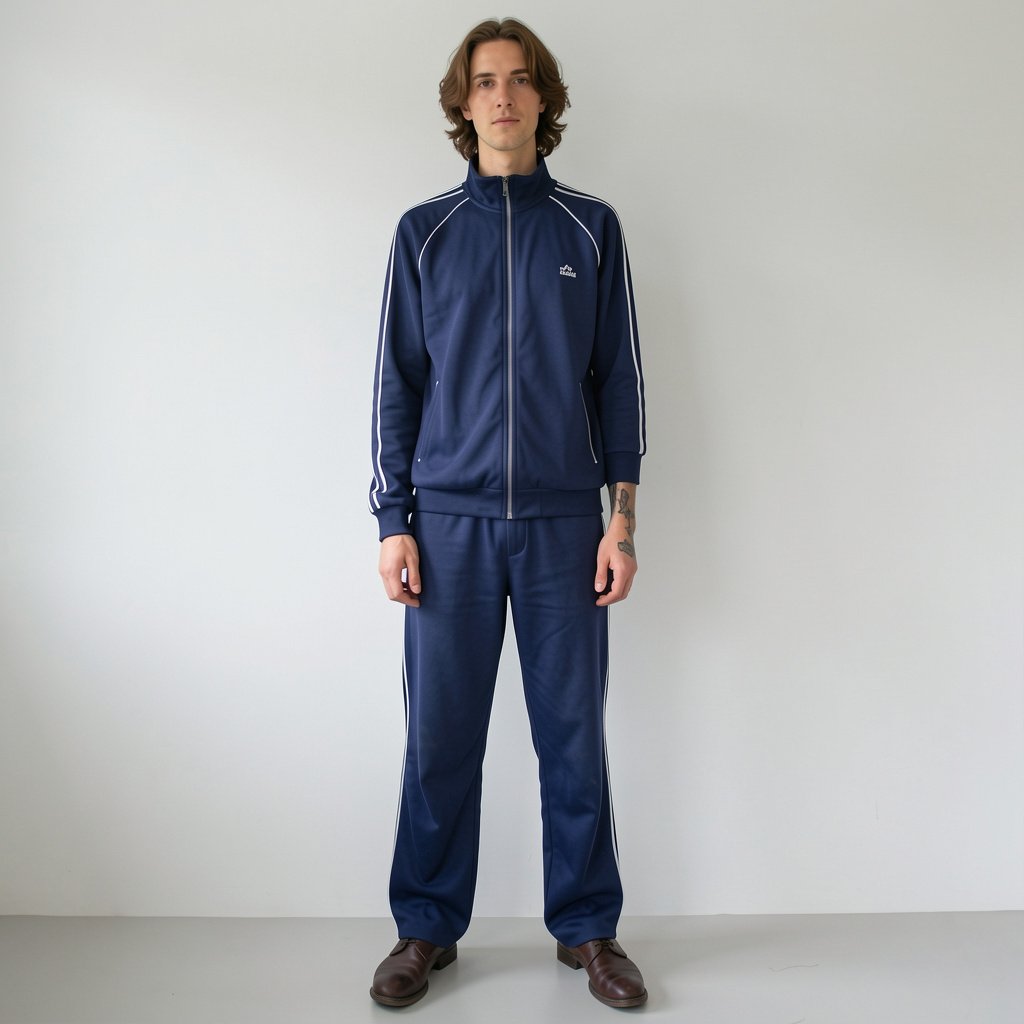}\hspace{-2.5pt}
  \includegraphics[width=0.18\linewidth]{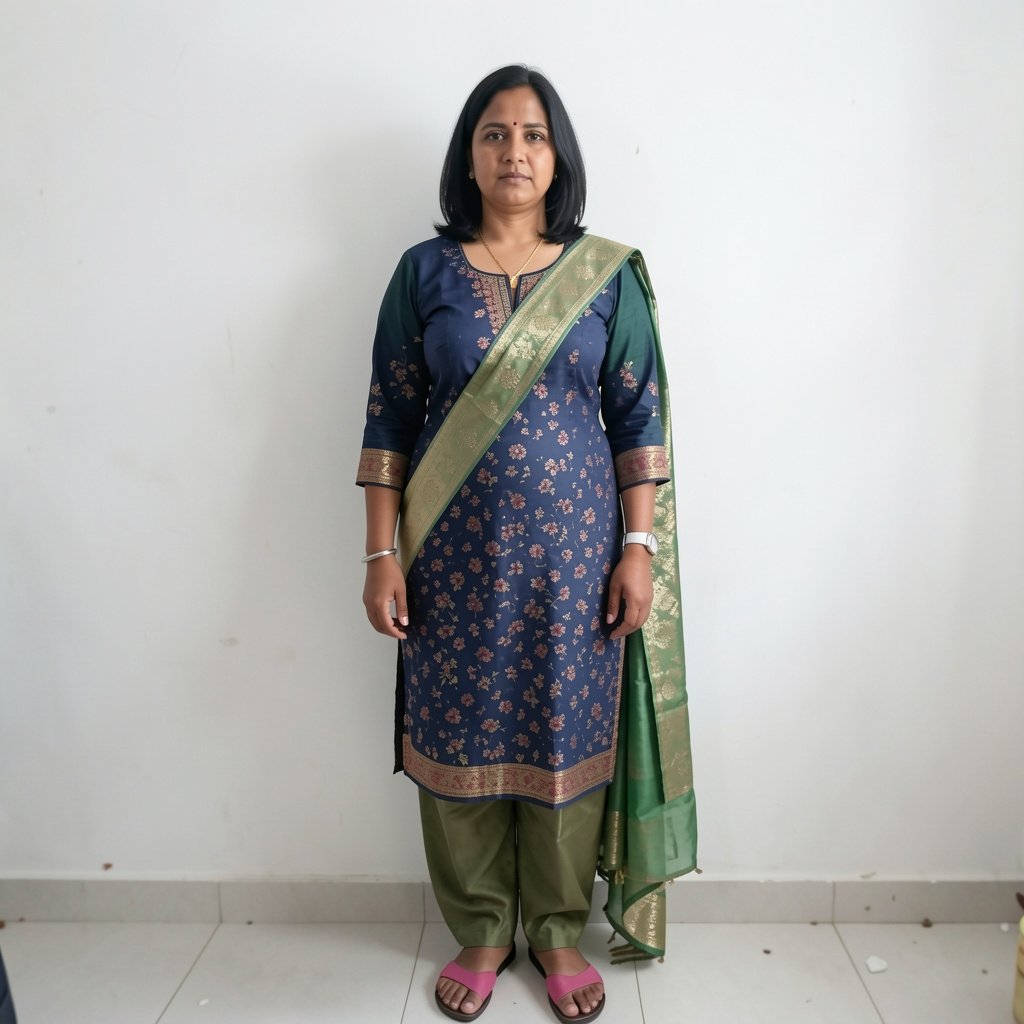}\hspace{-2.5pt}
  \includegraphics[width=0.18\linewidth]{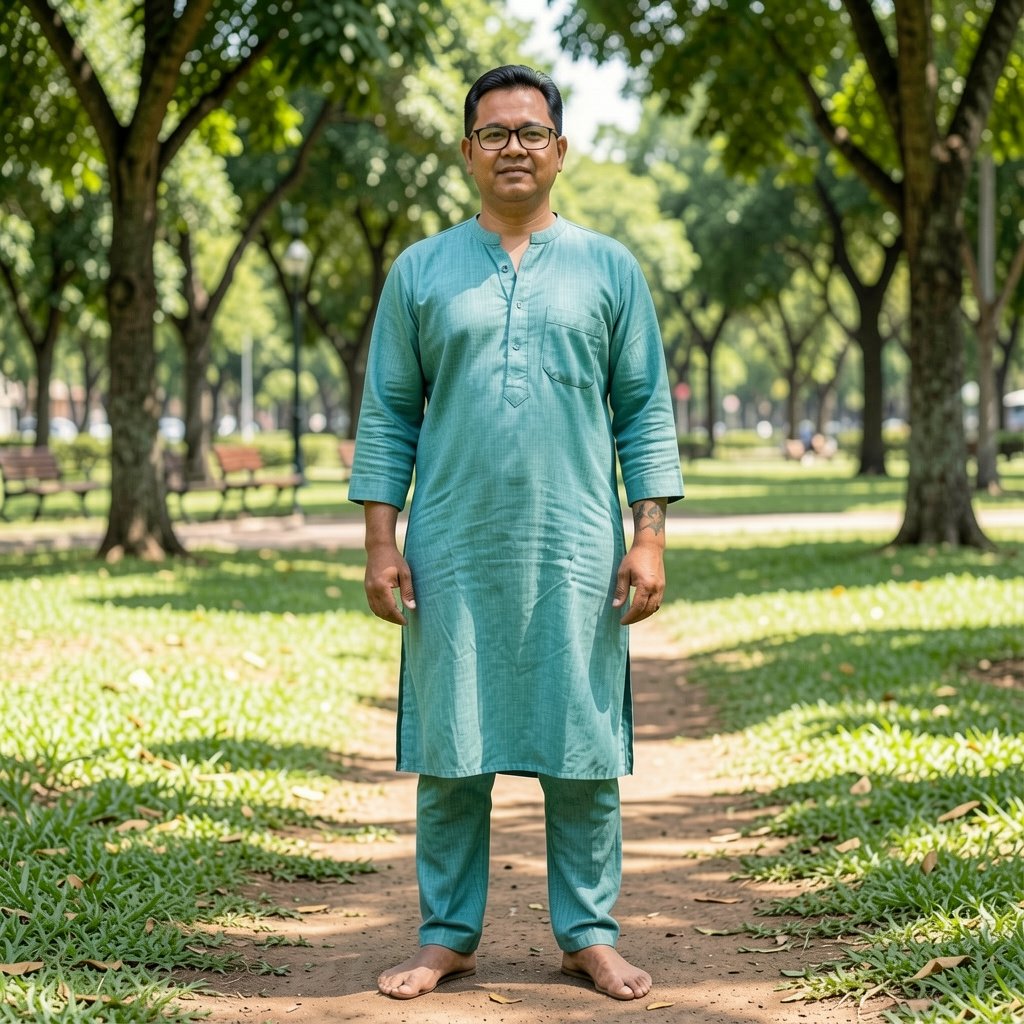}\hspace{-2.5pt}
  \includegraphics[width=0.18\linewidth]{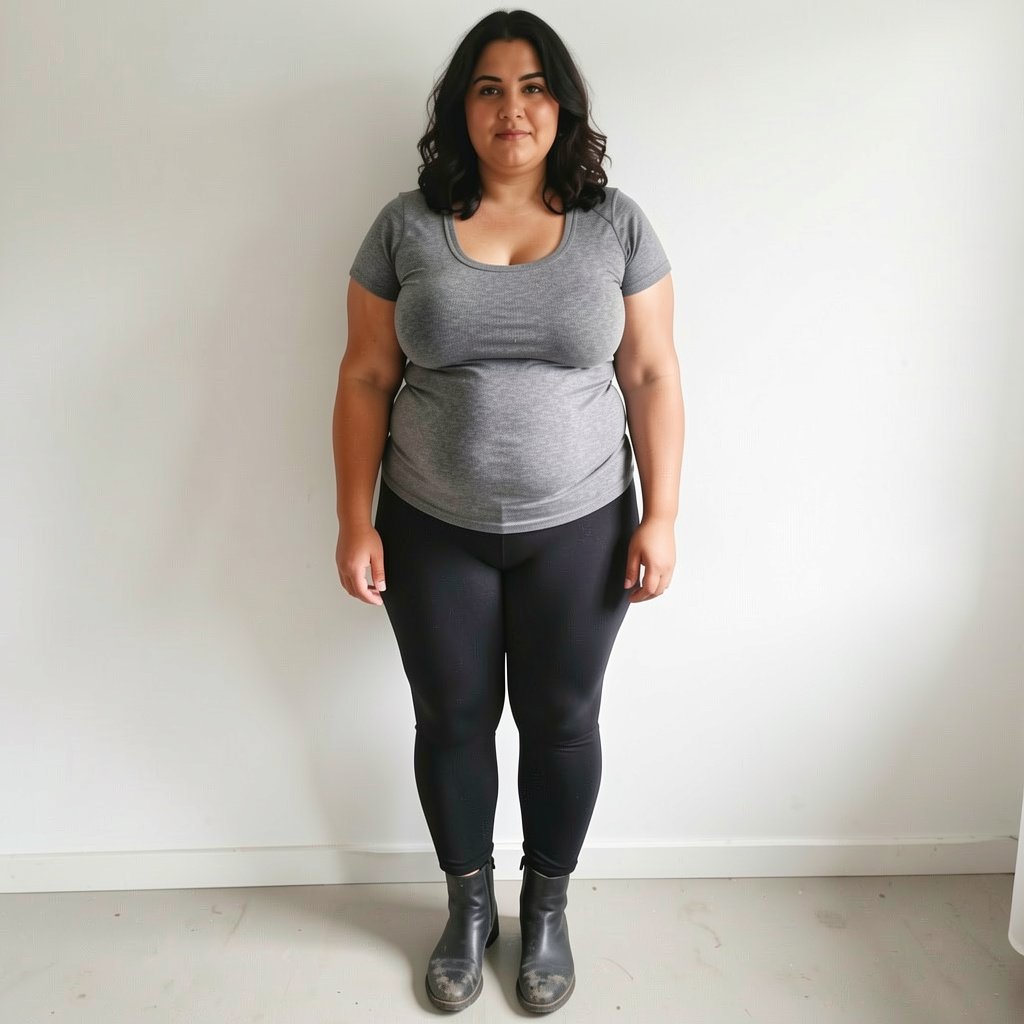}\\[-1pt]

  \includegraphics[width=0.18\linewidth]{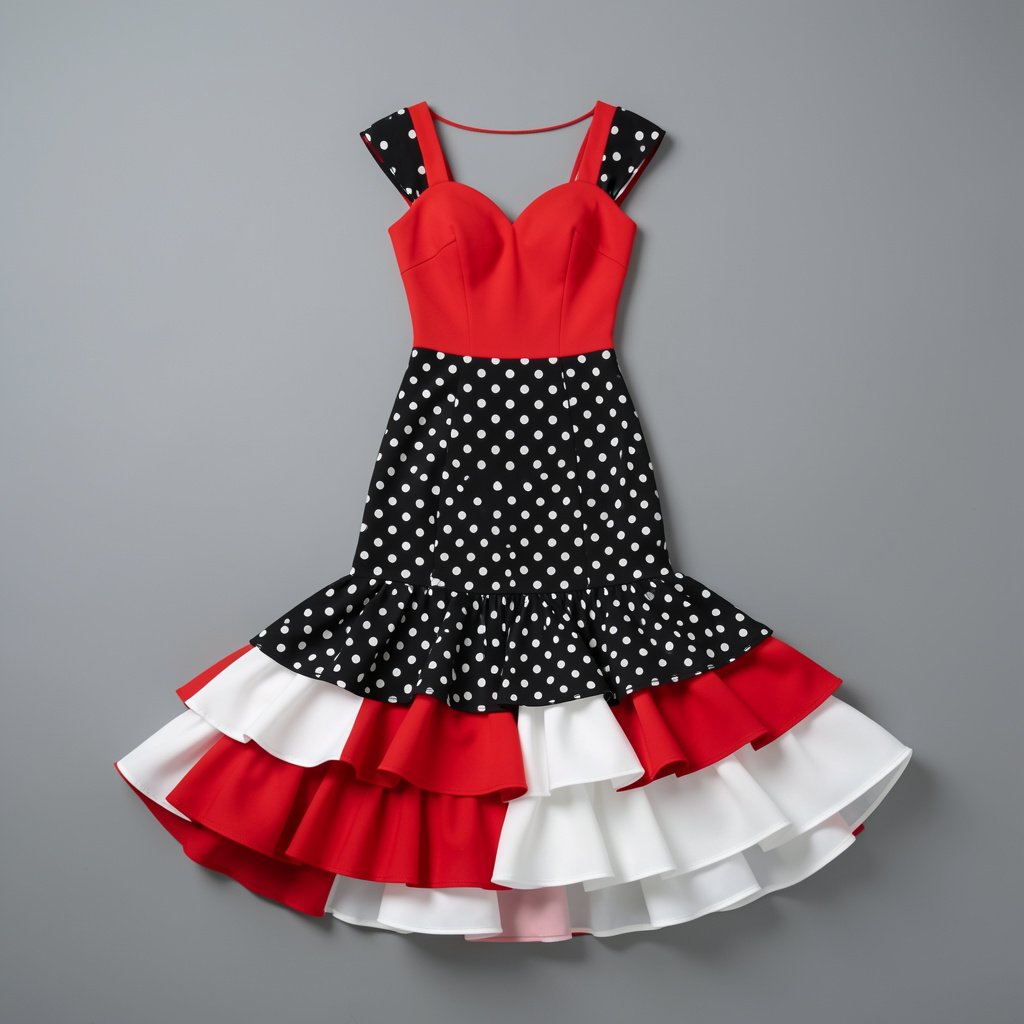}\hspace{-2.5pt}
  \includegraphics[width=0.18\linewidth]{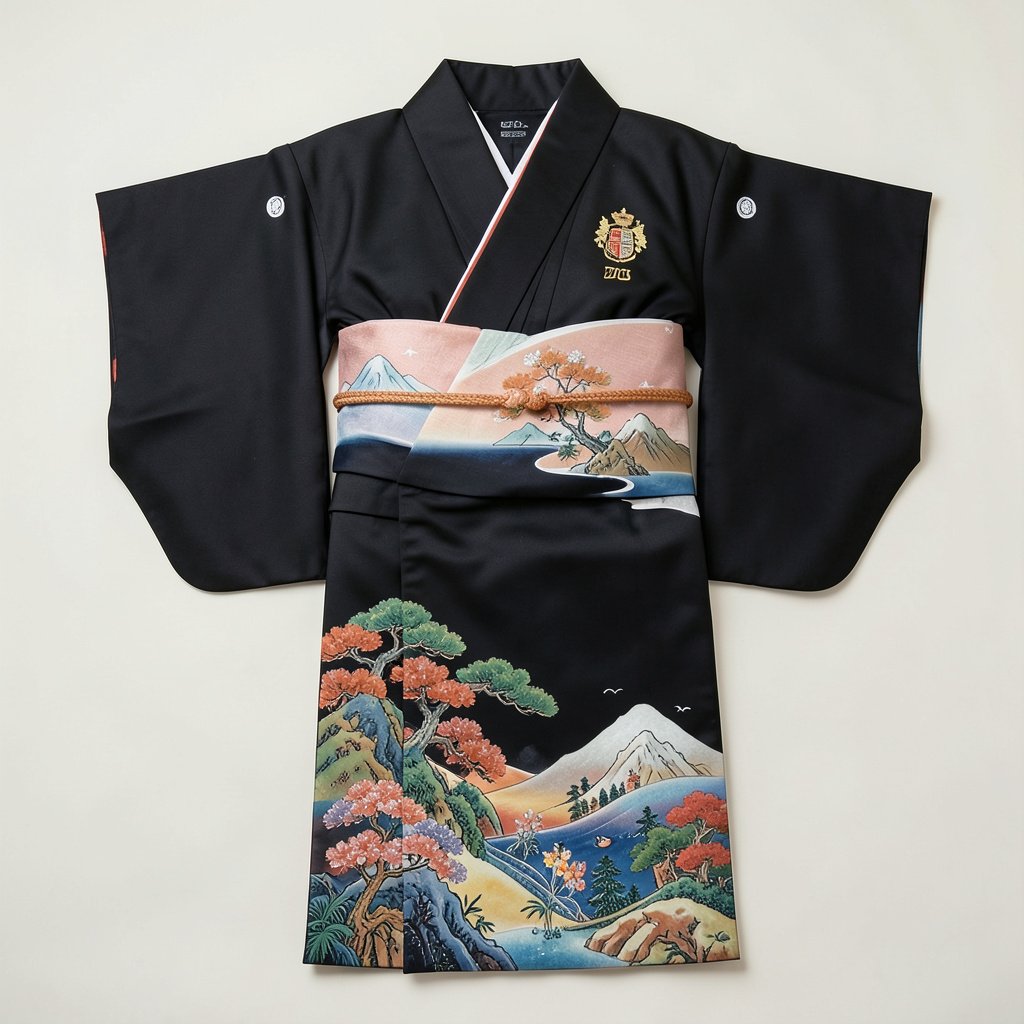}\hspace{-2.5pt}
  \includegraphics[width=0.18\linewidth]{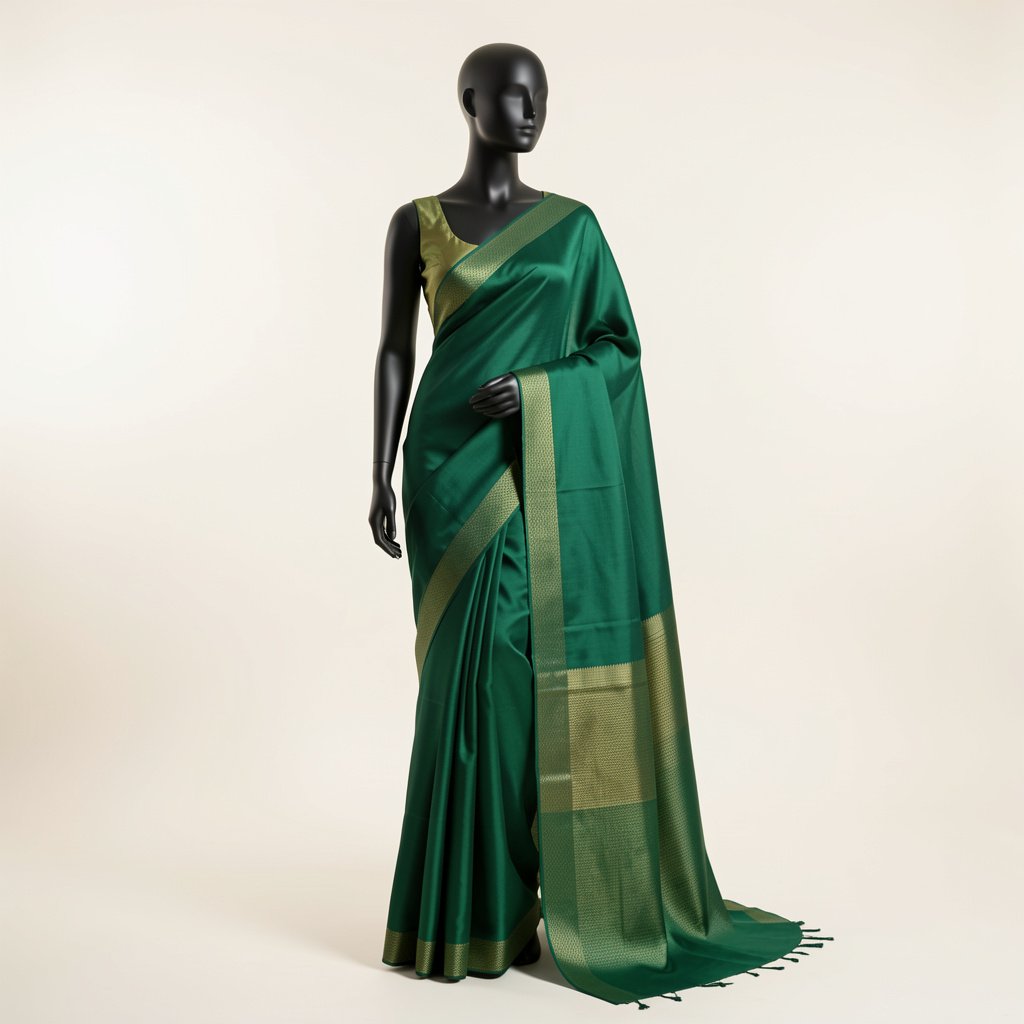}\hspace{-2.5pt}
  \includegraphics[width=0.18\linewidth]{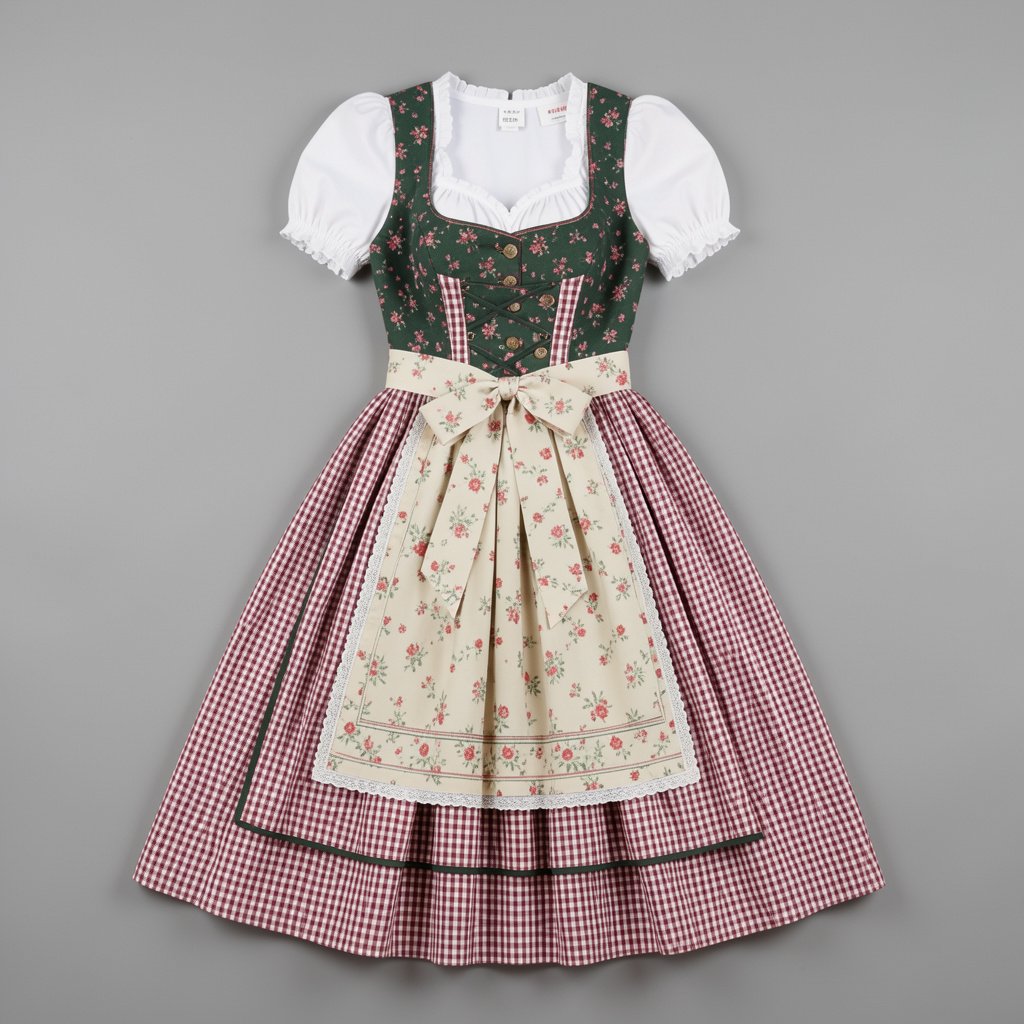}\hspace{-2.5pt}
  \includegraphics[width=0.18\linewidth]{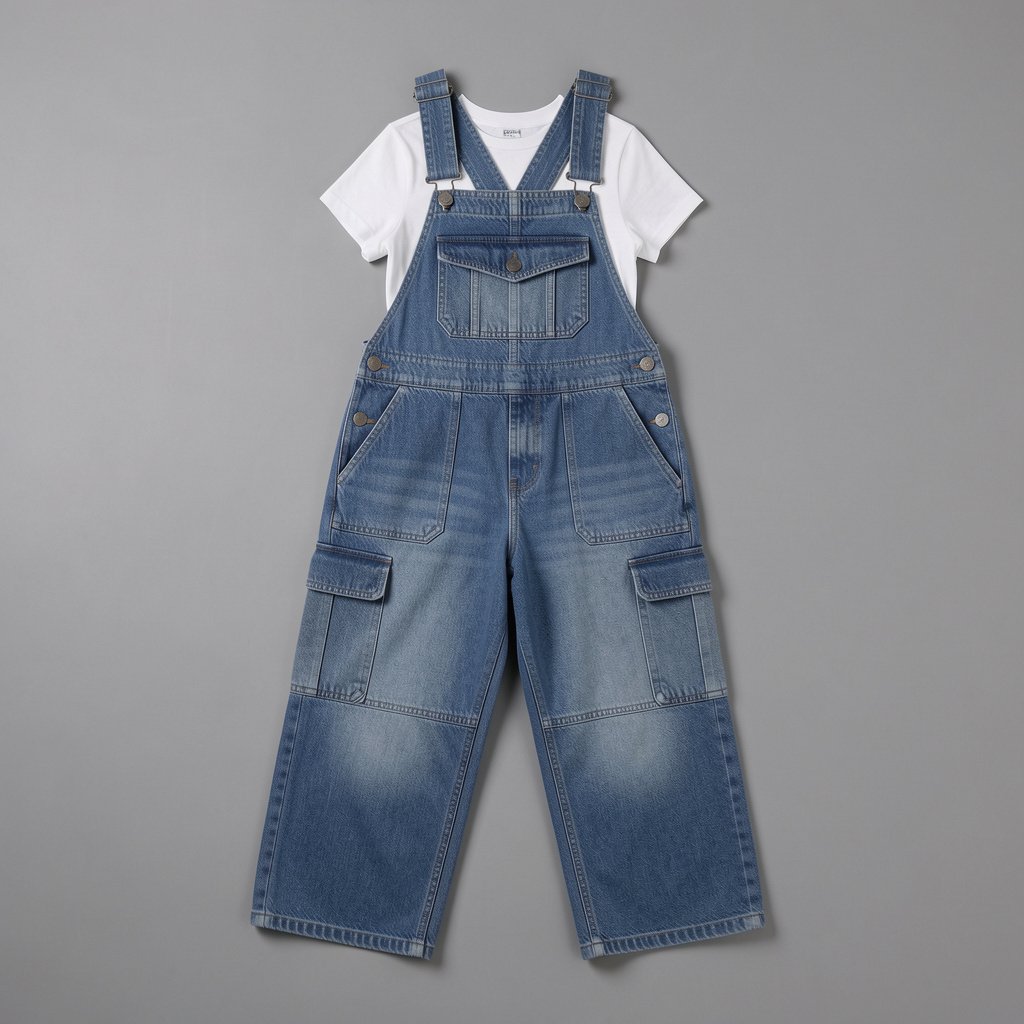}\\[-1pt]

  \caption{Diversity in CURVTON-205K: sampled person images (top) and in-shop garments (bottom) illustrating variation in identity cues, body shape, garment categories, and appearance.}
  \label{fig:person_and_garment}
\end{figure}

For an intuitive complement to these results, Fig.~\ref{fig:person_and_garment} shows representative samples from CURVTON-205K. The top row displays sampled person images and the bottom row shows sampled in-shop garments, illustrating variation in body shape, facial-appearance cues, garment categories, colors, and patterns. Overall, the quantitative and qualitative diagnostics provide evidence that CURVTON-205K covers a range of measured appearance factors, including garments, facial appearance, and body shape, as well as conditions relevant to in-the-wild evaluation, such as background, lighting, and occlusion. These analyses characterize visual and embedding-space coverage but should not be interpreted as a demographic or cultural fairness assessment.

\begin{table}[t]
\caption{In-the-wild evaluation of virtual try-on models. CLIP-I~\cite{Radford2021CLIP} and VLM measure semantic alignment, and Human Preference captures subjective evaluation of the generated Try-On images without ground truth supervision. FLUX.2 [klein] 9B and IDM-VTON are pretrained reference models. Bold: best; underline: second best.}
\label{tab:wild_grouped}
\centering
\scriptsize
\setlength{\tabcolsep}{4pt}
\renewcommand{\arraystretch}{0.9}
\begin{tabular}{l cc cc}
\toprule
\multirow{2}{*}{Train Data / Model}
  & \multirow{2}{*}{CLIP-I$\uparrow$}
  & \multirow{2}{*}{VLM$\uparrow$}
  & \multicolumn{2}{c}{Human Preference (\%)} \\
\cmidrule(lr){4-5}
  & & & Quality$\uparrow$
      & Realism$\uparrow$ \\
\midrule
VITON-HD
  & 0.71 & 0.69 & 10\% &
  11\% \\
DressCode
  & 0.74 & 0.73 & 12\% & 
  13\% \\
VHD + DC
  & 0.77 & 0.76 & 14\% & 
  15\% \\
FLUX.2 [klein] 9B (Pretrained)
  & 0.76 & 0.74 & 14\% & 
  14\% \\
IDM-VTON
  & \underline{0.80} & \underline{0.78} & \underline{15\%} & 
  \underline{19\%} \\
\midrule
\textbf{CURVTON-205K}
  & \textbf{0.83} & \textbf{0.86}
  & \textbf{35\%} & 
  \textbf{28\%} \\
\bottomrule
\end{tabular}
\end{table}





\section{Limitations}\label{sec:limitations}
While CURVTON-205K substantially expands mask-free supervision and controllable difficulty for virtual try-on, it inherits limitations common to large-scale generative data pipelines. Because the dataset is synthetically generated, its distribution may not fully capture real deployment settings, including camera characteristics, garment materials, lighting, and person--garment interactions; although our cross-dataset experiments indicate transfer to real-image benchmarks, further synthetic-to-real adaptation studies are needed to characterize performance in specific deployment domains. In a small fraction of cases, fine-grained garment attributes such as pockets or buttons are not perfectly preserved between the target garment and the synthesized try-on, leading to minor attribute drift; these effects are more likely in the hard tier, where reconstruction is intrinsically ambiguous. Finally, although CURVTON-205K broadens coverage across garment styles, person appearance, backgrounds, and visual conditions, it should not be read as exhaustively or demographically representing real-world populations and fashion practices. These limitations motivate future work on noise-aware training, generator-diverse synthesis, synthetic-to-real adaptation, and improved auditing of quality and representation in virtual try-on datasets.

\section{Conclusion}

We presented CURVTON-205K, a data-centric virtual try-on resource built
around the premise that robust VTON depends not only on model architecture,
but also on how training data covers challenging visual conditions.
CURVTON-205K provides 205,000 curated mask-free training triplets and
4,500 test triplets, organized into three difficulty tiers that progressively
vary pose, occlusion, viewpoint, background complexity, and garment--person
interaction. A scalable generation and curation pipeline combines specialized
generation and editing models with iterative VLM-guided refinement and learned
quality filtering. Across four VTON architectures, difficulty-stratified
evaluation reveals consistent performance degradation as visual complexity
increases, while augmenting real training data with CURVTON improves most
cross-benchmark measurements and substantially improves performance in the
in-the-wild StreetTryOn setting. Task-free analyses further show broad,
complementary coverage across several measured visual dimensions. Together,
these results position CURVTON-205K as a source of complementary synthetic
supervision and a structured resource for studying where VTON systems degrade
and how data design can improve robustness beyond aggregate benchmarks.


{
    \small
    \bibliographystyle{ieeenat_fullname}
    \bibliography{main_refs}

\begin{thebibliography}{51}
\providecommand{\natexlab}[1]{#1}
\providecommand{\url}[1]{\texttt{#1}}
\expandafter\ifx\csname urlstyle\endcsname\relax
  \providecommand{\doi}[1]{doi: #1}\else
  \providecommand{\doi}{doi: \begingroup \urlstyle{rm}\Url}\fi

\bibitem[Alldieck et~al.(2018)Alldieck, Magnor, Xu, Theobalt, and Pons-Moll]{Alldieck2018PeopleSnapshot}
Thiemo Alldieck, Marcus Magnor, Weipeng Xu, Christian Theobalt, and Gerard Pons-Moll.
\newblock Video based reconstruction of {3D} people models.
\newblock In \emph{CVPR}, pages 8387--8397, 2018.

\bibitem[Bengio et~al.(2009)Bengio, Louradour, Collobert, and Weston]{BengioLCW09_Curriculum}
Yoshua Bengio, J{\'e}r{\^o}me Louradour, Ronan Collobert, and Jason Weston.
\newblock Curriculum learning.
\newblock In \emph{ICML}, pages 41--48, 2009.

\bibitem[Beyer et~al.(2024)Beyer, Steiner, Pinto, Kolesnikov, Wang, Salz, et~al.]{Beyer2024PaliGemma}
Lucas Beyer, Andreas Steiner, Andr{\'e}~Susano Pinto, Alexander Kolesnikov, Xiao Wang, Daniel Salz, et~al.
\newblock {PaliGemma}: A versatile {3B} {VLM} for transfer.
\newblock arXiv:2407.07726, 2024.

\bibitem[Bi{\'n}kowski et~al.(2018)Bi{\'n}kowski, Sutherland, Arbel, and Gretton]{Binkowski2018KID}
Miko{\l}aj Bi{\'n}kowski, Danica~J. Sutherland, Michael Arbel, and Arthur Gretton.
\newblock Demystifying {MMD} {GANs}.
\newblock In \emph{ICLR}, 2018.

\bibitem[{Black Forest Labs}(2026{\natexlab{a}})]{Flux2Klein4B}
{Black Forest Labs}.
\newblock {FLUX.2} [klein] {4B}.
\newblock \url{https://huggingface.co/black-forest-labs/FLUX.2-klein-4B}, 2026{\natexlab{a}}.

\bibitem[{Black Forest Labs}(2026{\natexlab{b}})]{Flux2Klein9B}
{Black Forest Labs}.
\newblock {FLUX.2} [klein] {9B}.
\newblock \url{https://huggingface.co/black-forest-labs/FLUX.2-klein-9B}, 2026{\natexlab{b}}.

\bibitem[Carion et~al.(2020)Carion, Massa, Synnaeve, Usunier, Kirillov, and Zagoruyko]{Carion2020DETR}
Nicolas Carion, Francisco Massa, Gabriel Synnaeve, Nicolas Usunier, Alexander Kirillov, and Sergey Zagoruyko.
\newblock End-to-end object detection with transformers.
\newblock In \emph{ECCV}, pages 213--229, 2020.

\bibitem[Chen et~al.(2017)Chen, Papandreou, Schroff, and Adam]{Chen2017DeepLab}
Liang-Chieh Chen, George Papandreou, Florian Schroff, and Hartwig Adam.
\newblock Rethinking atrous convolution for semantic image segmentation.
\newblock arXiv:1706.05587, 2017.

\bibitem[Cheng et~al.(2022)Cheng, Misra, Schwing, Kirillov, and Girdhar]{Cheng2022Mask2Former}
Bowen Cheng, Ishan Misra, Alexander~G. Schwing, Alexander Kirillov, and Rohit Girdhar.
\newblock Masked-attention mask transformer for universal image segmentation.
\newblock In \emph{CVPR}, pages 1290--1299, 2022.

\bibitem[Chia et~al.(2022)Chia, Attanasio, Bianchi, Terragni, Magalh{\~a}es, Goncalves, Greco, and Tagliabue]{Chia2022FashionCLIP}
Patrick~John Chia, Giuseppe Attanasio, Federico Bianchi, Silvia Terragni, Ana~Rita Magalh{\~a}es, Diogo Goncalves, Ciro Greco, and Jacopo Tagliabue.
\newblock Contrastive language and vision learning of general fashion concepts.
\newblock \emph{Scientific Reports}, 12:\penalty0 18958, 2022.

\bibitem[Choi et~al.(2021)Choi, Park, Lee, and Choo]{Choi2021VITONHD}
Seunghwan Choi, Sunghyun Park, Minsoo Lee, and Jaegul Choo.
\newblock {VITON-HD}: High-resolution virtual try-on via misalignment-aware normalization.
\newblock In \emph{CVPR}, pages 14131--14140, 2021.

\bibitem[Choi et~al.(2024)Choi, Kwak, Lee, Choi, and Shin]{Choi2024IDMVTON}
Yisol Choi, Sangkyung Kwak, Kyungmin Lee, Hyungwon Choi, and Jinwoo Shin.
\newblock Improving diffusion models for authentic virtual try-on in the wild.
\newblock In \emph{ECCV}, 2024.

\bibitem[Chong et~al.(2025)Chong, Dong, Li, Zhang, Zhang, Zhang, Zhao, Jiang, and Liang]{Chong2025CatVTON}
Zheng Chong, Xiao Dong, Haoxiang Li, Shiyue Zhang, Wenqing Zhang, Xujie Zhang, Hanqing Zhao, Dongmei Jiang, and Xiaodan Liang.
\newblock {CatVTON}: Concatenation is all you need for virtual try-on with diffusion models.
\newblock In \emph{ICLR}, 2025.

\bibitem[Cui et~al.(2025)Cui, Mahajan, Shah, Gomathinayagam, Liu, and Lazebnik]{Yang2023StreetTryOn}
Aiyu Cui, Jay Mahajan, Viraj Shah, Preeti Gomathinayagam, Chang Liu, and Svetlana Lazebnik.
\newblock Street {TryOn}: Learning in-the-wild virtual try-on from unpaired person images.
\newblock In \emph{WACV}, 2025.

\bibitem[Dong et~al.(2019)Dong, Liang, Shen, Wang, Lai, Zhu, Hu, and Yin]{Guo2023MPV}
Haoye Dong, Xiaodan Liang, Xiaohui Shen, Bochao Wang, Hanjiang Lai, Jia Zhu, Zhiting Hu, and Jian Yin.
\newblock Towards multi-pose guided virtual try-on network.
\newblock In \emph{ICCV}, pages 9026--9035, 2019.

\bibitem[Dosovitskiy et~al.(2021)Dosovitskiy, Beyer, Kolesnikov, Weissenborn, Zhai, Unterthiner, Dehghani, Minderer, Heigold, Gelly, Uszkoreit, and Houlsby]{Dosovitskiy2021ViT}
Alexey Dosovitskiy, Lucas Beyer, Alexander Kolesnikov, Dirk Weissenborn, Xiaohua Zhai, Thomas Unterthiner, Mostafa Dehghani, Matthias Minderer, Georg Heigold, Sylvain Gelly, Jakob Uszkoreit, and Neil Houlsby.
\newblock An image is worth 16x16 words: Transformers for image recognition at scale.
\newblock In \emph{ICLR}, 2021.

\bibitem[Ge et~al.(2021{\natexlab{a}})Ge, Song, Ge, Yang, Liu, and Luo]{Fele2021DCTON}
Chongjian Ge, Yibing Song, Yuying Ge, Han Yang, Wei Liu, and Ping Luo.
\newblock Disentangled cycle consistency for highly-realistic virtual try-on.
\newblock In \emph{CVPR}, pages 16928--16937, 2021{\natexlab{a}}.

\bibitem[Ge et~al.(2021{\natexlab{b}})Ge, Song, Zhang, Ge, Liu, and Luo]{Ge2021PFAFN}
Yuying Ge, Yibing Song, Ruimao Zhang, Chongjian Ge, Wei Liu, and Ping Luo.
\newblock Parser-free virtual try-on via distilling appearance flows.
\newblock In \emph{CVPR}, pages 8485--8493, 2021{\natexlab{b}}.

\bibitem[Ghorbani et~al.(2021)Ghorbani, Mahdaviani, Thaler, Kording, Cook, Blohm, and Troje]{Ghorbani2021MoVi}
Saeed Ghorbani, Kimia Mahdaviani, Anne Thaler, Konrad Kording, Douglas~James Cook, Gunnar Blohm, and Nikolaus~F. Troje.
\newblock {MoVi}: A large multi-purpose human motion and video dataset.
\newblock \emph{PLOS ONE}, 16\penalty0 (6):\penalty0 e0253157, 2021.

\bibitem[Goel et~al.(2023)Goel, Pavlakos, Rajasegaran, Kanazawa, and Malik]{Goel2023HMR2}
Shubham Goel, Georgios Pavlakos, Jathushan Rajasegaran, Angjoo Kanazawa, and Jitendra Malik.
\newblock Humans in 4{D}: Reconstructing and tracking humans with transformers.
\newblock In \emph{ICCV}, 2023.

\bibitem[{Google DeepMind}(2025)]{GoogleNanoBananaPro}
{Google DeepMind}.
\newblock Introducing {Nano Banana Pro} ({Gemini 3 Pro Image}).
\newblock \url{https://blog.google/technology/ai/nano-banana-pro/}, 2025.

\bibitem[Gou et~al.(2023)Gou, Sun, Zhang, Si, Qian, and Zhang]{Gou2023DCIVTON}
Junhong Gou, Siyu Sun, Jianfu Zhang, Jianlou Si, Chen Qian, and Liqing Zhang.
\newblock Taming the power of diffusion models for high-quality virtual try-on with appearance flow.
\newblock In \emph{ACM MM}, pages 4304--4312. ACM, 2023.

\bibitem[Han et~al.(2018)Han, Wu, Wu, Yu, and Davis]{Han2018VITON}
Xintong Han, Zuxuan Wu, Zhe Wu, Ruichi Yu, and Larry~S. Davis.
\newblock {VITON}: An image-based virtual try-on network.
\newblock In \emph{CVPR}, pages 7543--7552, 2018.

\bibitem[Han et~al.(2019)Han, Hu, Huang, and Scott]{Han2019ClothFlow}
Xintong Han, Xiaojun Hu, Weilin Huang, and Matthew~R. Scott.
\newblock {ClothFlow}: A flow-based model for clothed person generation.
\newblock In \emph{ICCV}, pages 10471--10480, 2019.

\bibitem[Heusel et~al.(2017)Heusel, Ramsauer, Unterthiner, Nessler, and Hochreiter]{Heusel2017FID}
Martin Heusel, Hubert Ramsauer, Thomas Unterthiner, Bernhard Nessler, and Sepp Hochreiter.
\newblock {GANs} trained by a two time-scale update rule converge to a local {Nash} equilibrium.
\newblock In \emph{NeurIPS}, 2017.

\bibitem[Kim et~al.(2024)Kim, Gu, Park, Park, and Choo]{Kim2024StableVITON}
Jeongho Kim, Gyojung Gu, Minho Park, Sunghyun Park, and Jaegul Choo.
\newblock {StableVITON}: Learning semantic correspondence with latent diffusion model for virtual try-on.
\newblock In \emph{CVPR}, pages 8176--8185, 2024.

\bibitem[Li et~al.(2026)Li, Chen, Wen, Yin, Jiang, Wang, Luo, and Wu]{li2026openvtonbenchlargescalehighresolutionbenchmark}
Jin Li, Tao Chen, Kai Wen, Siqi Yin, Shuai Jiang, Weijie Wang, Jingwen Luo, and Chenhui Wu.
\newblock {OpenVTON-Bench}: A large-scale high-resolution benchmark for controllable virtual try-on evaluation.
\newblock In \emph{NeurIPS}, 2026.

\bibitem[Li et~al.(2025)Li, Shih, and Plummer]{LiSP25_SyntheticPairsVTON}
Nannan Li, Kevin~J. Shih, and Bryan~A. Plummer.
\newblock Enhancing virtual try-on with synthetic pairs and error-aware noise scheduling.
\newblock In \emph{CVPR}, pages 21238--21247, 2025.

\bibitem[Liu et~al.(2016)Liu, Luo, Qiu, Wang, and Tang]{Liu2016DeepFashion}
Ziwei Liu, Ping Luo, Shi Qiu, Xiaogang Wang, and Xiaoou Tang.
\newblock {DeepFashion}: Powering robust clothes recognition and retrieval with rich annotations.
\newblock In \emph{CVPR}, pages 1096--1104, 2016.

\bibitem[Lobba et~al.(2026)Lobba, Sanguigni, Ren, Cornia, Cucchiara, and Sebe]{dress_ed}
Davide Lobba, Fulvio Sanguigni, Bin Ren, Marcella Cornia, Rita Cucchiara, and Nicu Sebe.
\newblock {Dress-ED}: Instruction-guided editing for virtual try-on and try-off.
\newblock In \emph{ECCV}, 2026.

\bibitem[Loper et~al.(2015)Loper, Mahmood, Romero, Pons-Moll, and Black]{Loper2015SMPL}
Matthew Loper, Naureen Mahmood, Javier Romero, Gerard Pons-Moll, and Michael~J. Black.
\newblock {SMPL}: A skinned multi-person linear model.
\newblock \emph{ACM Transactions on Graphics (SIGGRAPH Asia)}, 34\penalty0 (6):\penalty0 248:1--248:16, 2015.

\bibitem[Minar et~al.(2020)Minar, Tuan, Ahn, Rosin, and Lai]{Minar2020CPVTONPLUS}
Matiur~Rahman Minar, Thai~Thanh Tuan, Heejune Ahn, Paul Rosin, and Yu-Kun Lai.
\newblock {CP-VTON+}: Clothing shape and texture preserving image-based virtual try-on.
\newblock In \emph{CVPRW}, pages 10--14, 2020.

\bibitem[{MMPose Contributors}(2020)]{mmpose2020}
{MMPose Contributors}.
\newblock Openmmlab pose estimation toolbox and benchmark.
\newblock \url{https://github.com/open-mmlab/mmpose}, 2020.

\bibitem[Morelli et~al.(2022)Morelli, Fincato, Cornia, Landi, Cesari, and Cucchiara]{Morelli2022DressCode}
Davide Morelli, Matteo Fincato, Marcella Cornia, Federico Landi, Fabio Cesari, and Rita Cucchiara.
\newblock {Dress Code}: High-resolution multi-category virtual try-on.
\newblock In \emph{ECCV}, 2022.

\bibitem[Morelli et~al.(2023)Morelli, Baldrati, Cartella, Cornia, Bertini, and Cucchiara]{Morelli2023LaDIVTON}
Davide Morelli, Alberto Baldrati, Giuseppe Cartella, Marcella Cornia, Marco Bertini, and Rita Cucchiara.
\newblock {LaDI-VTON}: Latent diffusion textual-inversion enhanced virtual try-on.
\newblock In \emph{ACM MM}, pages 8580--8589. ACM, 2023.

\bibitem[Oquab et~al.(2023)Oquab, Darcet, Moutakanni, Vo, Szafraniec, Khalidov, Fernandez, Haziza, Massa, El-Nouby, Assran, Ballas, Galuba, Howes, Huang, Li, Misra, Rabbat, Sharma, Synnaeve, Xu, J{\'e}gou, Mairal, Labatut, Joulin, and Bojanowski]{Oquab2023DINOv2}
Maxime Oquab, Timoth{\'e}e Darcet, Th{\'e}o Moutakanni, Huy~V. Vo, Marc Szafraniec, Vasil Khalidov, Pierre Fernandez, Daniel Haziza, Francisco Massa, Alaaeldin El-Nouby, Mahmoud Assran, Nicolas Ballas, Wojciech Galuba, Russell Howes, Po-Yao Huang, Shang-Wen Li, Ishan Misra, Michael Rabbat, Vasu Sharma, Gabriel Synnaeve, Hu Xu, Herv{\'e} J{\'e}gou, Julien Mairal, Patrick Labatut, Armand Joulin, and Piotr Bojanowski.
\newblock {DINOv2}: Learning robust visual features without supervision.
\newblock \emph{Transactions on Machine Learning Research (TMLR)}, 2023.

\bibitem[Qian et~al.(2025)Qian, Bocek-Rivele, Song, Tong, Yang, Lu, Hu, and Gan]{Qian2025PicoBanana}
Yusu Qian, Eli Bocek-Rivele, Liangchen Song, Jialing Tong, Yinfei Yang, Jiasen Lu, Wenze Hu, and Zhe Gan.
\newblock {Pico-Banana-400K}: A large-scale dataset for text-guided image editing.
\newblock \emph{arXiv preprint arXiv:2510.19808}, 2025.

\bibitem[{Qwen Team}(2025)]{Qwen3VL}
{Qwen Team}.
\newblock {Qwen3-VL} technical report.
\newblock arXiv:2511.21631, 2025.

\bibitem[Radford et~al.(2021)Radford, Kim, Hallacy, Ramesh, Goh, Agarwal, et~al.]{Radford2021CLIP}
Alec Radford, Jong~Wook Kim, Chris Hallacy, Aditya Ramesh, Gabriel Goh, Sandhini Agarwal, et~al.
\newblock Learning transferable visual models from natural language supervision.
\newblock In \emph{ICML}, 2021.

\bibitem[Richter et~al.(2016)Richter, Vineet, Roth, and Koltun]{RichterVRK16_PlayingForData}
Stephan~R. Richter, Vibhav Vineet, Stefan Roth, and Vladlen Koltun.
\newblock Playing for data: Ground truth from computer games.
\newblock In \emph{ECCV}, pages 102--118. Springer, 2016.

\bibitem[Ros et~al.(2016)Ros, Sellart, Materzynska, V{\'a}zquez, and L{\'o}pez]{RosSMVL16_Synthia}
Germ{\'a}n Ros, Laura Sellart, Joanna Materzynska, David V{\'a}zquez, and Antonio~M. L{\'o}pez.
\newblock The {SYNTHIA} dataset: A large collection of synthetic images for semantic segmentation of urban scenes.
\newblock In \emph{CVPR}, pages 3234--3243, 2016.

\bibitem[van~der Maaten and Hinton(2008)]{vanderMaaten2008tSNE}
Laurens van~der Maaten and Geoffrey Hinton.
\newblock Visualizing data using {t-SNE}.
\newblock \emph{Journal of Machine Learning Research}, 9:\penalty0 2579--2605, 2008.

\bibitem[Wang et~al.(2018)Wang, Zheng, Liang, Chen, Lin, and Yang]{Wang2018CPVTON}
Bochao Wang, Huabin Zheng, Xiaodan Liang, Yimin Chen, Liang Lin, and Meng Yang.
\newblock Toward characteristic-preserving image-based virtual try-on network.
\newblock In \emph{ECCV}, pages 589--604. Springer, 2018.

\bibitem[Wang et~al.(2025)Wang, Zhang, Di, Zhang, and Zuo]{Chen2024MVVTON}
Haoyu Wang, Zhilu Zhang, Donglin Di, Shiliang Zhang, and Wangmeng Zuo.
\newblock {MV-VTON}: Multi-view virtual try-on with diffusion models.
\newblock In \emph{AAAI}, 2025.

\bibitem[Wang et~al.(2004)Wang, Bovik, Sheikh, and Simoncelli]{Wang2004SSIM}
Zhou Wang, Alan~C. Bovik, Hamid~R. Sheikh, and Eero~P. Simoncelli.
\newblock Image quality assessment: From error visibility to structural similarity.
\newblock \emph{IEEE TIP}, 13\penalty0 (4):\penalty0 600--612, 2004.

\bibitem[Wu et~al.(2025)Wu, Li, Zhou, Lin, Gao, et~al.]{Wu2025QwenImage}
Chenfei Wu, Jiahao Li, Jingren Zhou, Junyang Lin, Kaiyuan Gao, et~al.
\newblock {Qwen-Image} technical report.
\newblock arXiv:2508.02324, 2025.

\bibitem[Xu et~al.(2022)Xu, Zhang, Zhang, and Tao]{Xu2022ViTPose}
Yufei Xu, Jing Zhang, Qiming Zhang, and Dacheng Tao.
\newblock Vitpose: Simple vision transformer baselines for human pose estimation.
\newblock In \emph{NeurIPS}, 2022.

\bibitem[Xu et~al.(2025)Xu, Gu, Chen, and Chen]{Xu2024OOTDiffusion}
Yuhao Xu, Tao Gu, Weifeng Chen, and Chengcai Chen.
\newblock {OOTDiffusion}: Outfitting fusion based latent diffusion for controllable virtual try-on.
\newblock In \emph{AAAI}, 2025.

\bibitem[Yang et~al.(2020)Yang, Zhang, Guo, Liu, Zuo, and Luo]{Yang2020ACGPN}
Han Yang, Ruimao Zhang, Xiaobao Guo, Wei Liu, Wangmeng Zuo, and Ping Luo.
\newblock Towards photo-realistic virtual try-on by adaptively generating-preserving image content.
\newblock In \emph{CVPR}, pages 7850--7859, 2020.

\bibitem[Zhang et~al.(2018)Zhang, Isola, Efros, Shechtman, and Wang]{Zhang2018LPIPS}
Richard Zhang, Phillip Isola, Alexei~A. Efros, Eli Shechtman, and Oliver Wang.
\newblock The unreasonable effectiveness of deep features as a perceptual metric.
\newblock In \emph{CVPR}, pages 586--595, 2018.

\bibitem[Zhang et~al.(2025)Zhang, Song, Zhan, Chang, Zeng, Chen, Luo, and Liu]{WildVTON}
Xuanpu Zhang, Dan Song, Pengxin Zhan, Tianyu Chang, Jianhao Zeng, Qingguo Chen, Weihua Luo, and An-An Liu.
\newblock {BooW-VTON}: Boosting in-the-wild virtual try-on via mask-free pseudo data training.
\newblock In \emph{CVPR}, pages 26399--26408, 2025.

\end{thebibliography}
}

\clearpage

\setcounter{topnumber}{3}
\setcounter{totalnumber}{5}
\renewcommand{\topfraction}{0.9}
\renewcommand{\bottomfraction}{0.8}
\renewcommand{\textfraction}{0.07}
\renewcommand{\floatpagefraction}{0.8}
\makeatletter
\setlength{\@fptop}{0pt}
\setlength{\@fpsep}{10pt plus 4pt}
\setlength{\@fpbot}{0pt plus 1fil}
\makeatother

\begin{center}
  {\Large\bf CURVTON-205K: A Mask-Free Difficulty-Stratified Synthetic Dataset
  for Robust and Scalable Virtual Try-On\par}
  \vspace{0.5em}
  {\large Supplementary Material\par}
  \vspace{1.0em}
\end{center}

\appendix

\section{Additional Curation Details}
\label{sec:additional_curation}

This section provides additional details on the automated and human-assisted
curation stages used to construct CURVTON-205K. We report how the validation
loop reduces artifact rates across difficulty tiers before the final
quality-screening step.

\subsection{VLM-Based Validation}
\label{sec:vlm_validation}

The VLM-based validation loop evaluates each candidate try-on triplet according
to garment faithfulness, semantic consistency, placement plausibility, and
visible artifacts. Samples flagged as inconsistent are regenerated for up to
four refinement attempts. Accordingly, the iterative results below include
the initial generation followed by successive refinement rounds.


\begin{table*}[!htbp]
\centering
\caption{
CURVTON synthetic try-on generation process and feedback iterations.
Artifact rates are reported by split and difficulty tier after each stage of
the pipeline, with VLM Thr denoting the acceptance threshold of the
VLM-based assessor in the closed-loop validation stage. This is distinct from
the ViT quality-filter threshold $\tau{=}0.85$ applied in the final automated
screening step (Section~\ref{sec:threshold_choice}). The final curated dataset contains 205K training triplets and 4.5K
test triplets. The test set is additionally manually verified by human
annotators to remove visually apparent artifacts and improve the reliability
of the final evaluation set across all difficulty tiers. Bold: best; underline: second best.
}
\label{tab:filter_iterations}

\scriptsize
\begin{tabular}{l c c c c c c c}
\toprule

\multirow{2}{*}{\textbf{Stage}}
  & \multirow{2}{*}{\shortstack{\textbf{Iteration}\\\textbf{(VLM Thr\,=\,0.8)}}}
  & \multicolumn{3}{c}{\textbf{Train Artifacts (\%)}}
  & \multicolumn{3}{c}{\textbf{Test Artifacts (\%)}} \\

\cmidrule(lr){3-5}
\cmidrule(lr){6-8}

  & & \textbf{Easy} & \textbf{Medium} & \textbf{Hard}
      & \textbf{Easy} & \textbf{Medium} & \textbf{Hard} \\

\midrule

Person \& cloth generation
  & $\times$
  & 17 & 25 & 35
  & $\times$ & $\times$ & $\times$ \\

\cmidrule(lr){1-8}

Person/Garment VLM Filtering
  & $\times$
  & 3 & 4.4 & 5.7
  & $\times$ & $\times$ & $\times$ \\

\cmidrule(lr){1-8}

Try-On Generation
  & Iteration 0
  & 38 & 48 & 60
  & 39 & 49 & 61 \\

  & Iteration I
  & 25 & 32 & 40
  & 26 & 33 & 41 \\

  & Iteration II
  & 15 & 19 & 24
  & 16 & 20 & 25 \\

  & Iteration III
  & \underline{8} & \underline{10} & \underline{13}
  & \underline{8} & \underline{11} & \underline{14} \\

  & Iteration IV
  & \textbf{3} & \textbf{4} & \textbf{5}
  & \textbf{3} & \textbf{4} & \textbf{5} \\

\bottomrule
\end{tabular}
\end{table*}


\subsection{Choice of the Quality-Filter Threshold}
\label{sec:threshold_choice}

The automated screening stage retains a candidate triplet when the quality
filter assigns it a score $s_n \ge \tau$, and the main paper uses
$\tau{=}0.85$. This value is deliberately far above the $0.5$ decision
boundary that a binary classifier would ordinarily imply, for the following
reason.

Note that $\tau$ governs the \emph{ViT} quality filter only. The earlier
closed-loop VLM assessor uses its own, separate acceptance threshold of $0.8$,
reported as VLM Thr in Table~\ref{tab:filter_iterations}; the two thresholds
belong to different stages and are not interchangeable.

By the time a triplet reaches the filter it has already survived the
closed-loop VLM validation stage of
Section~\ref{sec:vlm_validation}, which regenerates inconsistent samples for up
to four attempts. That stage alone brings the artifact rate down substantially
(Table~\ref{tab:filter_iterations}), so the pool entering the filter is already
a reasonably clean baseline rather than raw generator output. Setting
$\tau{=}0.5$ would therefore add very little: most of the pool sits above it,
and the filter would mainly re-confirm decisions the VLM loop had already made.

The filter is instead used as a second, stricter pass whose purpose is to raise
the quality floor further rather than to perform the initial separation. A
threshold of $0.85$ keeps only samples the filter is confident about, removing
the residual borderline cases that the VLM loop accepted but that still contain
mild artifacts. Because the candidate pool ($N \ge 209{,}500$) exceeds the
target dataset size ($205{,}000$ training and $4{,}500$ test triplets), we can
afford this strictness: discarding borderline samples costs coverage we can
replace by generating more, whereas retaining them would permanently lower the
quality of the released data. Acceptance rates are monitored across difficulty
tiers, garment categories, and appearance attributes so that the stricter
threshold does not silently prune long-tail groups.


\subsection{Cross-Generator and Cross-Evaluator Robustness}
\label{sec:cross_generator_evaluator}

To assess whether the observed effectiveness of the iterative validation
procedure depends on a particular image generator or VLM evaluator, we conduct
a small-scale cross-generator and cross-evaluator robustness study. We evaluate
multiple generator--evaluator configurations and measure artifact rates before
iterative refinement (Iteration 0) and after four refinement rounds
(Iteration IV) across the Easy, Medium, and Hard difficulty tiers. This
analysis complements Table~\ref{tab:filter_iterations} by examining whether
the artifact-reduction trend persists beyond the primary
FLUX.2-klein-9B~\cite{Flux2Klein9B} and Qwen3-VL 32B~\cite{Qwen3VL} configuration.


\begin{table*}[!htbp]
\centering
\caption{
Cross-generator and cross-evaluator robustness analysis.
Artifact rates report Easy/Medium/Hard percentages from the small-scale
screening experiments. Generator configuration and estimated cost are shown
once for each generator, with evaluator results grouped within the
corresponding block. ``Image Cond $n$'' denotes the maximum supported image
conditioning context. PaliGemma evaluates only the generated output, whereas
Qwen3-VL evaluates the complete source-person, target-garment, and
generated-output triplet. Iteration 0 denotes the initially generated outputs,
while Iteration IV denotes outputs after four rounds of iterative
VLM-guided refinement. Bold: best; underline: second best.
}
\label{tab:cross_generator_evaluator}

\scriptsize
\begin{tabular}{l c c l c c c}
\toprule

\textbf{Generator}
&
\shortstack{\textbf{Est.}\\\textbf{Cost}}
&
\shortstack{\textbf{Generator Image}\\\textbf{Cond $n$}}
&
\textbf{Evaluator(s)}
&
\shortstack{\textbf{Evaluator Image}\\\textbf{Cond $n$}}
&
\shortstack{\textbf{Artifact Rate (\%)}\\
\textbf{Easy / Medium / Hard}\\
\textbf{Iteration 0}}
&
\shortstack{\textbf{Artifact Rate (\%)}\\
\textbf{Easy / Medium / Hard}\\
\textbf{Iteration IV}}
\\

\midrule

\multirow{3}{*}{FLUX.2-klein-9B}
&
\multirow{3}{*}{\$25k}
&
\multirow{3}{*}{1--4}
&
Qwen3-VL 32B
&
3
&
\underline{44} / \underline{54} / \underline{66}
&
\underline{8} / \underline{11} / 14
\\

&
&
&
\shortstack[l]{Qwen3-VL-30B-A3B\\Thinking}
&
3
&
\underline{44} / \underline{54} / \underline{66}
&
9 / 12 / \underline{13}
\\

&
&
&
PaliGemma
&
1
&
\underline{44} / \underline{54} / \underline{66}
&
13 / 16 / 20
\\

\midrule

\multirow{2}{*}{FLUX.2-klein-4B}
&
\multirow{2}{*}{\$18k}
&
\multirow{2}{*}{1--4}
&
Qwen3-VL 32B
&
3
&
55 / 63 / 74
&
15 / 18 / 22
\\

&
&
&
PaliGemma \textit{(optional)}
&
1
&
55 / 63 / 74
&
18 / 22 / 27
\\

\midrule

\multirow{2}{*}{\shortstack[l]{Qwen-Image-Edit-\\2509}}
&
\multirow{2}{*}{\$90k}
&
\multirow{2}{*}{1--3}
&
Qwen3-VL 32B
&
3
&
65 / 73 / 82
&
21 / 25 / 30
\\

&
&
&
PaliGemma
&
1
&
65 / 73 / 82
&
28 / 33 / 38
\\

\midrule

Nano Banana Pro
&
\$160k
&
3
&
Qwen3-VL 32B
&
3
&
\textbf{40} / \textbf{48} / \textbf{60}
&
\textbf{5} / \textbf{8} / \textbf{10}
\\

\bottomrule
\end{tabular}
\end{table*}

Across all tested generator--evaluator configurations, iterative refinement
substantially reduces artifact rates across the three difficulty tiers.
For FLUX.2-klein-9B, the reduction remains consistent when replacing the
primary Qwen3-VL 32B evaluator with Qwen3-VL-30B-A3B Thinking~\cite{Qwen3VL} or PaliGemma~\cite{Beyer2024PaliGemma},
although the absolute residual artifact rates vary across evaluators.
Similarly, substantial reductions are observed for FLUX.2-klein-4B~\cite{Flux2Klein4B},
Qwen-Image-Edit-2509~\cite{Wu2025QwenImage}, and Nano Banana Pro~\cite{GoogleNanoBananaPro}. These results indicate that the
observed artifact-reduction trend is not restricted to a single
generator--evaluator configuration. Because this analysis is conducted on a
small-scale screening subset, we interpret it as evidence of robustness of the
refinement trend rather than as a comprehensive ranking of generators or
evaluators.

\section{Experiments and Additional Results}
\label{app:additional_results}
This section complements the main experimental analysis with implementation details, extended benchmark tables, scalability analyses, and additional curriculum ablations. These results provide fuller quantitative support for the compact tables reported in the main paper.

\subsection{Experimental Details}\label{suppl:experimental_details}
We summarize the compute environment, implementation details, and training setup used for data generation and model evaluation.

\paragraph{Compute Environment.}
Initial experiments and prototype development were conducted on the Lightning.ai and Vast.ai cloud GPU platforms. The majority of our experimental work, including the complete data generation pipeline, was executed on Vast.ai infrastructure. Data generation alone required over 3{,}000 H200 GPU hours, and training and evaluation of all reported models required a further 700 H200 GPU hours.

\paragraph{Train/Test Split Disjointness.}
The $205{,}000$ training and $4{,}500$ test triplets form disjoint partitions of
the curated dataset,
$\mathcal{D}_{\text{train}} \cap \mathcal{D}_{\text{test}} = \emptyset$.
The split is applied once, after quality screening, by partitioning the retained
pool; no triplet is duplicated across the two sides, so no evaluated
$(I_p, I_c, I_t)$ triplet is ever seen during training. The test partition is
additionally verified by human annotators using the rubric of
Section~\ref{app:rubric}, so the evaluation set is both held out and manually
checked. All numbers reported on the CURVTON-205K benchmark are computed on this
held-out partition only.

\paragraph{Dataset Adaptation and Mask-Free Training.}

VITON-HD and DressCode follow a paired shop-to-model structure, providing a real person image wearing the target garment together with the corresponding in-shop garment, but without a separate source-person image required by our triplet formulation. As our evaluation protocol operates on complete triplets $(I_p, I_c, I_t)$, a structural mismatch arises when applying these benchmarks directly. To resolve this while maintaining consistency with the testing protocol established by BooW-VTON~\cite{WildVTON}, we complete each pair into a triplet by synthesizing the missing source-person image $I_p$ via our generation pipeline, while retaining the original benchmark person image as the real ground-truth try-on target $I_t$ and the corresponding in-shop garment as $I_c$. This construction preserves the original ground-truth person--garment correspondence while providing the complete triplet structure required by our evaluation protocol. Furthermore, all baseline architectures were trained under a unified mask-free regime. Concretely, the masked person representation, conventionally supplied as a zeroed or in-painted region indicating the garment placement area, was replaced throughout training with the original, unmasked person image $I_p$. This substitution aligns the conditioning signal of all baselines with the mask-free supervision structure of CURVTON-205K and prevents the introduction of mask-induced spatial artifacts as a confounding factor in cross-architecture comparisons.

The training experiments across the different model architectures were implemented using PyTorch and distributed across multiple GPU nodes. We employed mixed-precision training to optimize memory usage and training speed. Hyperparameters for each model architecture were tuned using a subset of the training data, with final models trained on the complete CURVTON-205K dataset. Curriculum-aware variants use the difficulty-tier scheduling protocols described in the main paper.

For the curriculum training experiments, we consider two strategies: no curriculum and standard curriculum. In the no-curriculum setting, samples from all difficulty tiers are mixed throughout training. In the standard-curriculum setting, harder samples are gradually introduced into the training batches, so that training begins with easier tiers and avoids abrupt transitions between difficulty levels.

\begin{table*}[t]
  \caption{Stratified performance of all models trained and evaluated on the CURVTON-205K benchmark test set. SSIM$\uparrow$~\cite{Wang2004SSIM} higher is better; LPIPS$\downarrow$~\cite{Zhang2018LPIPS}, FID$\downarrow$~\cite{Heusel2017FID} lower is better. \textbf{Top:} overall scores and results by difficulty tier (E\,=\,Easy, M\,=\,Med, H\,=\,Hard). \textbf{Bottom:} results by cloth type (UB\,=\,Upper Body, LB\,=\,Lower Body, G\,=\,Garment, Tr\,=\,Traditional, NTr\,=\,Non-Traditional). Both panels list the same models and rows. Each model reports two settings, \textit{Zero Shot} and \textit{CURVTON Training}, except generalist models which are evaluated in \textit{Zero Shot} only. Bold: best; underline: second best.}
  \label{tab:stratified_perf_suppl}
  \centering
  \scriptsize
  \setlength{\tabcolsep}{3pt}
  \renewcommand{\arraystretch}{1.05}

  \begin{tabular}{l l ccc ccccccccc}
    \toprule
      & & \multicolumn{3}{c}{Overall} & \multicolumn{9}{c}{Difficulty Tier} \\
    \cmidrule(lr){3-5}\cmidrule(lr){6-14}
      & & \multirow{2}{*}{SSIM$\uparrow$} & \multirow{2}{*}{LPIPS$\downarrow$} & \multirow{2}{*}{FID$\downarrow$}
      & \multicolumn{3}{c}{LPIPS$\downarrow$} & \multicolumn{3}{c}{SSIM$\uparrow$} & \multicolumn{3}{c}{FID$\downarrow$} \\
    \cmidrule(lr){6-8}\cmidrule(lr){9-11}\cmidrule(lr){12-14}
      & & & & & E & M & H & E & M & H & E & M & H \\
    \midrule
    \multirow{2}{*}{OOTDiffusion~\cite{Xu2024OOTDiffusion}} & Zero Shot & 0.75 & 0.19 & 18.90 & 0.18 & 0.18 & 0.22 & 0.78 & 0.77 & 0.70 & 25.20 & 24.40 & 25.50 \\
     & CURVTON Training & 0.85 & 0.13 & \underline{8.75} & \underline{0.12} & 0.12 & 0.15 & \underline{0.86} & 0.85 & \underline{0.83} & 9.88 & 11.54 & 14.12 \\
    \midrule \multirow{2}{*}{CatVTON~\cite{Chong2025CatVTON}} & Zero Shot & 0.45 & 0.36 & 23.85 & 0.30 & 0.36 & 0.41 & 0.55 & 0.41 & 0.37 & 22.55 & 24.40 & 27.55 \\
     & CURVTON Training & \underline{0.86} & \underline{0.12} & 9.14 & \underline{0.12} & \textbf{0.09} & \textbf{0.12} & \underline{0.86} & \underline{0.86} & 0.81 & \underline{9.14} & \underline{9.23} & \underline{12.88} \\
    \midrule \multirow{2}{*}{IDM-VTON~\cite{Choi2024IDMVTON}} & Zero Shot & 0.80 & 0.15 & 17.05 & 0.14 & 0.14 & 0.17 & 0.82 & 0.80 & 0.77 & 16.20 & 17.30 & 19.30 \\
     & CURVTON Training & \textbf{0.87} & \textbf{0.11} & \textbf{8.44} & \textbf{0.10} & \underline{0.10} & \textbf{0.12} & \textbf{0.88} & \textbf{0.87} & \textbf{0.84} & \textbf{8.28} & \textbf{8.85} & \textbf{10.29} \\
    \midrule \multirow{1}{*}{DCI-VTON~\cite{Gou2023DCIVTON}} & Zero Shot & 0.81 & 0.17 & 19.20 & 0.16 & 0.16 & 0.18 & 0.82 & 0.81 & 0.79 & 18.00 & 19.30 & 20.80 \\
    \midrule \multirow{1}{*}{FLUX.2 [klein] 9B (Generator)} & Zero Shot & 0.77 & 0.15 & 14.50 & 0.14 & 0.15 & 0.18 & 0.79 & 0.78 & 0.73 & 13.10 & 15.35 & 17.95 \\
    \bottomrule
  \end{tabular}

  \vspace{6pt}

  \begin{tabular}{l l ccccc ccccc ccccc}
    \toprule
      & & \multicolumn{15}{c}{Cloth Type} \\
    \cmidrule(lr){3-17}
      & & \multicolumn{5}{c}{LPIPS$\downarrow$} & \multicolumn{5}{c}{SSIM$\uparrow$} & \multicolumn{5}{c}{FID$\downarrow$} \\
    \cmidrule(lr){3-7}\cmidrule(lr){8-12}\cmidrule(lr){13-17}
      & & UB & LB & G & Tr & NTr & UB & LB & G & Tr & NTr & UB & LB & G & Tr & NTr \\
    \midrule
    \multirow{2}{*}{OOTDiffusion~\cite{Xu2024OOTDiffusion}} & Zero Shot & 0.16 & 0.20 & 0.21 & 0.22 & 0.16 & 0.78 & 0.75 & 0.72 & 0.73 & 0.81 & 19.50 & 27.15 & 29.70 & 29.95 & 17.15 \\
     & CURVTON Training & 0.12 & 0.11 & 0.15 & \underline{0.18} & 0.12 & 0.86 & 0.87 & 0.80 & 0.78 & \underline{0.88} & 11.42 & 11.83 & \underline{13.79} & \textbf{14.22} & 13.83 \\
    \midrule \multirow{2}{*}{CatVTON~\cite{Chong2025CatVTON}} & Zero Shot & 0.35 & 0.34 & 0.38 & 0.36 & 0.32 & 0.45 & 0.45 & 0.42 & 0.50 & 0.54 & 22.60 & 23.55 & 28.90 & 31.35 & 20.16 \\
     & CURVTON Training & \underline{0.11} & \underline{0.10} & \underline{0.14} & \textbf{0.16} & \textbf{0.10} & \underline{0.87} & \underline{0.88} & \textbf{0.82} & \textbf{0.80} & \textbf{0.90} & \underline{10.62} & \underline{11.34} & 14.94 & \underline{15.57} & \underline{12.70} \\
    \midrule \multirow{2}{*}{IDM-VTON~\cite{Choi2024IDMVTON}} & Zero Shot & 0.14 & 0.13 & 0.18 & 0.24 & 0.16 & 0.81 & 0.83 & 0.75 & 0.71 & 0.80 & 16.70 & 17.25 & 22.00 & 25.55 & 15.15 \\
     & CURVTON Training & \textbf{0.10} & \textbf{0.09} & \textbf{0.13} & \underline{0.18} & \textbf{0.10} & \textbf{0.88} & \textbf{0.89} & \textbf{0.82} & 0.77 & \underline{0.88} & \textbf{10.21} & \textbf{10.57} & \textbf{13.54} & 15.75 & \textbf{7.98} \\
    \midrule \multirow{1}{*}{DCI-VTON~\cite{Gou2023DCIVTON}} & Zero Shot & 0.18 & 0.14 & 0.19 & 0.19 & 0.25 & 0.80 & 0.84 & 0.77 & \underline{0.79} & 0.72 & 20.15 & 17.05 & 24.10 & 18.05 & 25.55 \\
    \midrule \multirow{1}{*}{FLUX.2 [klein] 9B (Generator)} & Zero Shot & 0.14 & 0.15 & 0.18 & 0.35 & 0.31 & 0.79 & 0.79 & 0.73 & 0.47 & 0.52 & 14.75 & 16.05 & 17.50 & 23.27 & 20.23 \\
    \bottomrule
  \end{tabular}
\end{table*}

\subsection{Embedding-Based Diversity Analysis}
\begin{figure*}[t]
  \centering
  \includegraphics[width=0.45\linewidth]{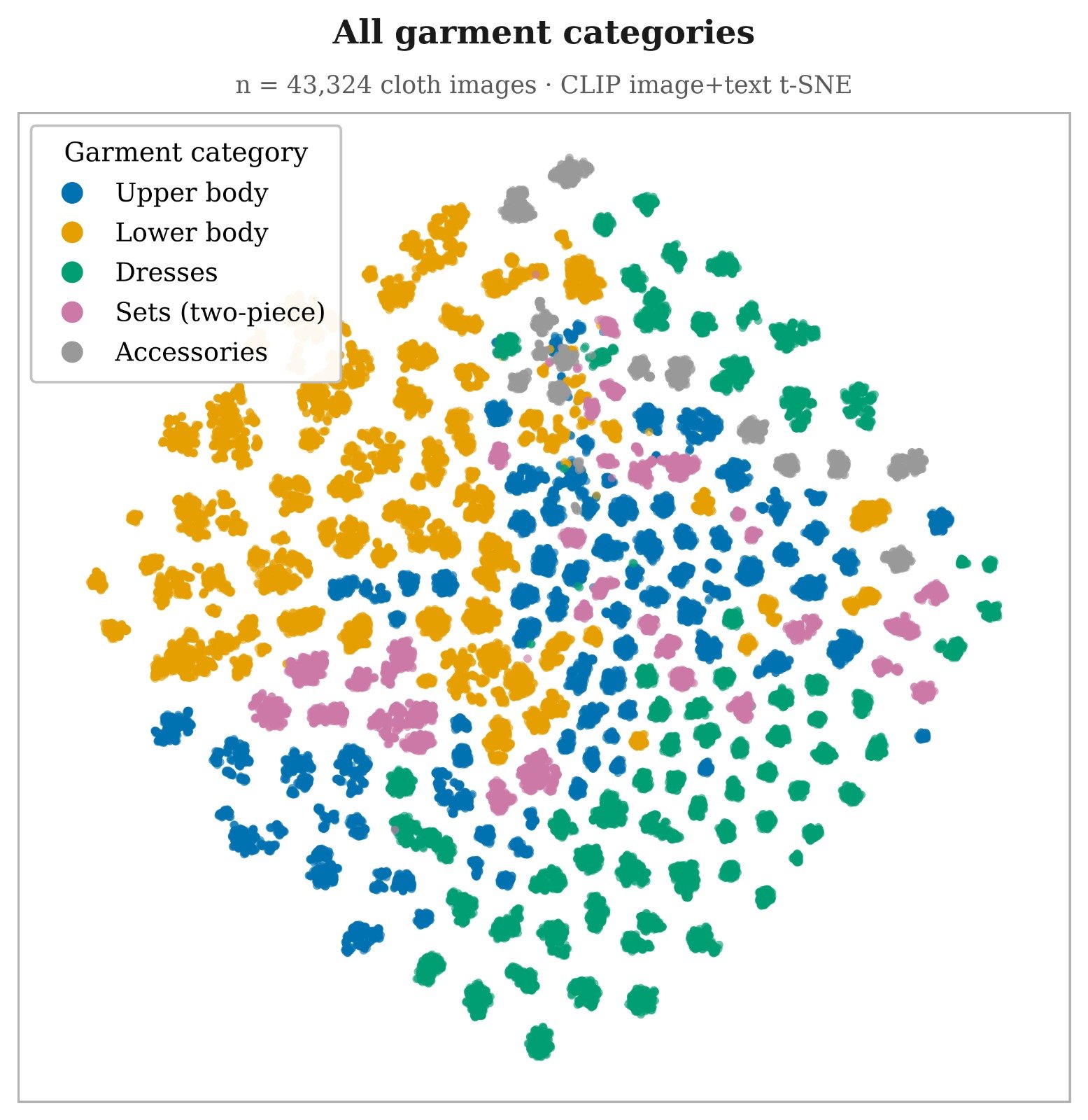}\hfill
  \includegraphics[width=0.45\linewidth]{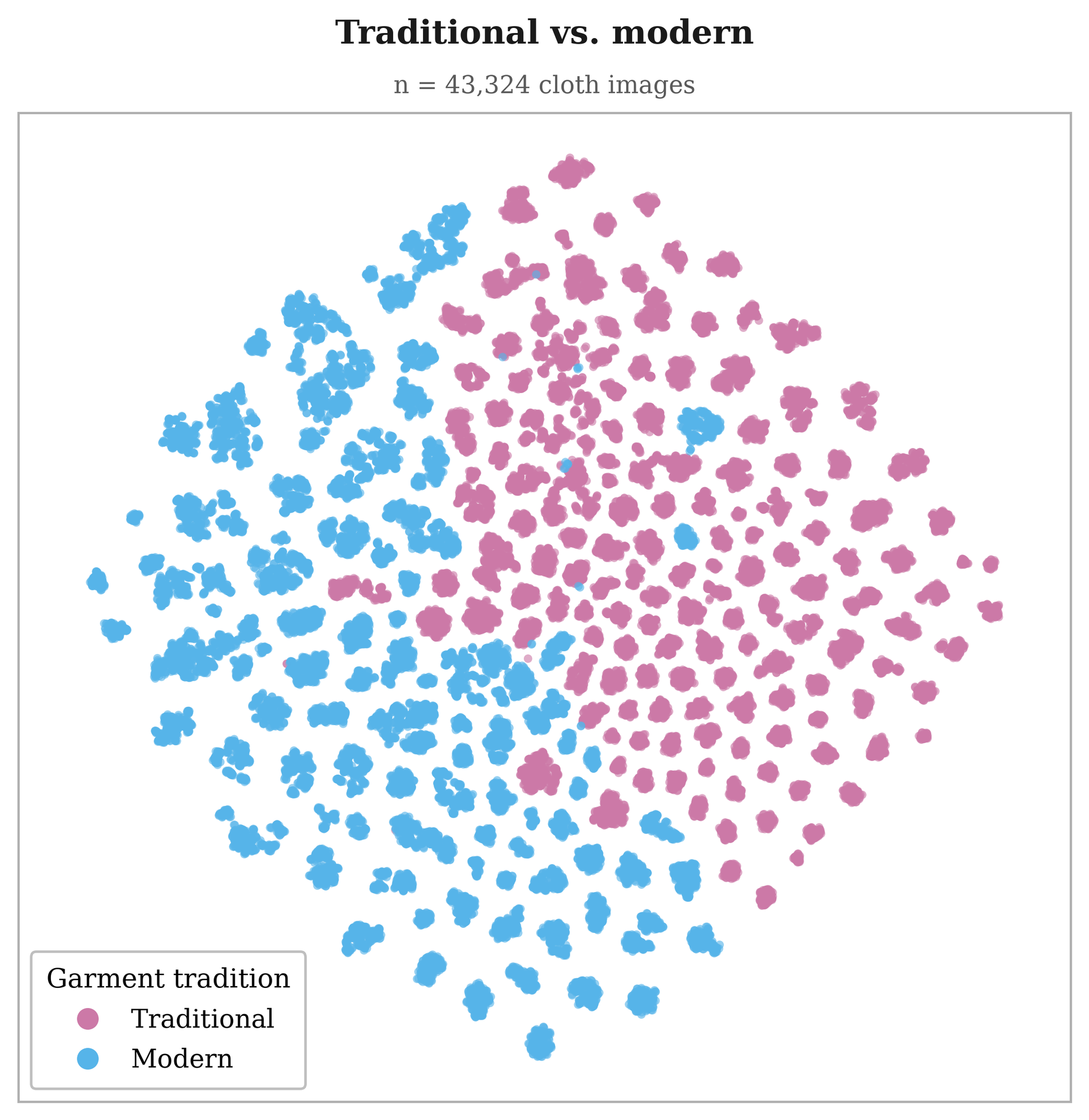}\\[4pt]
  \includegraphics[width=0.8\linewidth]{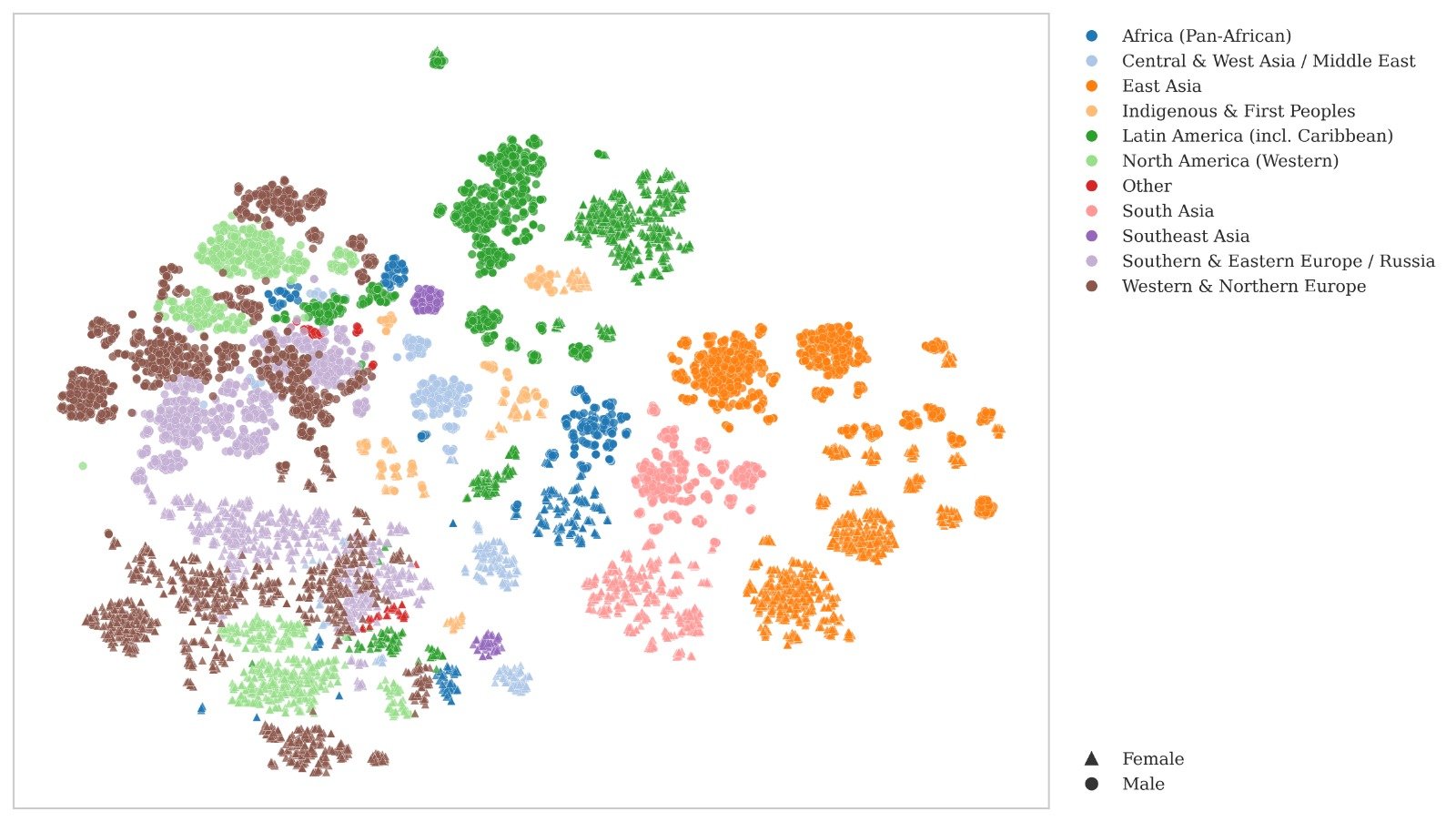}
  \caption{Diversity in CURVTON-205K for garments and cultural representation. We use 2D t-SNE projections of pretrained embeddings. Top left: CLIP image and text embeddings of the 43,324 garment images, colored by garment category. Top right: the same projection, colored by whether the garment is traditional or modern. Bottom: facial-appearance embeddings taken from face regions, colored by cultural group and marked by gender.}
  \label{fig:diversity_and_fairness}
\end{figure*}

Fig.~\ref{fig:diversity_and_fairness} provides additional qualitative diagnostics of dataset diversity using two-dimensional t-SNE~\cite{vanderMaaten2008tSNE} projections of pretrained garment and facial-appearance embeddings. These projections are intended as visual summaries of coverage rather than quantitative metrics; nevertheless, CURVTON-205K shows broader occupancy and less pronounced mode concentration across the embedding spaces.

\subsection{Full Curriculum Ablation Studies}
We report extended curriculum-aware ablations to separate the effect of difficulty-tier composition from the effect of sample ordering. These results complement the compact curriculum analysis in the main paper. Table~\ref{tab:curriculum_ablation_suppl} expands the tier-composition analysis by including single-tier, pairwise-tier, and scale-controlled variants. Table~\ref{tab:curriculum_results_suppl} expands the ordering analysis across training budgets and training regimes.

\begin{table*}[t]
    \caption{Extended difficulty-tier composition and scale ablation. Models are trained with different combinations of easy, medium, and hard samples, as well as different fractions of the full three-tier dataset, and evaluated across difficulty-specific test splits. Bold: best; underline: second best.}    
    \vspace{-4pt}
  \label{tab:curriculum_ablation_suppl}
  \centering
  \scriptsize
  \setlength{\tabcolsep}{1.3pt}
  \renewcommand{\arraystretch}{0.95}
  \begin{tabular}{l cccc cccc cccc cccc}
    \toprule
    & \multicolumn{4}{c}{Easy Test}
    & \multicolumn{4}{c}{Medium Test}
    & \multicolumn{4}{c}{Hard Test}
    & \multicolumn{4}{c}{All Mixed} \\
    \cmidrule(lr){2-5}\cmidrule(lr){6-9}\cmidrule(lr){10-13}\cmidrule(lr){14-17}
    Train Data
      & SSIM$\uparrow$ & LPIPS$\downarrow$ & FID$\downarrow$ & KID$\downarrow$
      & SSIM$\uparrow$ & LPIPS$\downarrow$ & FID$\downarrow$ & KID$\downarrow$
      & SSIM$\uparrow$ & LPIPS$\downarrow$ & FID$\downarrow$ & KID$\downarrow$
      & SSIM$\uparrow$ & LPIPS$\downarrow$ & FID$\downarrow$ & KID$\downarrow$ \\
    \midrule
    Easy only
      & \textbf{0.86} & \textbf{0.11} & \underline{9.86}  & \textbf{0.95}
      & 0.83 & 0.12 & 12.70 & 0.92
      & 0.76 & 0.18 & 19.87 & 2.29
      & \underline{0.85} & \textbf{0.11} & 10.29 & 0.94 \\
    Medium only
      & 0.82 & 0.18 & 12.89 & 1.75
      & 0.83 & 0.12 & 11.02 & 0.46
      & 0.78 & 0.15 & 15.32 & 0.61
      & 0.82 & 0.17 & 12.52 & 1.73 \\
    Hard only
      & 0.83 & 0.16 & 13.08 & 2.51
      & 0.84 & 0.12 & 11.47 & 1.11
      & 0.77 & 0.15 & 15.10 & 1.08
      & 0.83 & 0.15 & 13.15 & 2.75 \\
    Medium + Hard
      & 0.81 & 0.20 & 12.75 & 1.82
      & 0.82 & 0.13 & 11.11 & 0.39
      & 0.77 & 0.16 & 15.25 & 0.59
      & 0.80 & 0.16 & 13.92 & \textbf{0.64} \\
    Easy + Hard (full)
      & 0.83 & 0.16 & 11.46 & 1.20
      & 0.84 & 0.11 & 10.42 & 0.48
      & 0.75 & 0.17 & 15.50 & 0.95
      & 0.81 & 0.15 & 13.54 & \underline{0.70} \\
    \midrule
    Easy + Med + Hard (10\%)
      & 0.81 & 0.19 & 11.08 & 1.38
      & 0.82 & 0.14 & 10.35 & 0.30
      & 0.74 & 0.18 & 14.94 & 0.64
      & 0.77 & 0.21 & 12.02 & 2.54 \\
    Easy + Med + Hard (20\%)
      & 0.80 & 0.20 & 12.23 & 1.70
      & 0.81 & 0.14 & 10.91 & 0.47
      & 0.77 & 0.16 & 14.01 & \underline{0.30}
      & 0.80 & 0.19 & 11.99 & 1.79 \\
    Easy + Med + Hard (50\%)
      & 0.83 & 0.16 & 10.35 & 1.42
      & \underline{0.85} & \underline{0.10} & \underline{9.51}  & \underline{0.26}
      & \underline{0.80} & \underline{0.13} & \underline{13.15} & 0.37
      & 0.83 & 0.15 & \underline{10.18} & 1.23 \\
    Easy + Med + Hard (100\%)
      & \textbf{0.86} & \underline{0.12} & \textbf{9.14}  & \underline{0.96}
      & \textbf{0.86} & \textbf{0.09} & \textbf{9.23}  & \textbf{0.20}
      & \textbf{0.81} & \textbf{0.12} & \textbf{12.88} & \textbf{0.22}
      & \textbf{0.86} & \underline{0.12} & \textbf{9.14}  & 0.97 \\
    \bottomrule
  \end{tabular}
\end{table*}

\begin{table*}[t]
    \caption{Curriculum vs.\ non-curriculum under finetuning \textbf{(Stable Diffusion)} regime, comparing minimal \textbf{(3.6k iters)}, limited \textbf{(7.2k iters)},
    medium \textbf{(14.4k iters)}, and full \textbf{(28.8k iters)} training budgets across difficulty splits, including a mixed test set. Bold: best; underline: second best.}
    \vspace{-6pt}
    \label{tab:curriculum_results_suppl}
    \centering
    \scriptsize
    \setlength{\tabcolsep}{1.0pt}
    \renewcommand{\arraystretch}{1.0}
    \begin{tabular}{l l l ccc ccc ccc ccc}
        \toprule
        & & & \multicolumn{3}{c}{Easy}
          & \multicolumn{3}{c}{Medium}
          & \multicolumn{3}{c}{Hard}
          & \multicolumn{3}{c}{All Mixed} \\
        \cmidrule(lr){4-6}\cmidrule(lr){7-9}\cmidrule(lr){10-12}\cmidrule(lr){13-15}
        Regime & Budget & Curriculum
          & SSIM$\uparrow$ & LPIPS$\downarrow$ & FID$\downarrow$
          & SSIM$\uparrow$ & LPIPS$\downarrow$ & FID$\downarrow$
          & SSIM$\uparrow$ & LPIPS$\downarrow$ & FID$\downarrow$
          & SSIM$\uparrow$ & LPIPS$\downarrow$ & FID$\downarrow$ \\
        \midrule

        \multirow{8}{*}{\shortstack[l]{Finetuning \\ (Stable Diffusion)}}
        & \multirow{2}{*}{\shortstack[l]{Minimal \\ (3.6k iters)}}
          & No Curriculum
          & 0.78 & 0.17 & 21.4
          & 0.74 & 0.21 & 26.7
          & 0.66 & 0.27 & 36.3
          & 0.73 & 0.22 & 28.1 \\
        & & Standard Curriculum
          & 0.79 & 0.16 & 20.2
          & 0.77 & 0.19 & 25.3
          & 0.68 & 0.25 & 33.8
          & 0.75 & 0.21 & 26.4 \\

        \cmidrule(lr){2-15}

        & \multirow{2}{*}{\shortstack[l]{Limited \\ (7.2k iters)}}
          & No Curriculum
          & 0.81 & 0.14 & 19.3
          & 0.77 & 0.18 & 23.8
          & 0.70 & 0.23 & 32.7
          & 0.76 & 0.19 & 25.2 \\
        & & Standard Curriculum
          & 0.82 & 0.13 & 18.4
          & 0.78 & 0.17 & 22.4
          & 0.72 & 0.21 & 29.8
          & 0.77 & 0.17 & 23.5 \\

        \cmidrule(lr){2-15}

        & \multirow{2}{*}{\shortstack[l]{Medium \\ (14.4k iters)}}
          & No Curriculum
          & 0.83 & 0.13 & 17.9
          & 0.79 & 0.16 & 21.7
          & 0.73 & 0.21 & 28.4
          & 0.78 & 0.17 & 22.7 \\
        & & Standard Curriculum
          & 0.83 & \underline{0.12} & 17.3
          & 0.80 & \underline{0.15} & 21.1
          & 0.74 & 0.20 & 27.6
          & 0.79 & \underline{0.16} & 22.0 \\

        \cmidrule(lr){2-15}

        & \multirow{2}{*}{\shortstack[l]{Full \\ (28.8k iters)}}
          & No Curriculum
          & \underline{0.84} & \underline{0.12} & \underline{16.8}
          & \underline{0.81} & \underline{0.15} & \underline{20.3}
          & \underline{0.76} & \underline{0.19} & \underline{26.1}
          & \underline{0.80} & \underline{0.16} & \underline{21.1} \\
        & & Standard Curriculum
          & \textbf{0.85} & \textbf{0.11} & \textbf{16.2}
          & \textbf{0.82} & \textbf{0.14} & \textbf{19.4}
          & \textbf{0.77} & \textbf{0.18} & \textbf{25.3}
          & \textbf{0.81} & \textbf{0.15} & \textbf{20.3} \\
        \bottomrule
    \end{tabular}
\end{table*}

\section{Supplementary Qualitative Results}
\label{app:qualitative_results}


These sections include additional qualitative samples of the CURVTON-205K dataset, sorted by difficulty level, garment type, and pipeline components. In each triplet, there is the source person image on the left-hand side, the target garment from an in-store product on the middle, and the generated output of try-on result on the right-hand side. The samples below are carefully selected from the test split to showcase the diversity across the main dimensions of the dataset.

\subsection{Try-On Examples Across Difficulty Tiers}
\label{app:difficulty_tier_examples}

\begin{figure*}[!htbp]
  \centering
  \includegraphics[width=\textwidth]{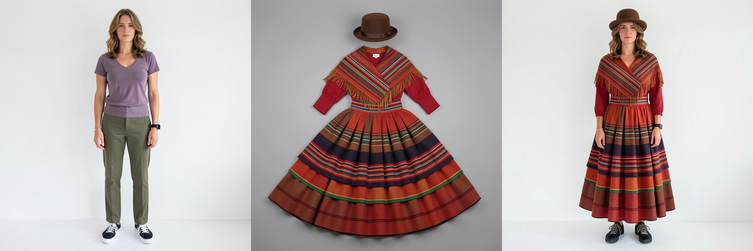}\\[4pt]
  \includegraphics[width=\textwidth]{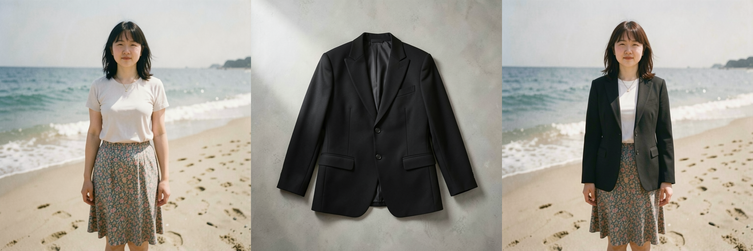}\\[4pt]
  \includegraphics[width=\textwidth]{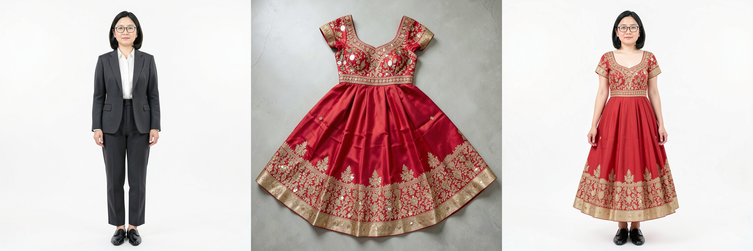}
  \caption{\textbf{Try-on Triplets of Easy tier.} Left: source person, center: target garment, right: synthesized try-on output. Samples of easy-tier try-ons show near front-facing poses, minimal occlusions and clean backgrounds, comprising traditional outfits like Cholita dress (row 1), Ghagra Choli (row 3), \`{A}o B\`{a} B\`{a} (row 4), as well as regular Western clothes.}
  \label{fig:app_easy}
\end{figure*}
\begin{figure*}[!htbp]
  \centering
  \includegraphics[width=\textwidth]{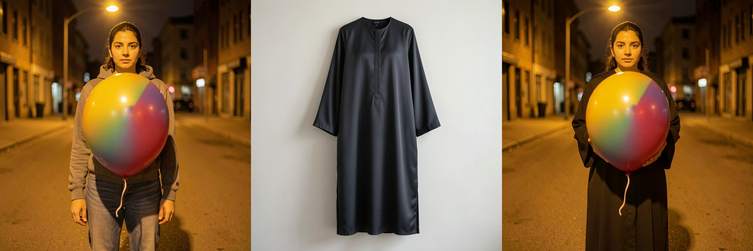}\\[4pt]
  \includegraphics[width=\textwidth]{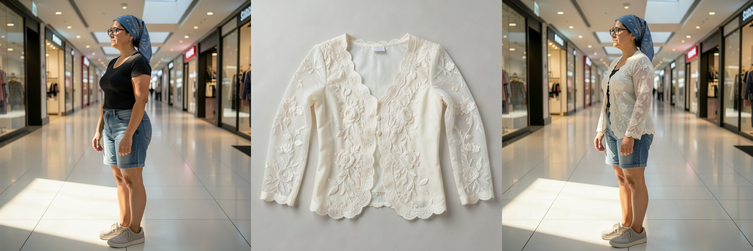}\\[4pt]
  \includegraphics[width=\textwidth]{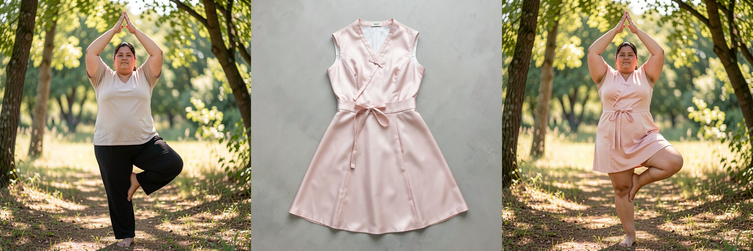}
  \caption{\textbf{Try-on triplets of Medium tier.} As compared to the Easy tier, the Medium-tier samples have moderate pose variations, with some occlusions and comparatively more complex backgrounds. Garment diversity includes a range of culturally-specific clothing from various cultures, namely, Abaya (row 1), Kebaya (row 2), Ruqun (row 3), Hanfu (row 4), and Balochi suit (row 6).}
  \label{fig:app_medium}
\end{figure*}
\begin{figure*}[!htbp]
  \centering
  \includegraphics[width=\textwidth]{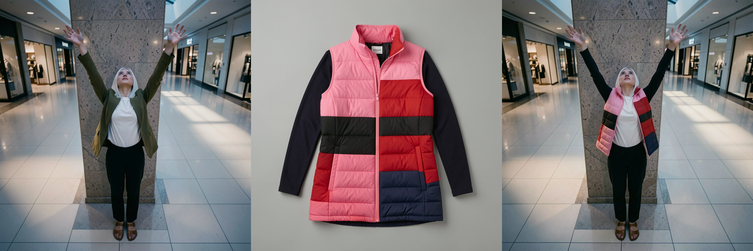}\\[4pt]
  \includegraphics[width=\textwidth]{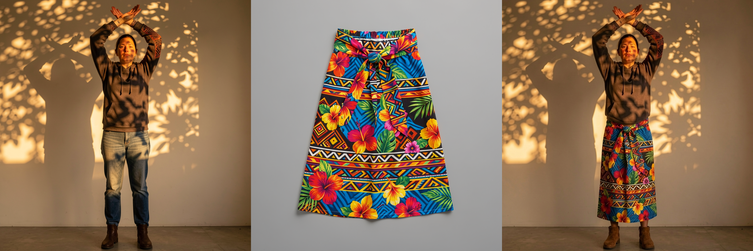}\\[4pt]
  \includegraphics[width=\textwidth]{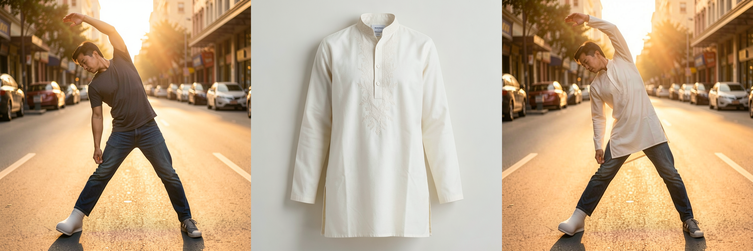}
  \caption{\textbf{Try-on triplets of Hard tier.} Samples of the Hard tier are characterized by extremely complex poses, significant occlusions and challenging in-the-wild environments, which involve long-tail traditional dresses, such as Lavalava (row 2), Panjabi (row 3) and Hakama (row 5). These cases stress both garment-identity preservation and geometric plausibility under strong visual complexity.}
  \label{fig:app_hard}
\end{figure*}

Figures~\ref{fig:app_easy}--\ref{fig:app_hard}, show try-on triplets sampled from easy, medium, and hard difficulty tiers, respectively. The difficulty tier rises gradually due to larger pose offsets, stronger occlusion effects, more cluttered backgrounds, and increasing garment-to-person interaction complexity. The easy difficulty tier corresponds to nearly frontal poses against either a studio background or a minimal background. The medium difficulty tier includes slight pose offset, moderate occlusions, and cluttered backgrounds, while the hard difficulty tier implies challenging poses, large occlusion rates, and culturally diverse clothing.

\begin{figure*}[!htbp]
  \centering
  \includegraphics[width=\textwidth]{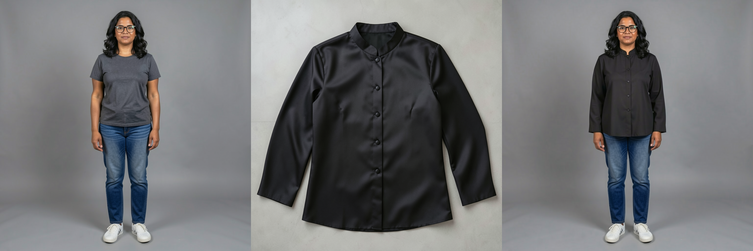}\\[4pt]
  \includegraphics[width=\textwidth]{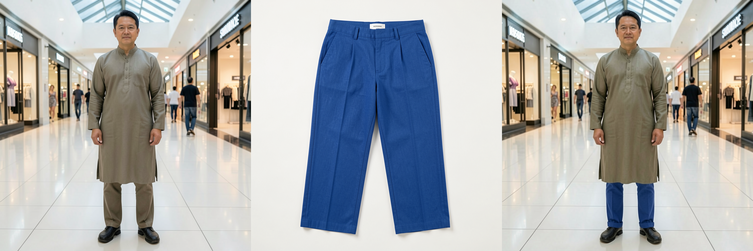}\\[4pt]
  \includegraphics[width=\textwidth]{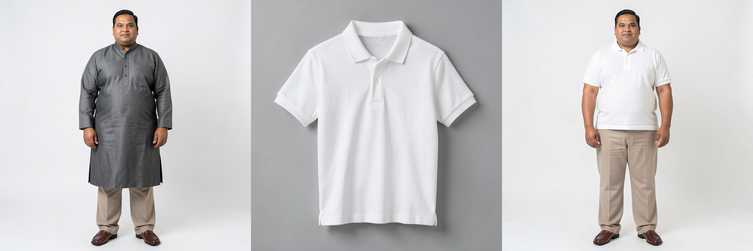}
  \contfig{fig:app_easy}
\end{figure*}

\begin{figure*}[!htbp]
  \centering
  \includegraphics[width=\textwidth]{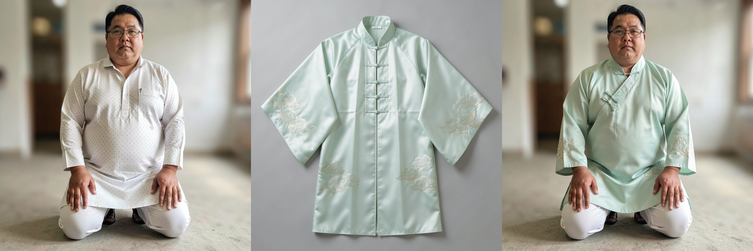}\\[4pt]
  \includegraphics[width=\textwidth]{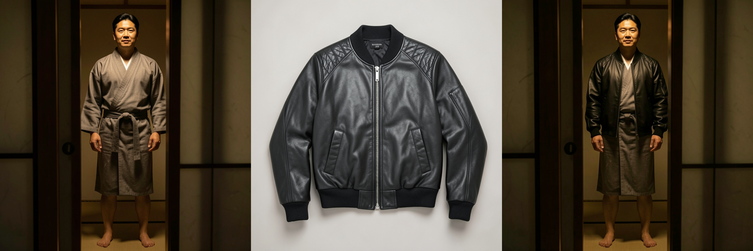}\\[4pt]
  \includegraphics[width=\textwidth]{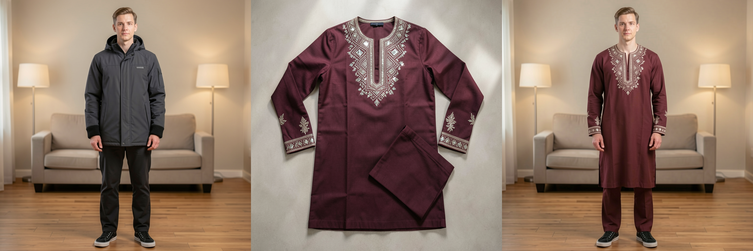}
  \contfig{fig:app_medium}
\end{figure*}

\begin{figure*}[!htbp]
  \centering
  \includegraphics[width=\textwidth]{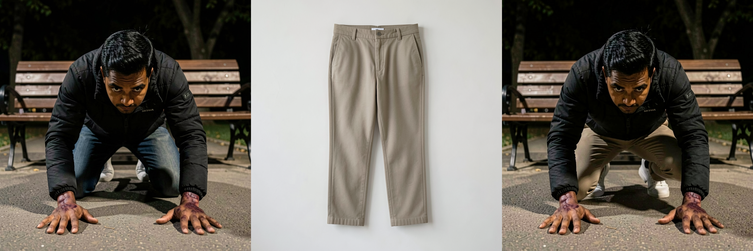}\\[4pt]
  \includegraphics[width=\textwidth]{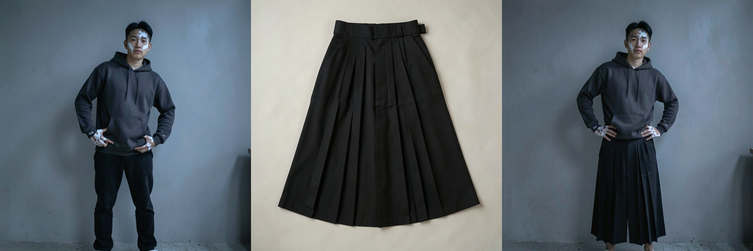}
  \contfig{fig:app_hard}
\end{figure*}

\subsection{Try-On Examples Across Garment Categories}
\label{app:garment_category_examples}

\begin{figure*}[!htbp]
  \centering
  \includegraphics[width=0.23\textwidth]{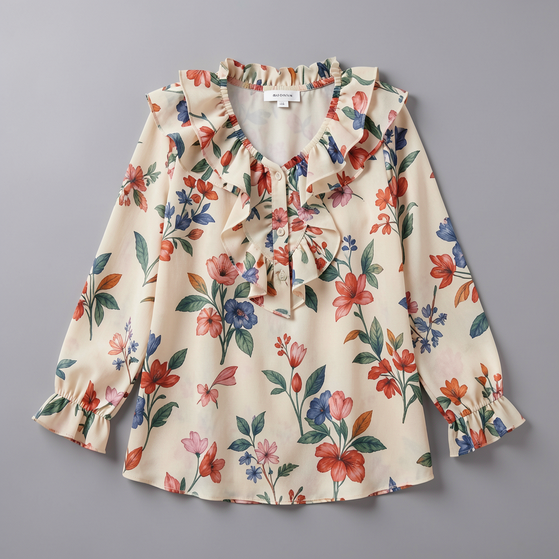}\hfill
  \includegraphics[width=0.23\textwidth]{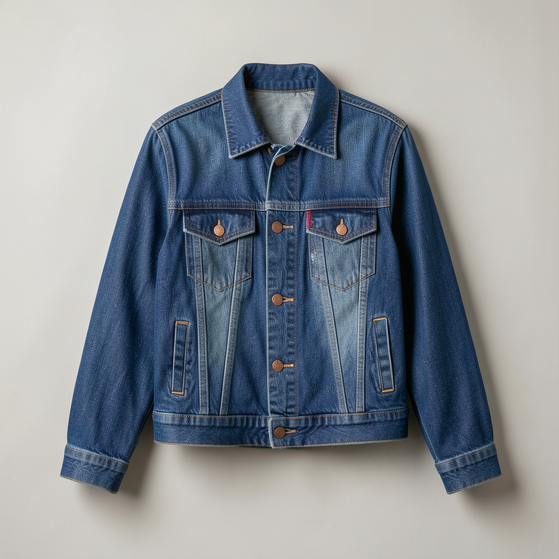}\hfill
  \includegraphics[width=0.23\textwidth]{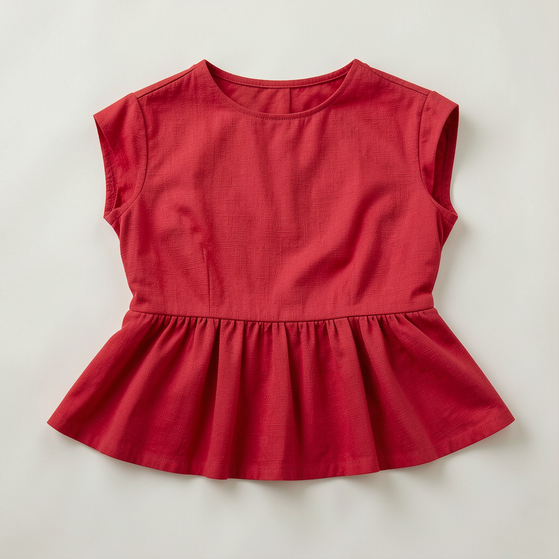}\hfill
  \includegraphics[width=0.23\textwidth]{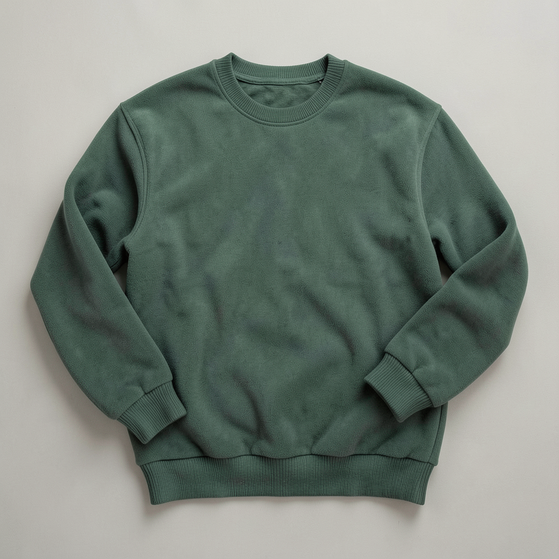}\\[4pt]
  \includegraphics[width=0.23\textwidth]{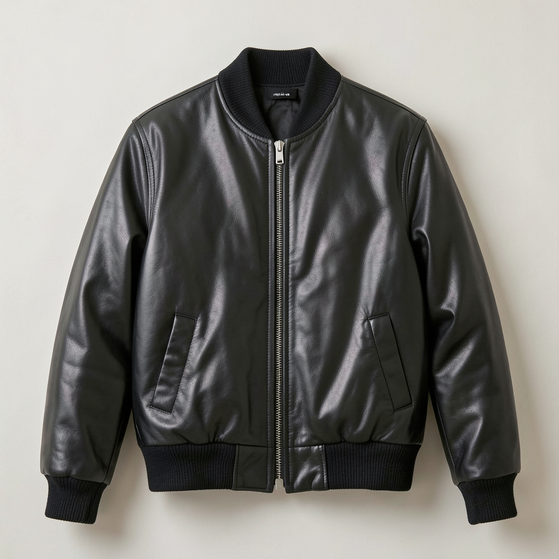}\hfill
  \includegraphics[width=0.23\textwidth]{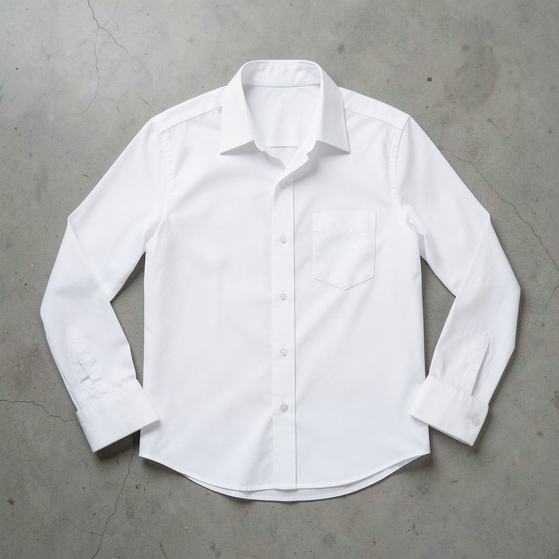}\hfill
  \includegraphics[width=0.23\textwidth]{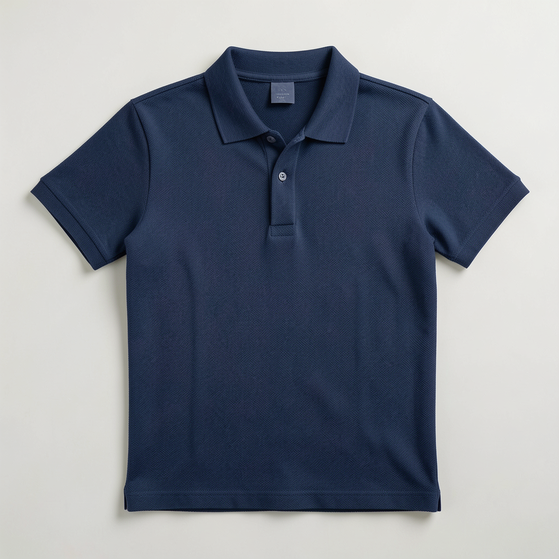}\hfill
  \includegraphics[width=0.23\textwidth]{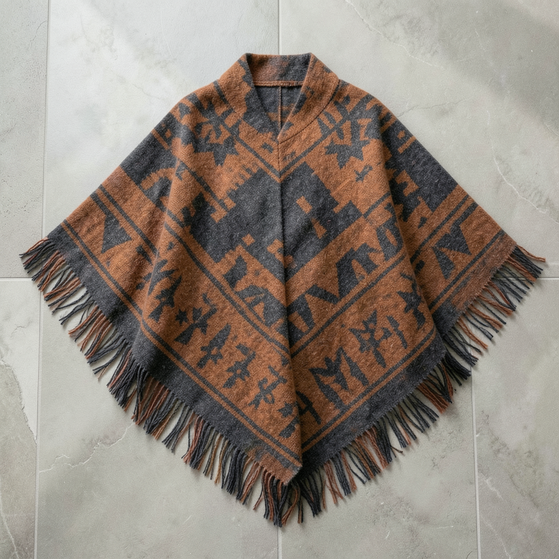}
  \caption{\textbf{Try-on of upper-body garments.} Top row (left to right): women wearing a ruffle top, a denim jacket, a peplum top, and a sweatshirt. Bottom row (left to right): men wearing a bomber jacket, a dress shirt, a polo shirt, and a poncho. These results highlight consistent collar, sleeve, and torso coverage over different types of garments.}
  \label{fig:app_upper_body}
\end{figure*}

\begin{figure*}[!htbp]
  \centering
  \includegraphics[width=0.23\textwidth]{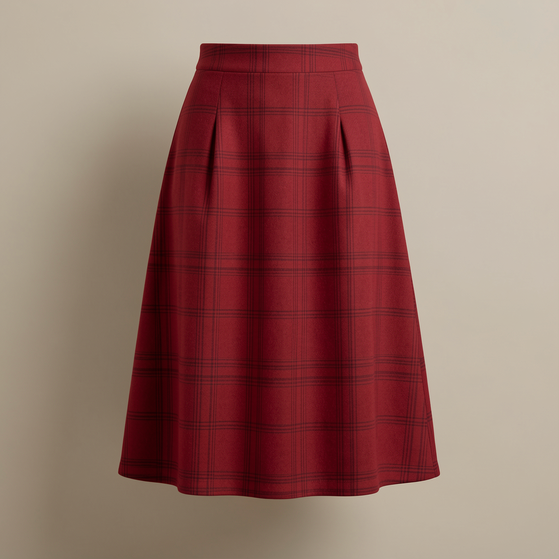}\hfill
  \includegraphics[width=0.23\textwidth]{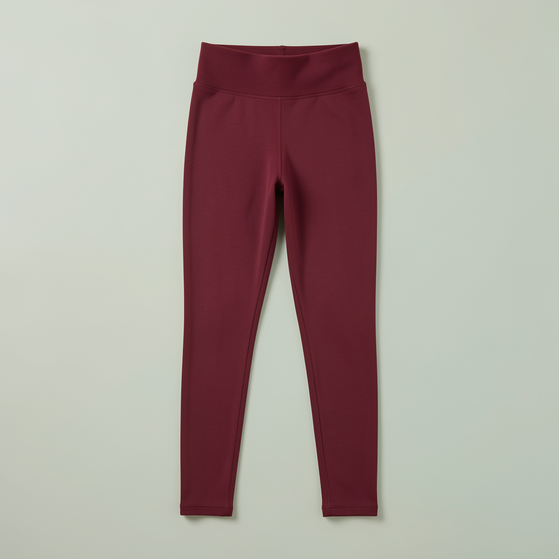}\hfill
  \includegraphics[width=0.23\textwidth]{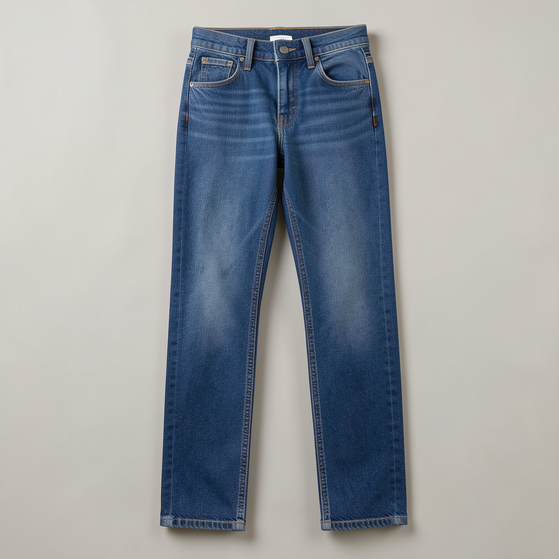}\hfill
  \includegraphics[width=0.23\textwidth]{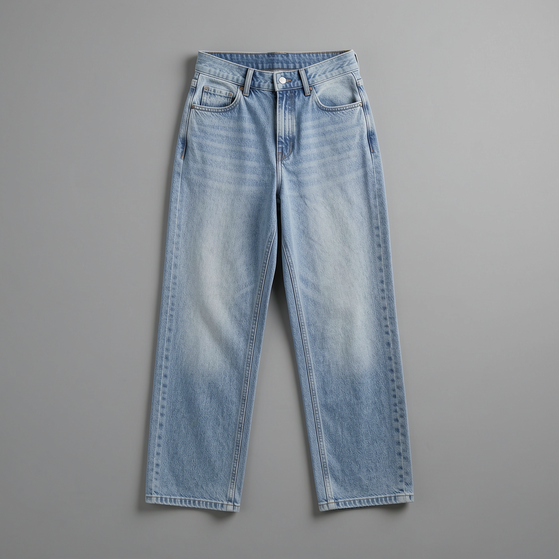}\\[4pt]
  \includegraphics[width=0.23\textwidth]{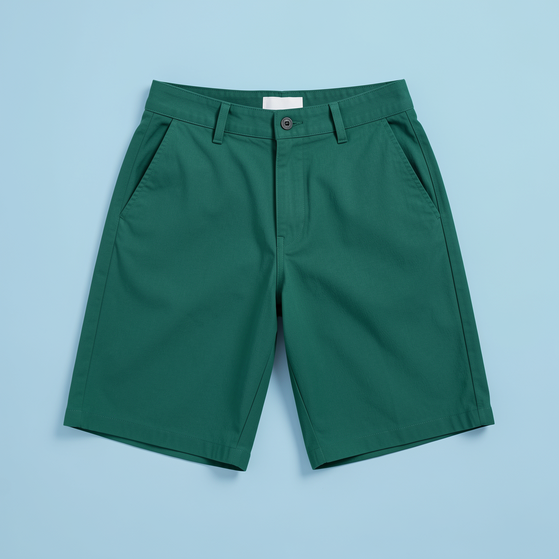}\hfill
  \includegraphics[width=0.23\textwidth]{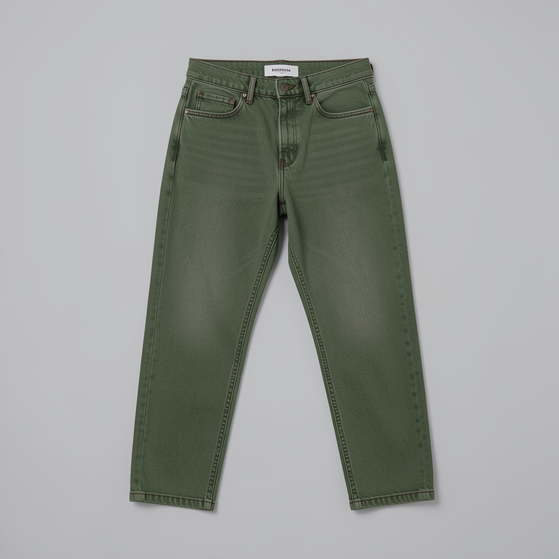}\hfill
  \includegraphics[width=0.23\textwidth]{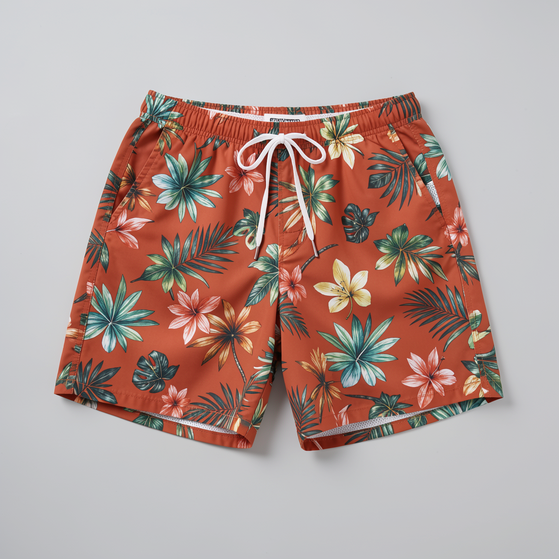}\hfill
  \includegraphics[width=0.23\textwidth]{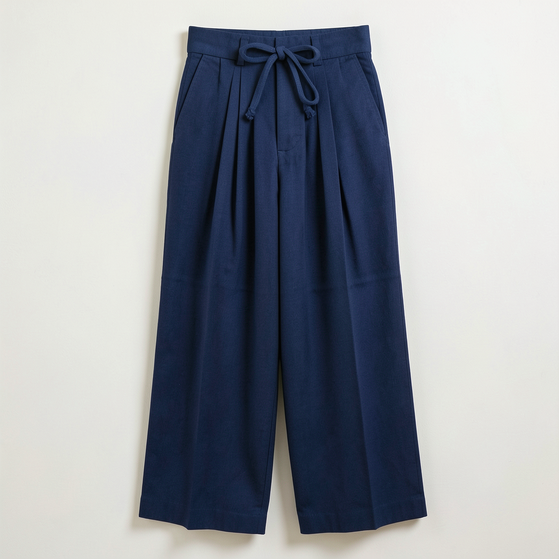}
  \caption{\textbf{Try-on of lower-body garments.} Top row (left to right): women wearing a skirt, leggings, jeans, and mom jeans. Bottom row (left to right): men wearing shorts, jeans, swim trunks, and a traditional hakama. These results indicate good handling of diverse silhouette types, ranging from fitted, tapered, and flowing lower body garments.}
  \label{fig:app_lower_body}
\end{figure*}

\begin{figure*}[!htbp]
  \centering
  \includegraphics[width=0.23\textwidth]{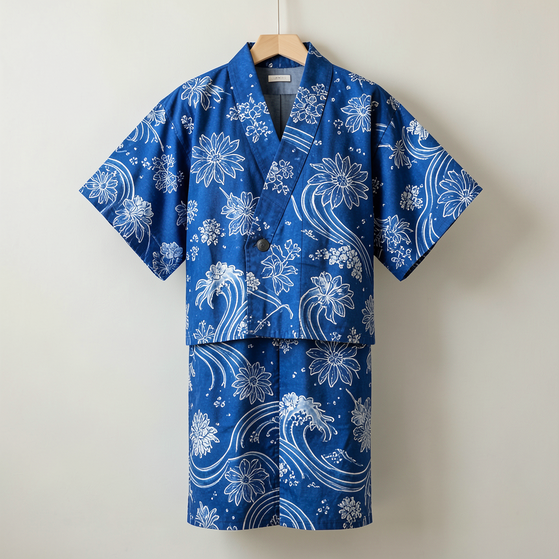}\hfill
  \includegraphics[width=0.23\textwidth]{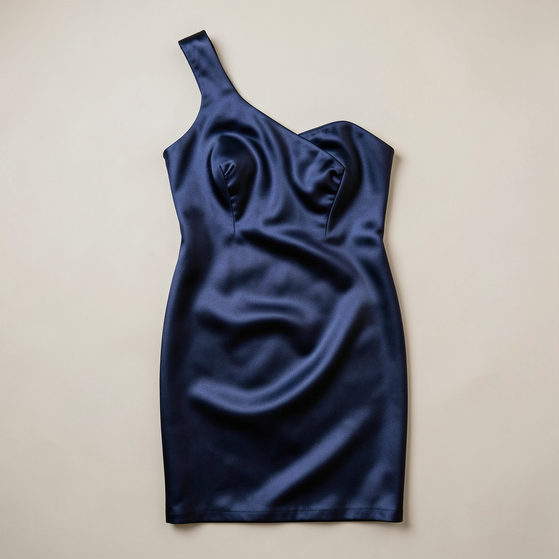}\hfill
  \includegraphics[width=0.23\textwidth]{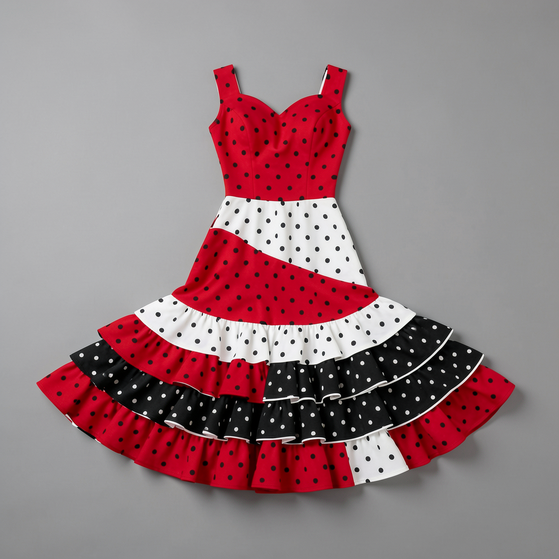}\hfill
  \includegraphics[width=0.23\textwidth]{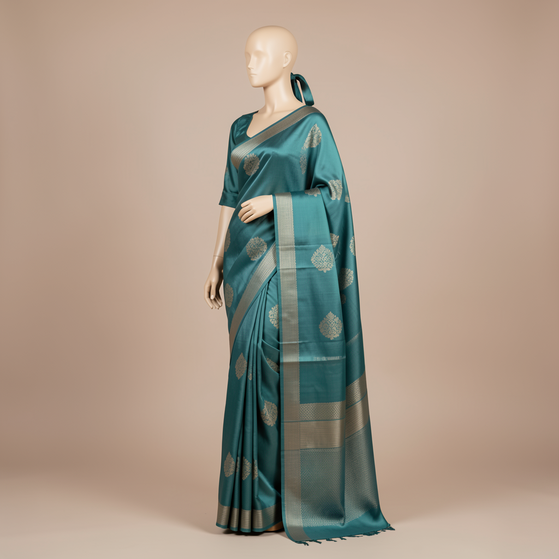}\\[4pt]
  \includegraphics[width=0.23\textwidth]{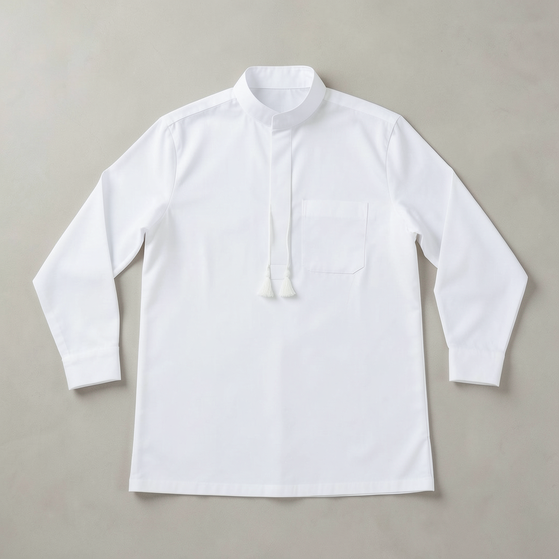}\hfill
  \includegraphics[width=0.23\textwidth]{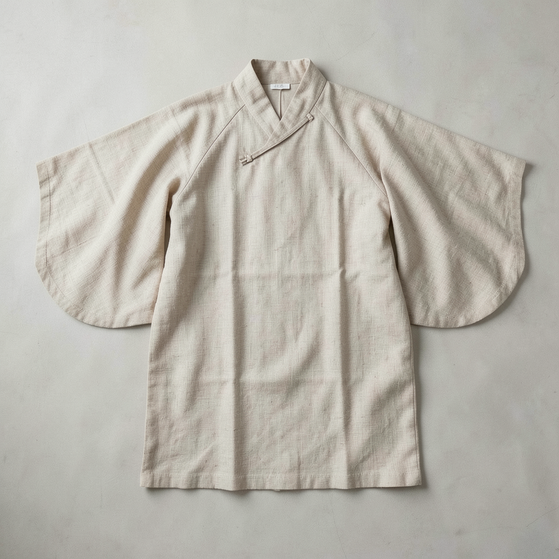}\hfill
  \includegraphics[width=0.23\textwidth]{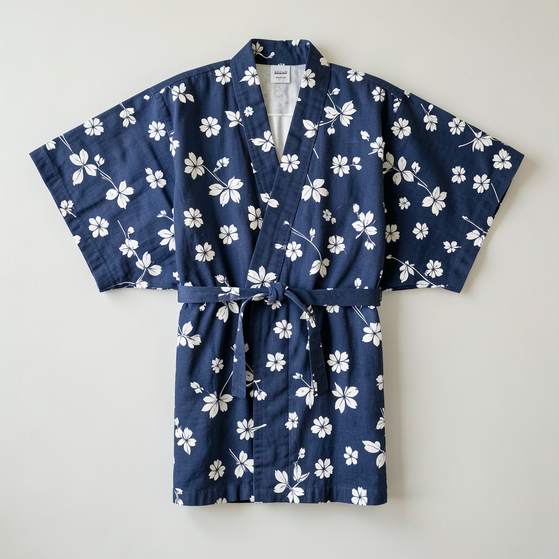}\hfill
  \includegraphics[width=0.23\textwidth]{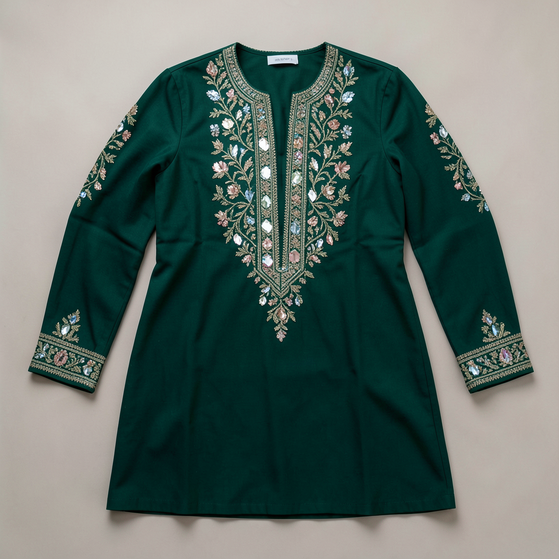}
  \caption{\textbf{Full-body and traditional garment try-on examples.} Top row: women wearing a Ryusou, a cocktail dress, a Flamenco dress, and a Muga Silk Saree. Bottom row: men wearing a Kandura, a Shenyi, a Yukata, and a Pashtun dress. Wide variety of garments in our corpus enables us evaluate try-ons of culturally specific garments not covered by other VTON datasets.}
  \label{fig:app_dresses}
\end{figure*}

Figures~\ref{fig:app_upper_body}--\ref{fig:app_dresses} show the quality of try-ons for the three major categories of garments used in the CURVTON-205K corpus: upper body, lower body, and full-body/dress type garments. The selection in these figures covers both common Western clothing as well as culturally distinctive clothing in order to showcase the wide range of the 200-type dictionary used to guide try-on synthesis. The results demonstrate consistent garment-identity preservation and plausible placement across varied garment structures.


\subsection{Closed-Loop VLM Feedback Mechanism: Illustrative Examples}
\label{app:closed_loop_examples}

\begin{figure*}[!htbp]
  \centering
  \includegraphics[width=\textwidth]{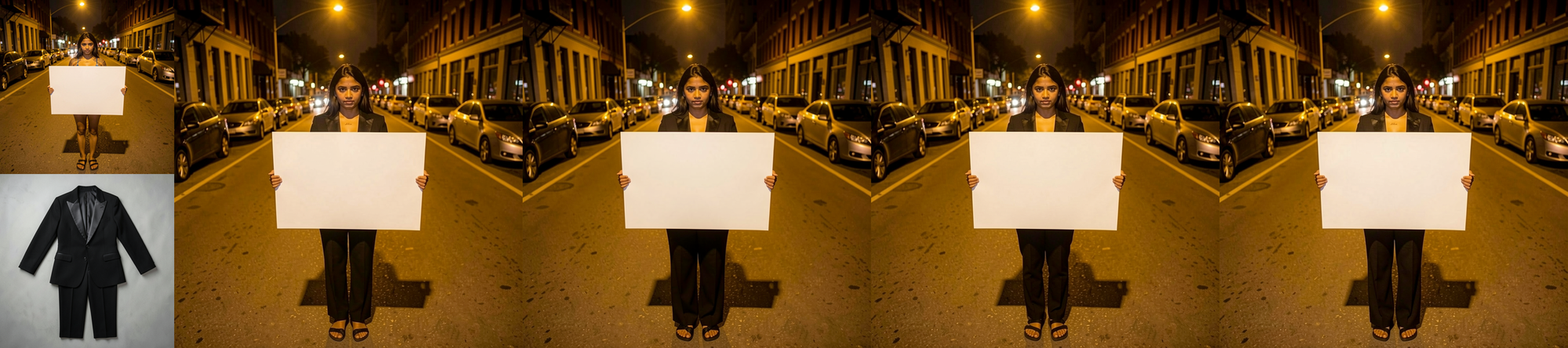}\\[6pt]
  \includegraphics[width=\textwidth]{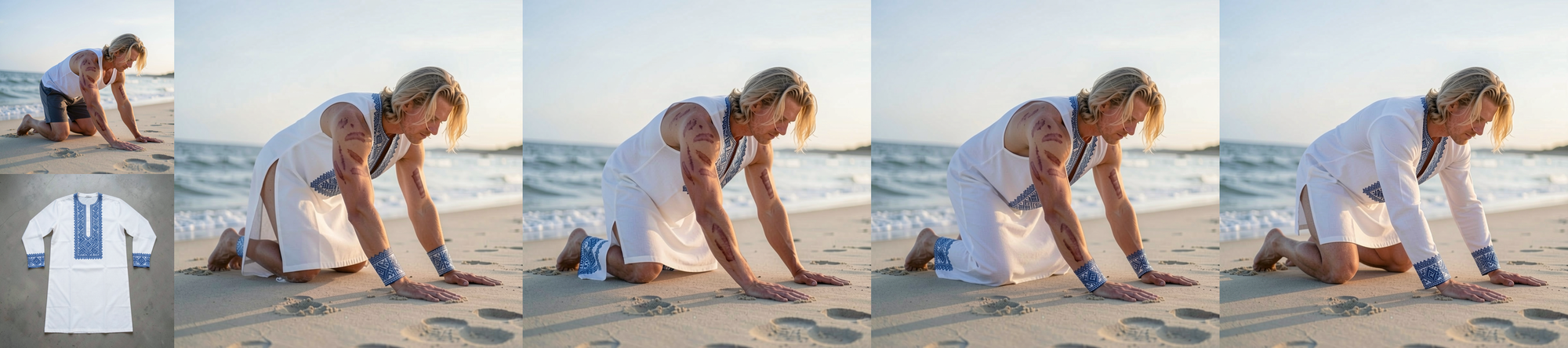}\\[6pt]
  \includegraphics[width=\textwidth]{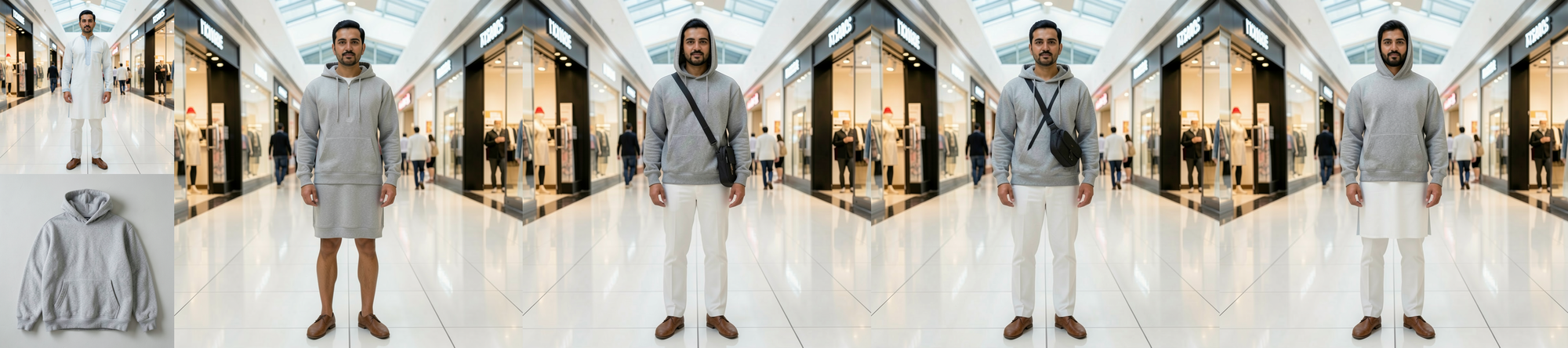}
  \caption{\textbf{Closed-loop VLM feedback: accepted triplets.} For triplets passing the VLM-based quality check, each figure shows the source person (left), the target garment (middle), and the generated try-on (right). First row: a formal power suit for a female person; second row: a traditional Ghalek outfit for a male person; third row: a casual hoodie. These accepted samples exhibit consistent garment identity, plausible placement, and no visible artifacts, illustrating the quality maintained by the automated validation loop.}
  \label{fig:app_closed_loop}
\end{figure*}

Figure~\ref{fig:app_closed_loop} illustrates representative output triplets produced by the closed-loop VLM-guided regeneration mechanism described in Section~\ref{subsec:data_synth} of the main paper. In each example, the VLM assessor accepts the synthesized try-on once it finishes evaluating the garment faithfulness, placement plausibility, semantic consistency, and the absence of visible artifacts. These examples span diverse garment styles and subject appearances to showcase the range of cases handled by the automated quality loop. A formal power suit (left), a traditional Ghalek (center), and a casual hoodie (right).

\subsection{Extended Person Diversity Gallery}
\label{app:person_diversity}

\begin{figure*}[!htbp]
  \centering
  \includegraphics[width=0.185\textwidth]{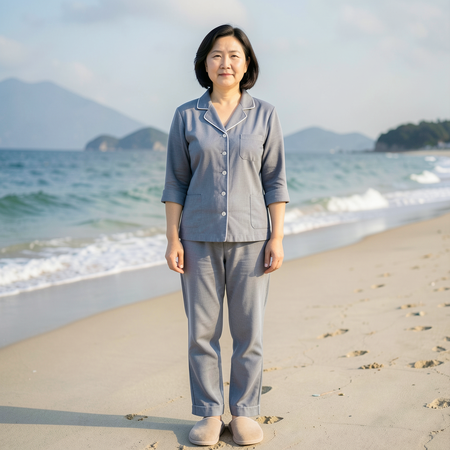}\hfill
  \includegraphics[width=0.185\textwidth]{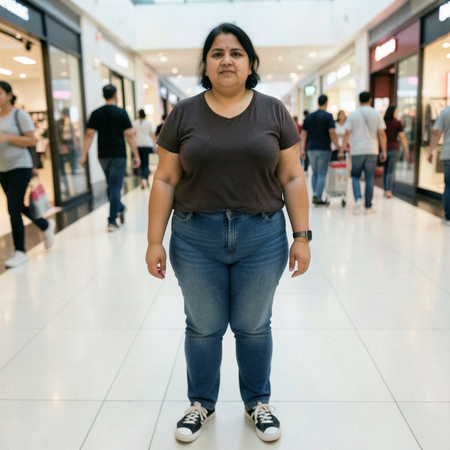}\hfill
  \includegraphics[width=0.185\textwidth]{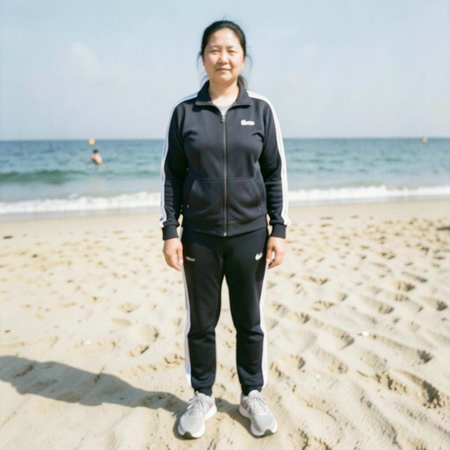}\hfill
  \includegraphics[width=0.185\textwidth]{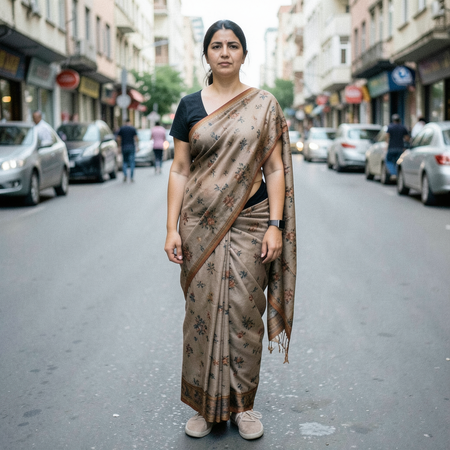}\hfill
  \includegraphics[width=0.185\textwidth]{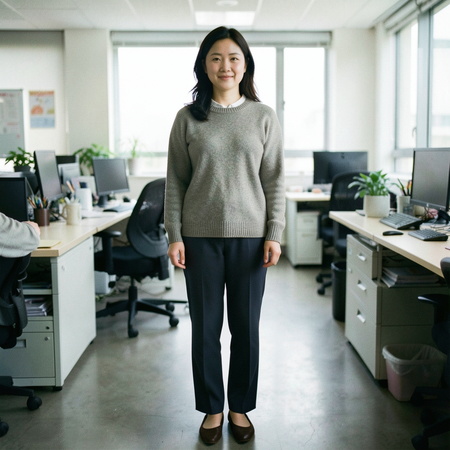}\\[3pt]
  \includegraphics[width=0.185\textwidth]{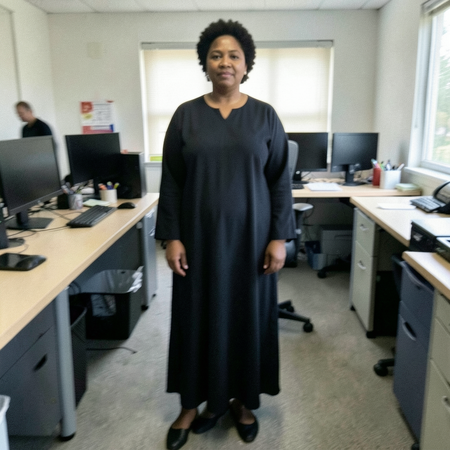}\hfill
  \includegraphics[width=0.185\textwidth]{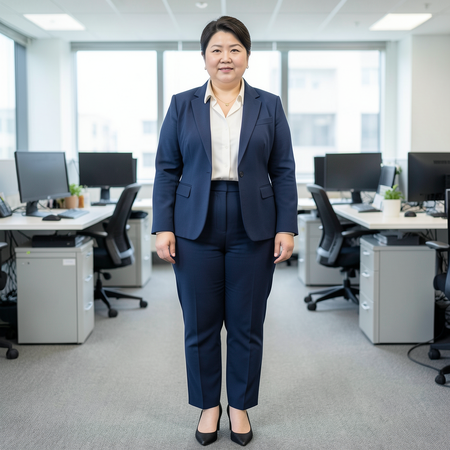}\hfill
  \includegraphics[width=0.185\textwidth]{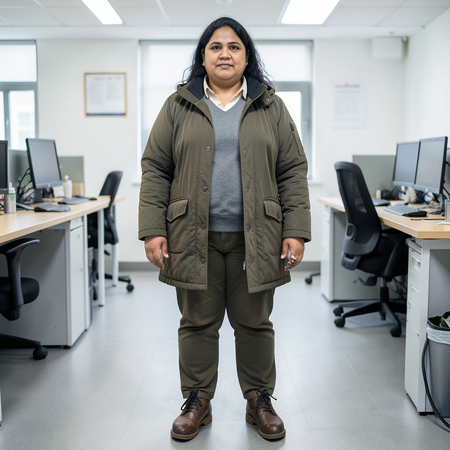}\hfill
  \includegraphics[width=0.185\textwidth]{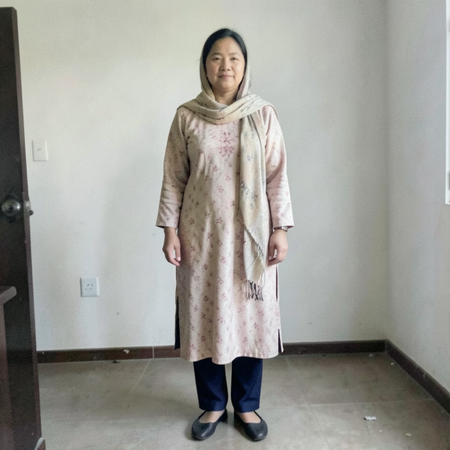}\hfill
  \includegraphics[width=0.185\textwidth]{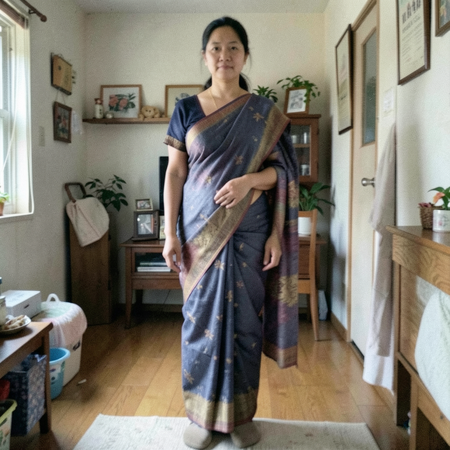}\\[6pt]
  \includegraphics[width=0.185\textwidth]{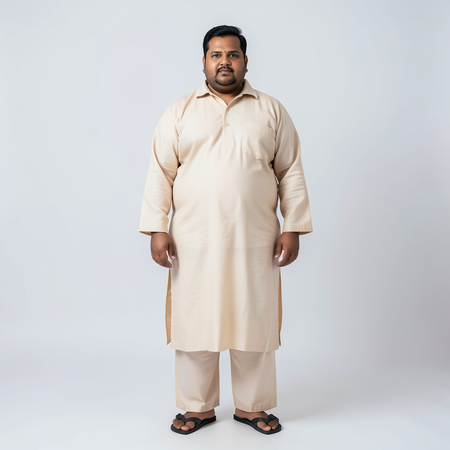}\hfill
  \includegraphics[width=0.185\textwidth]{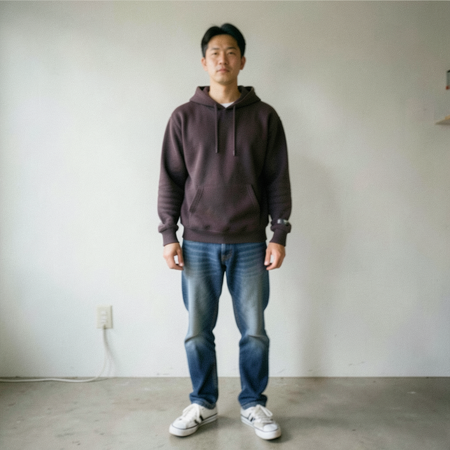}\hfill
  \includegraphics[width=0.185\textwidth]{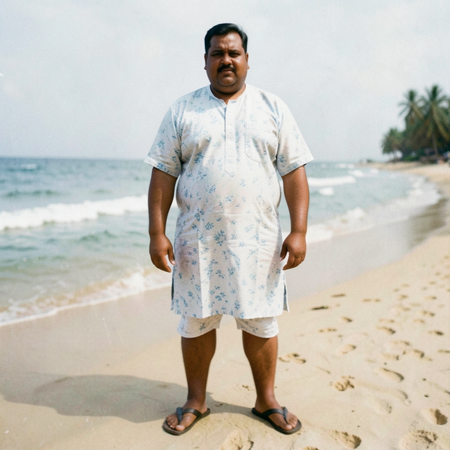}\hfill
  \includegraphics[width=0.185\textwidth]{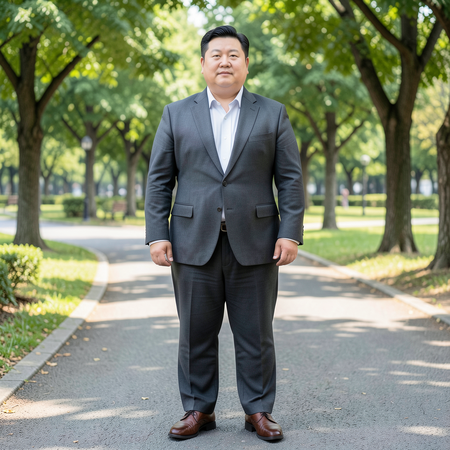}\hfill
  \includegraphics[width=0.185\textwidth]{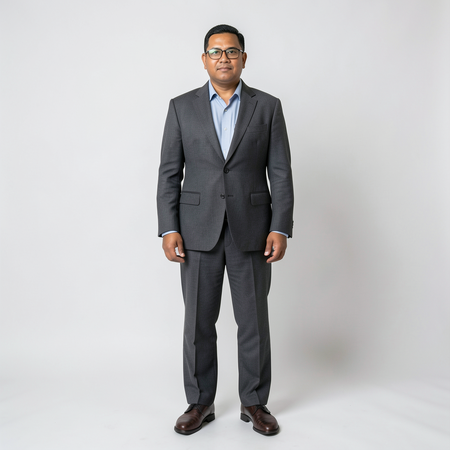}\\[3pt]
  \includegraphics[width=0.185\textwidth]{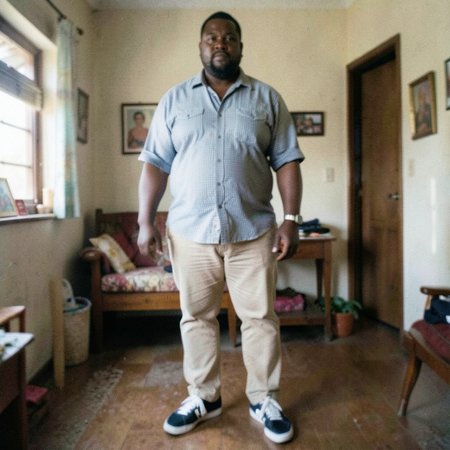}\hfill
  \includegraphics[width=0.185\textwidth]{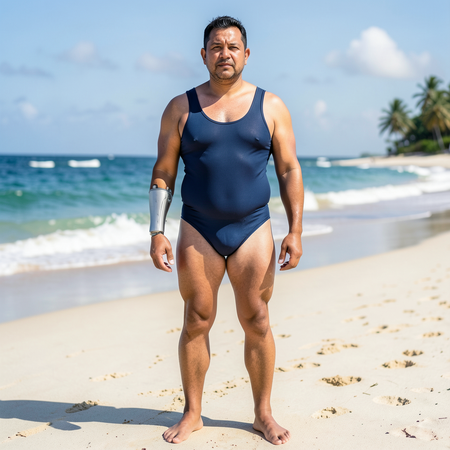}\hfill
  \includegraphics[width=0.185\textwidth]{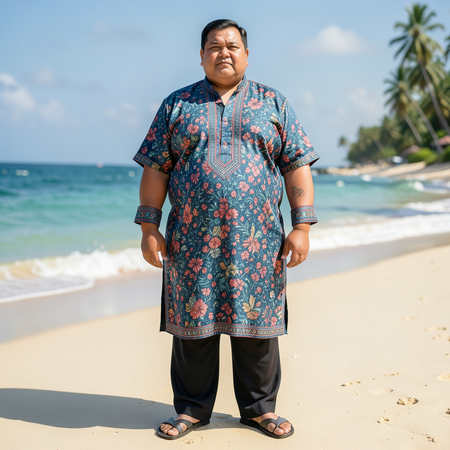}\hfill
  \includegraphics[width=0.185\textwidth]{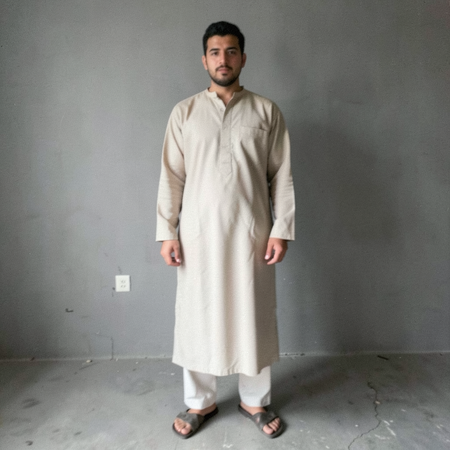}
  \caption{\textbf{Extended person diversity gallery.} First two rows of female persons and next two rows of male persons from the CURVTON-205K dataset .Body shape, facial appearance, skin tone, accessories, and photography conditions (both studio and in-the-wild) are covered thanks to the diversity introduced with factorized attribute sampling strategy $\mathbb{P}_{\mathcal{D}_p}$. This breadth of person appearance is a key differentiator of CURVTON-205K relative to existing paired VTON benchmarks.}
  \label{fig:app_person_diversity}
\end{figure*}

Figure~\ref{fig:app_person_diversity} presents a collection of source person images sampled from CURVTON-205K to illustrate the diversity in demographic and appearance targeted by the person-stream attribute dictionary $\mathcal{D}_p$. The collection shows variation in body shape, facial-appearance details, geographic and cultural style indicators, photographic conditions (studio vs. in-the-wild), accessories, and background context. This vastness is a direct result of sampling attributes from the factorized distribution $\mathbb{P}_{\mathcal{D}_p}$ described in Section~\ref{subsec:data_synth} of the main paper, which is designed to avoid long-tail collapse and improve coverage of groups that are under-represented.

\subsection{Additional Closed-Loop VLM Feedback Examples}
\label{app:closed_loop_extra}

\begin{figure*}[!htbp]
  \centering
  \includegraphics[width=\textwidth]{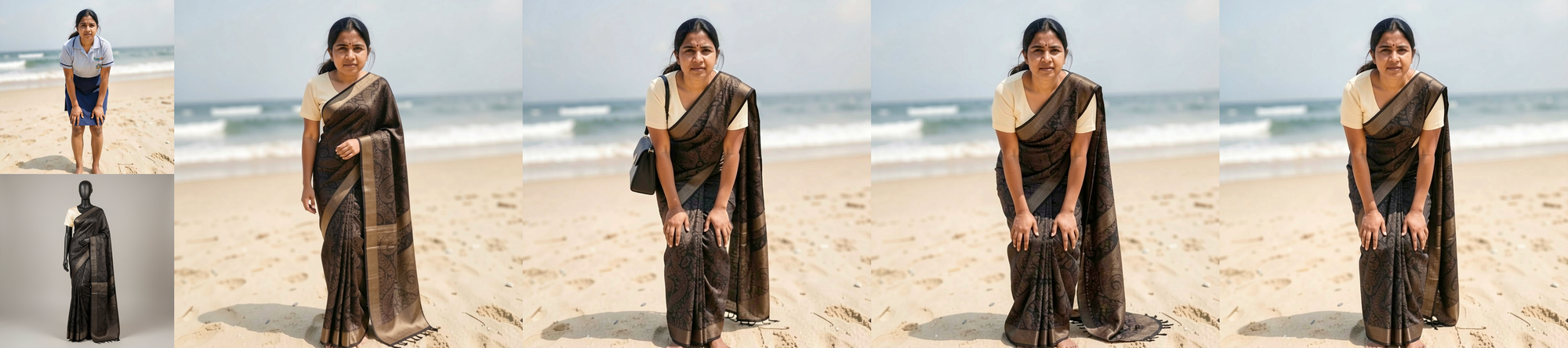}\\[6pt]
  \includegraphics[width=\textwidth]{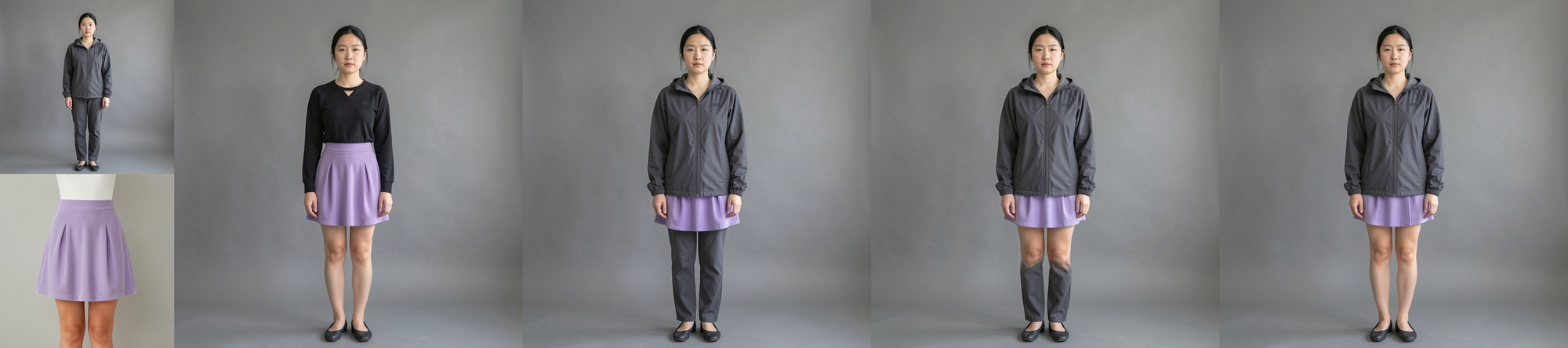}\\[6pt]
  \includegraphics[width=\textwidth]{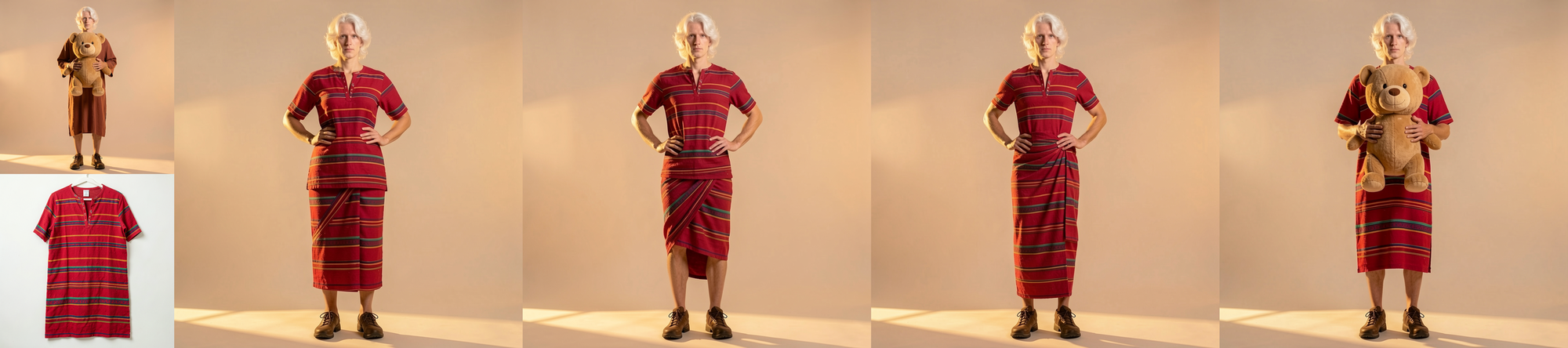}
  \caption{\textbf{Additional closed-loop VLM feedback: accepted triplets.} For accepted triplets based on VLM quality check, each figure provides the source person (left), target garment (middle), and generated try-on (right). Top: Jamdani Saree on a female subject; center: mini skirt on a female subject; bottom: Chakma dress on a male subject. These samples further illustrate the breadth of garment types handled by the automated validation loop.}
  \label{fig:app_closed_loop_extra}
\end{figure*}

Figure~\ref{fig:app_closed_loop_extra} shows three more samples of accepted triplets using our VLM-based automated validation process. This figure complements our discussion of the closed-loop validation system in Section~\ref{app:closed_loop_examples}. The examples include a traditional Jamdani Saree on a female subject, a casual mini skirt, and a Chakma dress on a male subject, illustrating the quality loop's performance on both conventional Western garments and culturally specific traditional clothing.

\subsection{Extended Medium-Tier Try-On Examples}
\label{app:medium_extended}

\begin{figure*}[!htbp]
  \centering
  \includegraphics[width=0.31\textwidth]{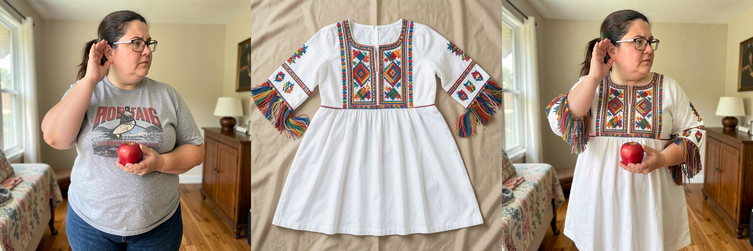}\hfill
  \includegraphics[width=0.31\textwidth]{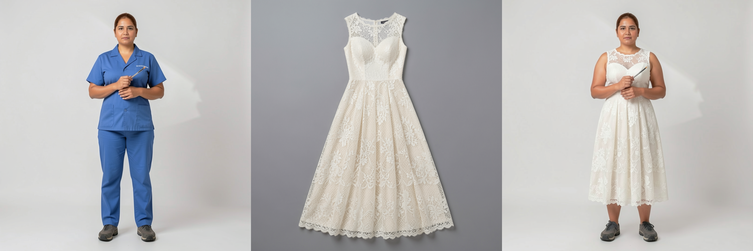}\hfill
  \includegraphics[width=0.31\textwidth]{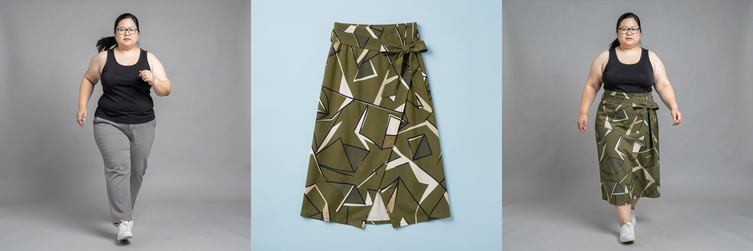}\\[4pt]
  \includegraphics[width=0.31\textwidth]{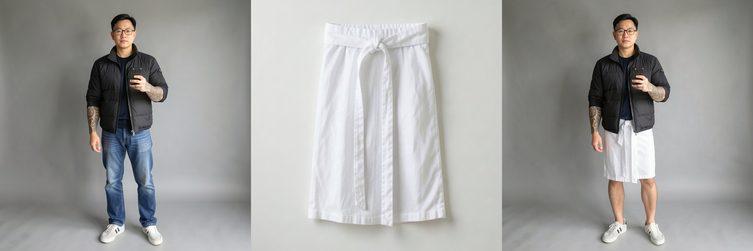}\hfill
  \includegraphics[width=0.31\textwidth]{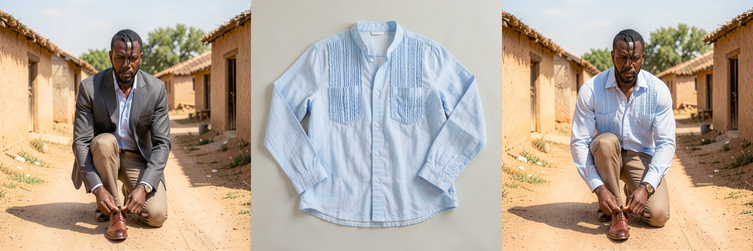}\hfill
  \includegraphics[width=0.31\textwidth]{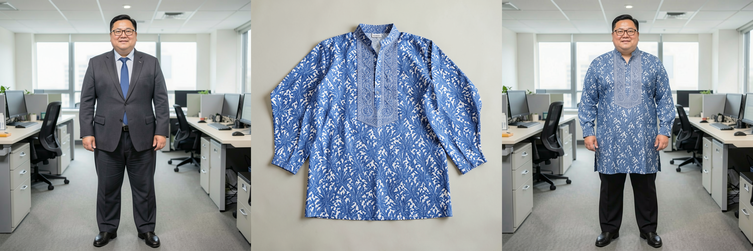}
  \caption{\textbf{Extended medium-tier try-on triplets.} Top row: female subjects wearing (left to right) a Berber Amazigh dress, a lace dress, and a wrap skirt. Bottom row: male subjects wear a Fundoshi, a Guayabera shirt, and a Dashiki. The culturally-specific clothes indicate the representation of rare classes of clothing in CURVTON-205K.}
  \label{fig:app_medium_ext}
\end{figure*}

Figure~\ref{fig:app_medium_ext} shows six more medium-level try-on triplets in addition to those in Figure~\ref{fig:app_medium}. The examples shown here cover increased cultural diversity, with a Berber Amazigh dress, a lace dress, a wrap skirt, a Fundoshi, a Guayabera shirt, and a Dashiki. 


\subsection{Extended Hard-Tier Try-On Examples}
\label{app:hard_extended}

\begin{figure*}[!htbp]
  \centering
  \includegraphics[width=0.47\textwidth]{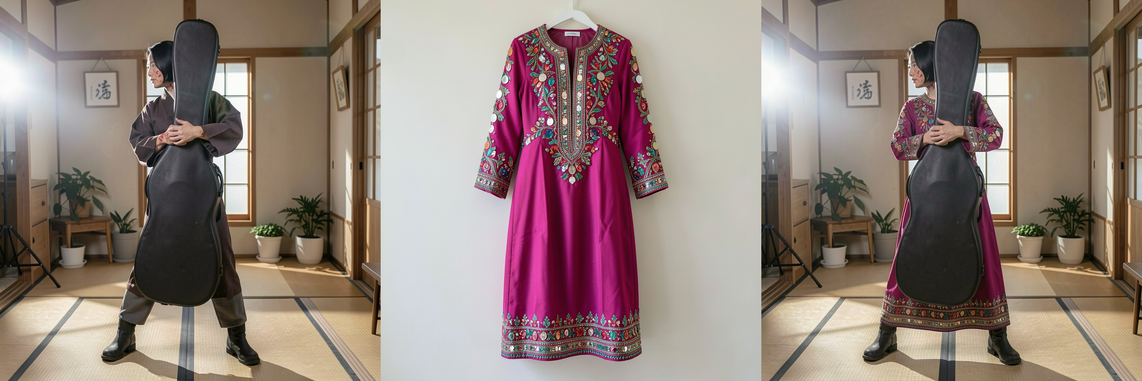}\hfill
  \includegraphics[width=0.47\textwidth]{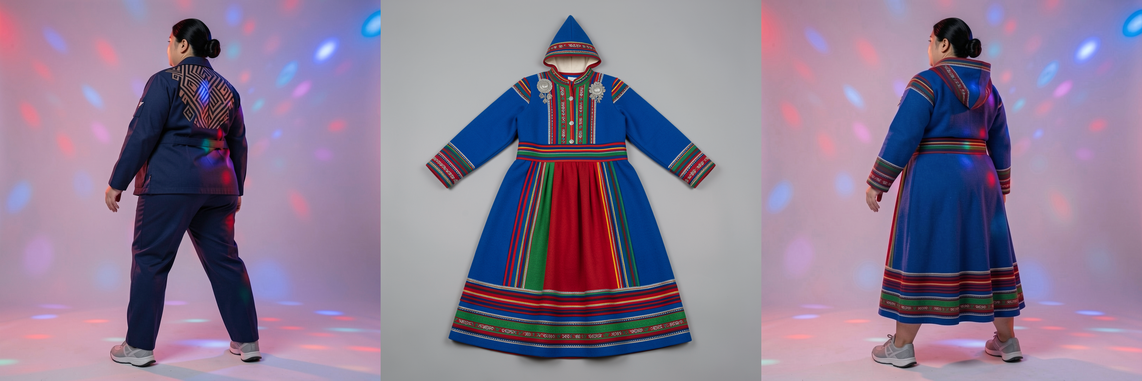}
  \caption{\textbf{Extended hard-tier try-on triplets.} Left image: an Afghan Kuchi dress worn by a female subject with a complex pose; right image: a G\"{o}kti worn by a female subject with a cluttered background. These images show the difficulty in maintaining garment identity in challenging conditions.}
  \label{fig:app_hard_ext}
\end{figure*}

Figure~\ref{fig:app_hard_ext} gives two more hard-level try-on triplets. The images here include extremely large pose deviations with rare traditional clothes: an Afghan Kuchi dress and a \textit{G\"{o}kti} worn by female subjects. They show the difficulty of preserving the identity of the clothes when dealing with pose deviation and high levels of occlusions.

\subsection{Traditional Garment Coverage}
\label{app:traditional_garments}

\begin{figure*}[!htbp]
  \centering
  \includegraphics[width=0.23\textwidth]{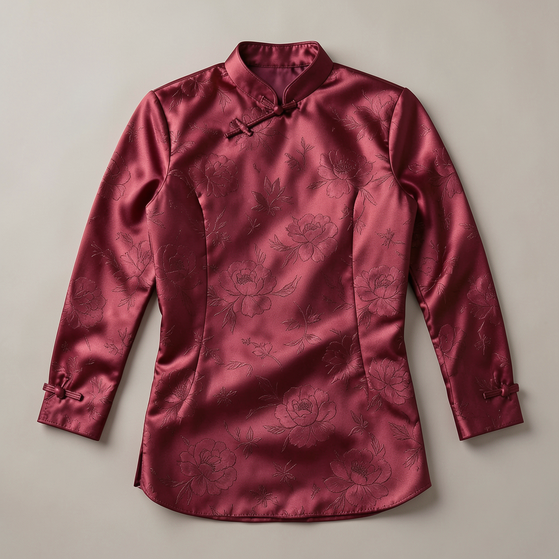}\hfill
  \includegraphics[width=0.23\textwidth]{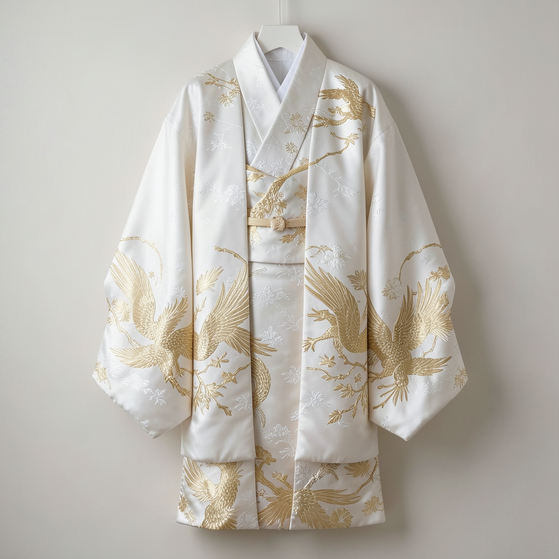}\hfill
  \includegraphics[width=0.23\textwidth]{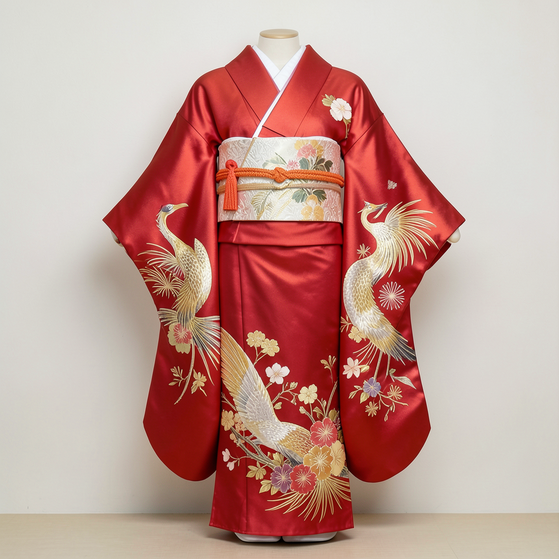}\hfill
  \includegraphics[width=0.23\textwidth]{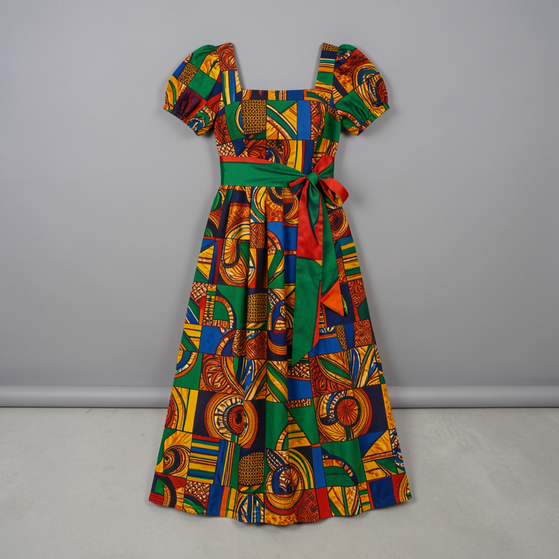}\\[4pt]
  \includegraphics[width=0.23\textwidth]{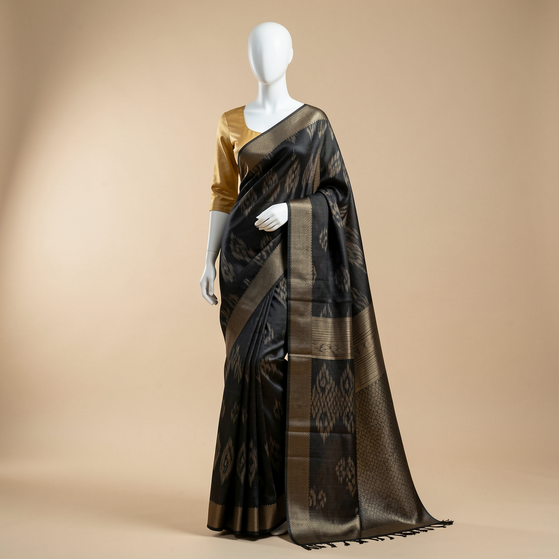}\hfill
  \includegraphics[width=0.23\textwidth]{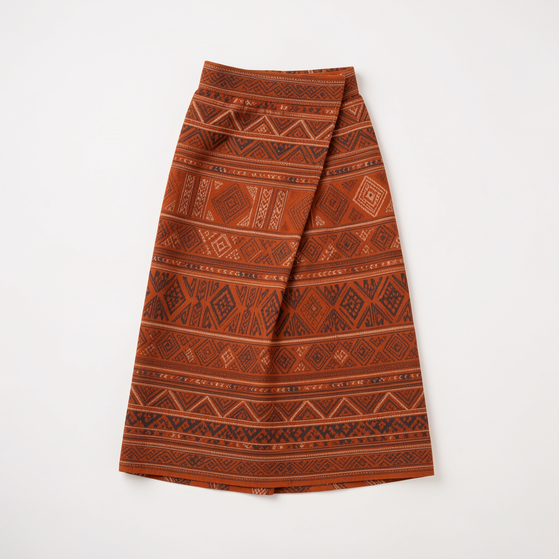}\hfill
  \includegraphics[width=0.23\textwidth]{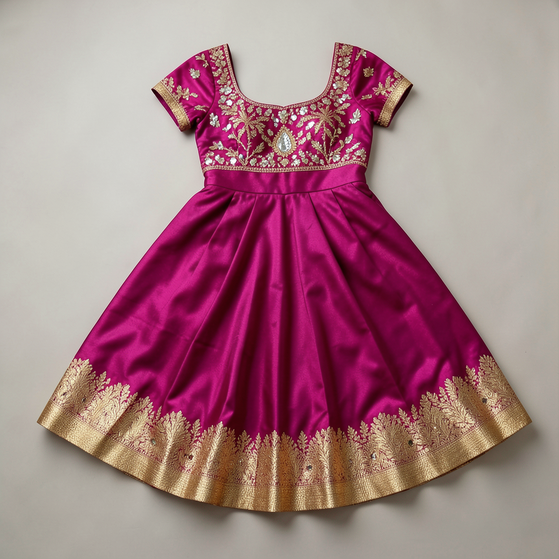}\hfill
  \includegraphics[width=0.23\textwidth]{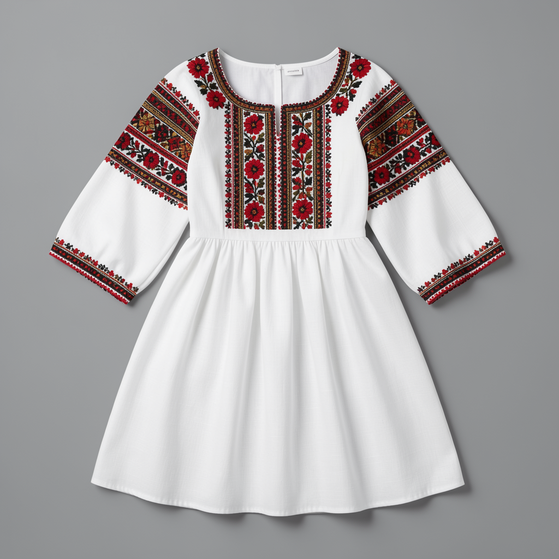}
  \caption{\textbf{Traditional garment try-on — female subjects.} Top row (left to right): Qipao, Uchikake, Furisode, Gomesi. Bottom row: Muga Silk Saree, Limbu Mekhli, Ghagra Choli, Romanian Ie. These clothes come from East Asia, South Asia, East Africa, and Eastern Europe, illustrating the cultural breadth of CURVTON-205K.}
  \label{fig:app_trad_female}
\end{figure*}

\begin{figure*}[!htbp]
  \centering
  \includegraphics[width=0.23\textwidth]{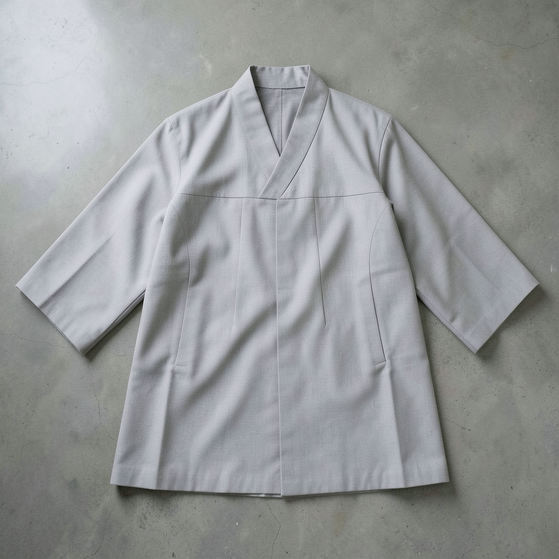}\hfill
  \includegraphics[width=0.23\textwidth]{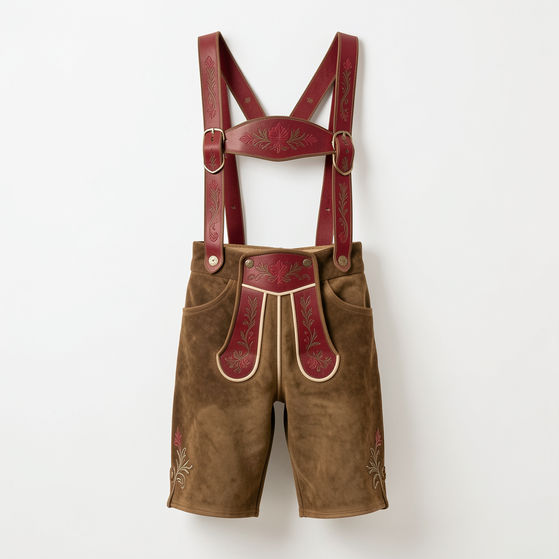}\hfill
  \includegraphics[width=0.23\textwidth]{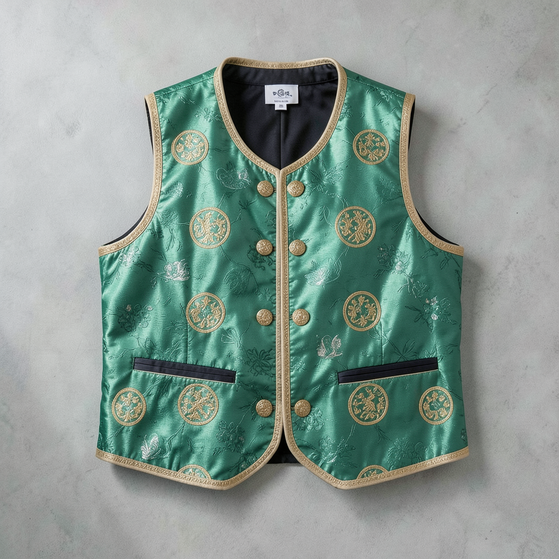}\hfill
  \includegraphics[width=0.23\textwidth]{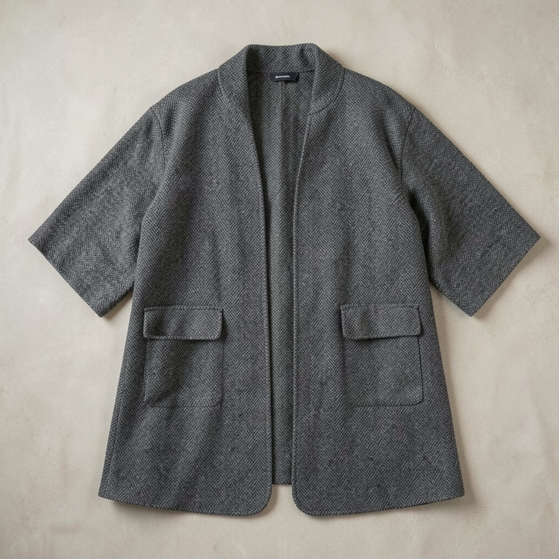}\\[4pt]
  \includegraphics[width=0.23\textwidth]{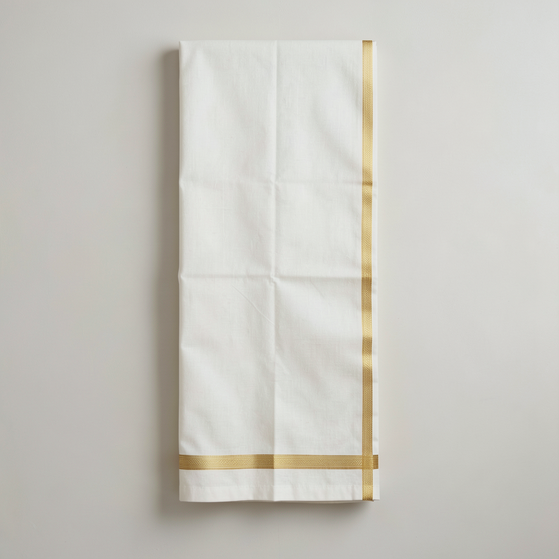}\hfill
  \includegraphics[width=0.23\textwidth]{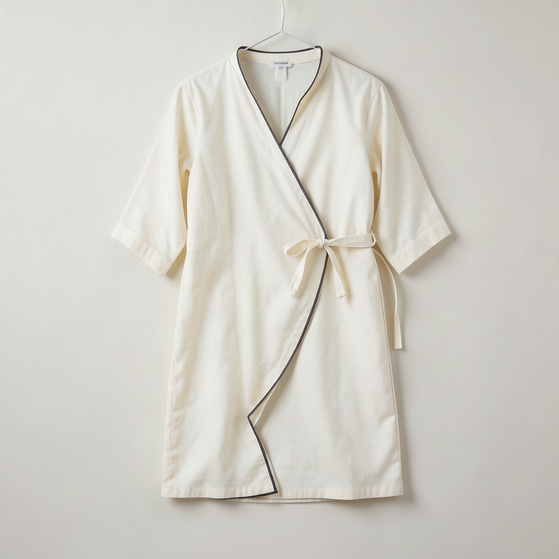}\hfill
  \includegraphics[width=0.23\textwidth]{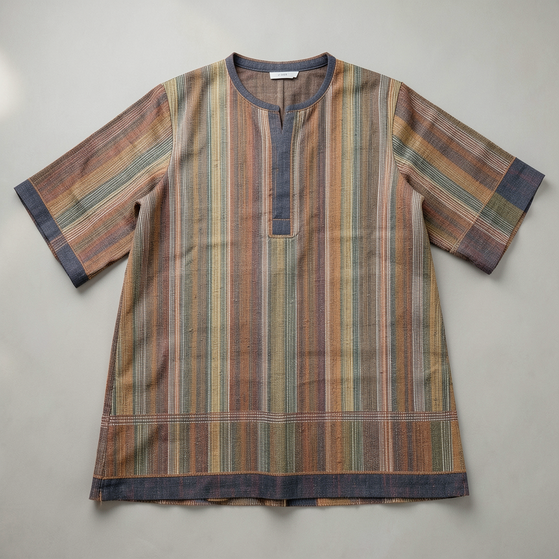}\hfill
  \includegraphics[width=0.23\textwidth]{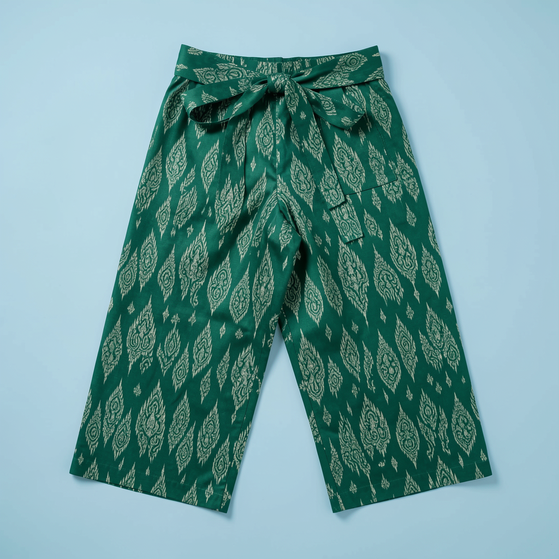}
  \caption{\textbf{Traditional garment try-on — male subjects.} Top row (left to right): Durumagi, Lederhosen, Korean Jeonbok, Ruana. Bottom row: Mundu, Angrakha, Khet Partug, Sampot. These clothes belong to Korea, Germany, South Asia, South America, and Southeast Asia, demonstrating the dataset's global cultural coverage.}
  \label{fig:app_trad_male}
\end{figure*}

Figures~\ref{fig:app_trad_female} and~\ref{fig:app_trad_male} show try-on triplets of garments that belong to the traditional garments collection of the CURVTON-205K garment dictionary. This subset covers traditional garments belonging to different world cultures, including East Asia, South Asia, Southeast Asia, Central Asia, Middle East, Africa, and Europe.

\subsection{Non-Traditional Garment Coverage}
\label{app:nontraditional_garments}

\begin{figure*}[!htbp]
  \centering
  \includegraphics[width=0.23\textwidth]{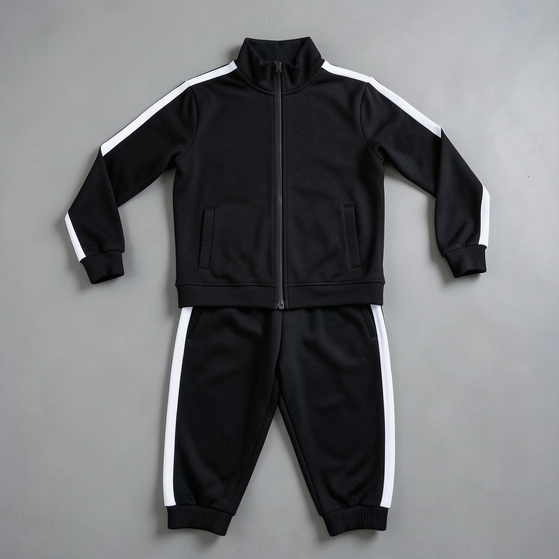}\hfill
  \includegraphics[width=0.23\textwidth]{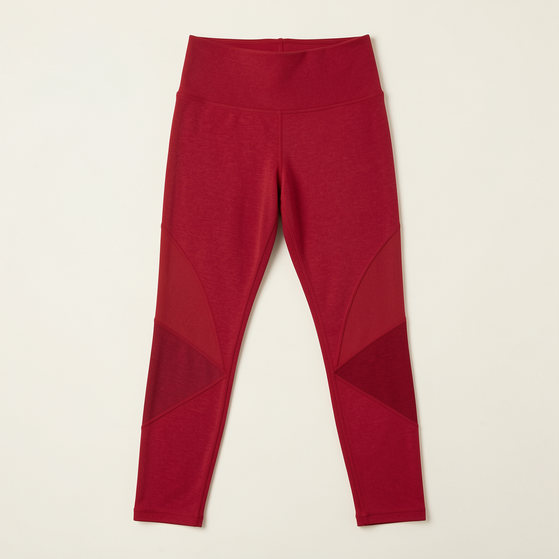}\hfill
  \includegraphics[width=0.23\textwidth]{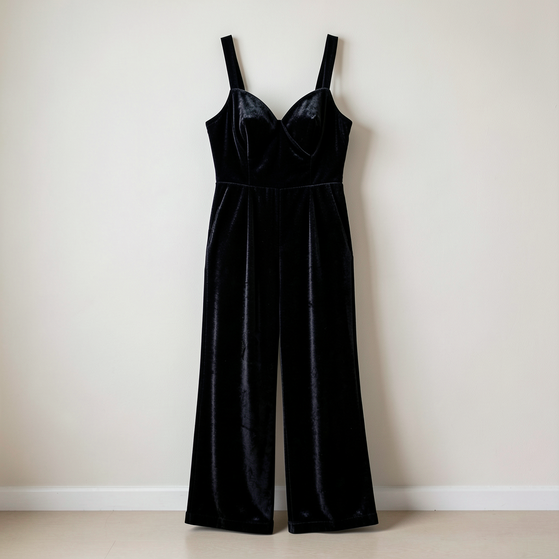}\hfill
  \includegraphics[width=0.23\textwidth]{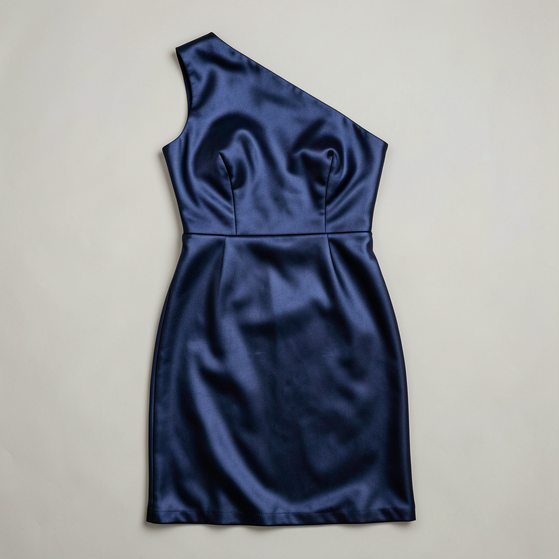}\\[4pt]
  \includegraphics[width=0.23\textwidth]{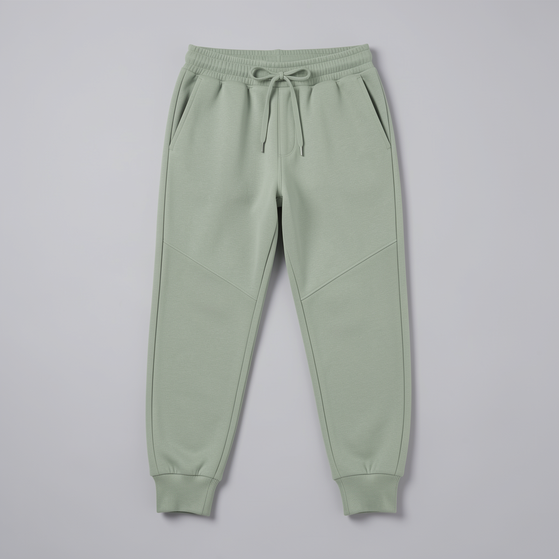}\hfill
  \includegraphics[width=0.23\textwidth]{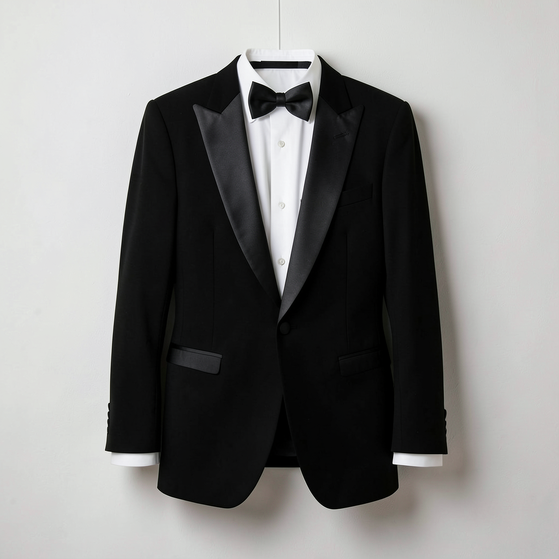}\hfill
  \includegraphics[width=0.23\textwidth]{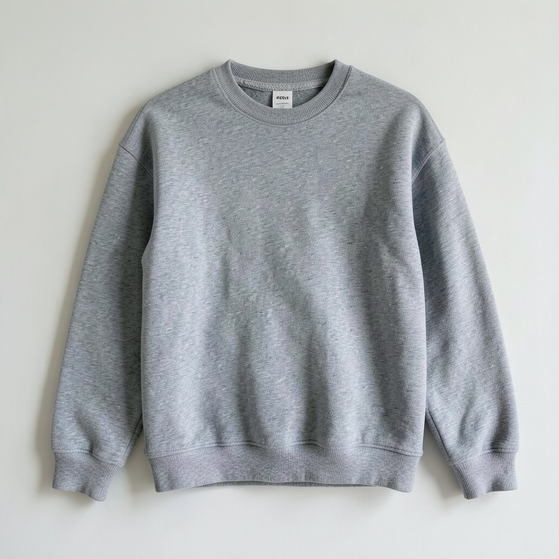}\hfill
  \includegraphics[width=0.23\textwidth]{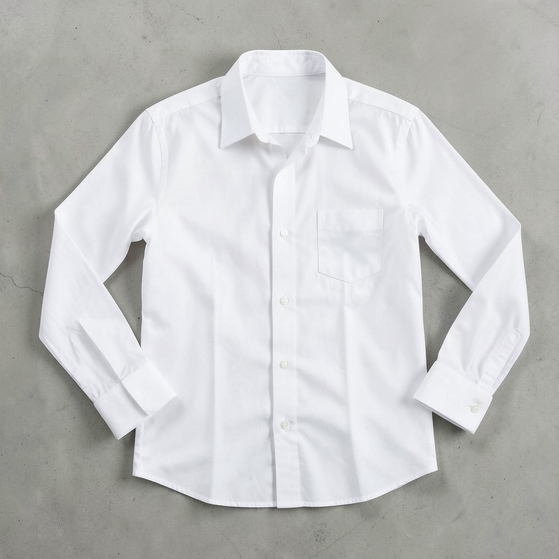}
  \caption{\textbf{Non-traditional (contemporary) garment try-on examples.} From top to bottom: female subjects wearing a tracksuit, yoga pants, jumpsuit, and cocktail dress; male subjects wearing joggers, tuxedo, sweatshirt, and dress shirt. Examples indicate consistency of garment identity representation in various contemporary Western  attires.}

  \label{fig:app_nontrad}
\end{figure*}

Figure~\ref{fig:app_nontrad} 
shows example try-on triplets for non-traditional (contemporary Western) garments. These augment the traditional garments shown above and verify the balanced distribution of traditional vs. contemporary garment styles present in CURVTON-205K.

\section{Dataset Visual Comparison}
\label{app:dataset_comparison}

This section provides a qualitative visual comparison between CURVTON-205K and three widely-used virtual try-on benchmarks: VITON-HD~\cite{Choi2021VITONHD}, DressCode~\cite{Morelli2022DressCode}, and StreetTryOn~\cite{Yang2023StreetTryOn}. Each row below shows representative samples from a single dataset. VITON-HD and DressCode are predominantly composed of studio-shot images of Western garments with frontal poses while StreetTryOn adds in-the-wild backgrounds, yet still limited to Western upper body clothing. In contrast, CURVTON-205K contains diverse pose angles and backgrounds along with a large vocabulary of 200 different types of traditional garments from over 60 cultural categories.

\begin{figure*}[!htbp]
  \centering
  \includegraphics[width=0.185\textwidth]{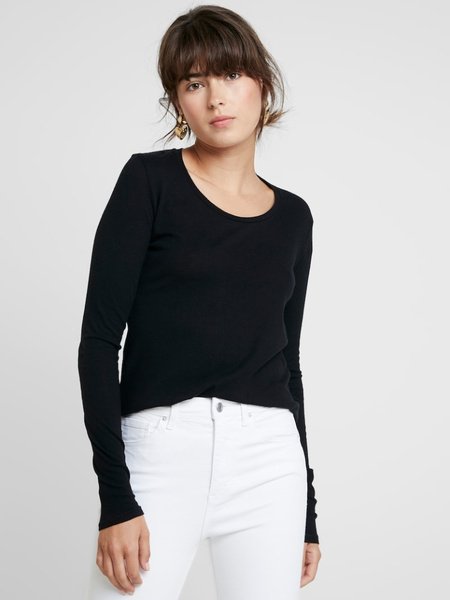}\hfill
  \includegraphics[width=0.185\textwidth]{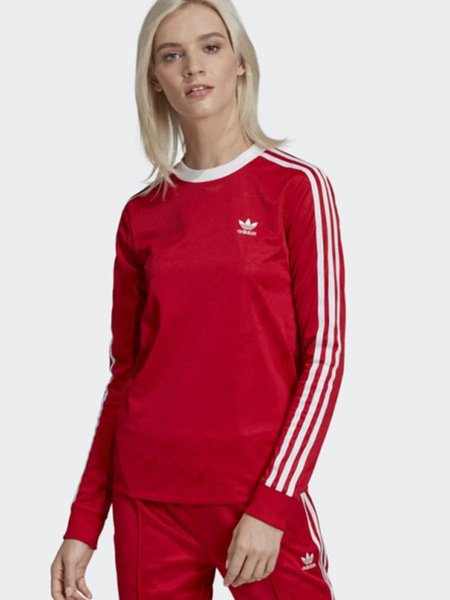}\hfill
  \includegraphics[width=0.185\textwidth]{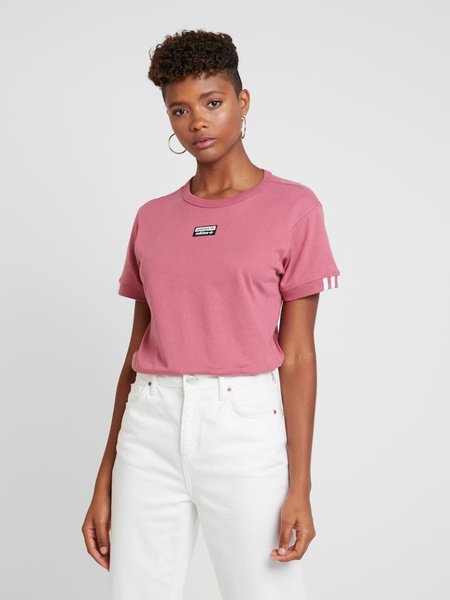}\hfill
  \includegraphics[width=0.185\textwidth]{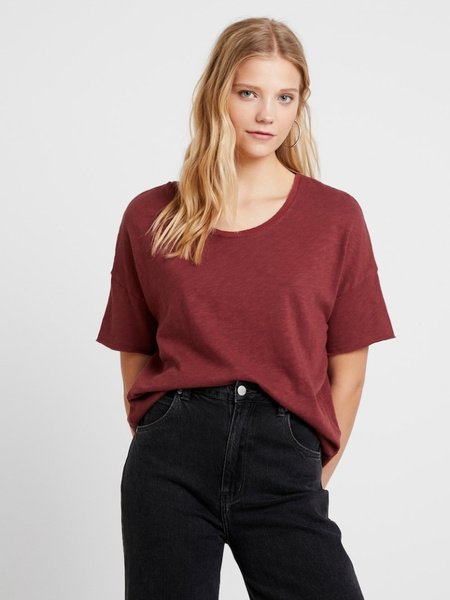}\hfill
  \includegraphics[width=0.185\textwidth]{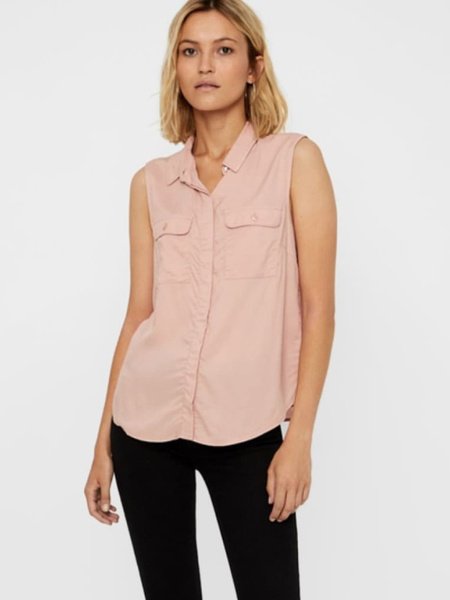}
  \caption{\textbf{VITON-HD samples.}  Representative person images in the VITON-HD dataset. Samples mainly consist of front poses, uniform backgrounds, and upper body clothing. Pose, background and garment diversity are limited in comparison to CURVTON-205K.}
  \label{fig:dataset_comp_vitonhd}
\end{figure*}

\begin{figure*}[!htbp]
  \centering
  \includegraphics[width=0.185\textwidth]{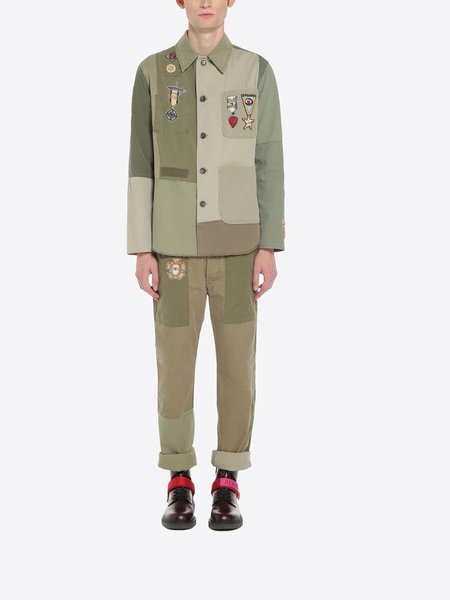}\hfill
  \includegraphics[width=0.185\textwidth]{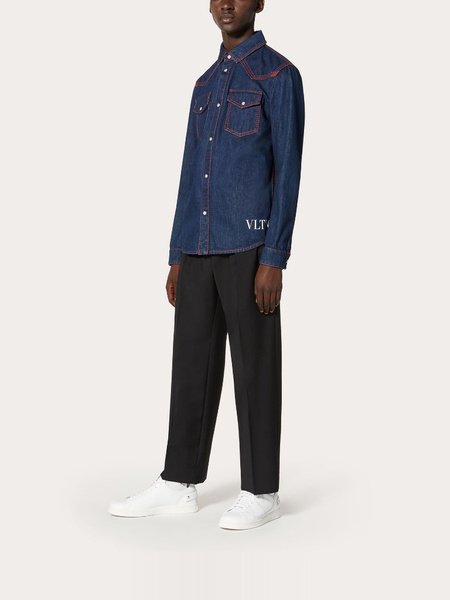}\hfill
  \includegraphics[width=0.185\textwidth]{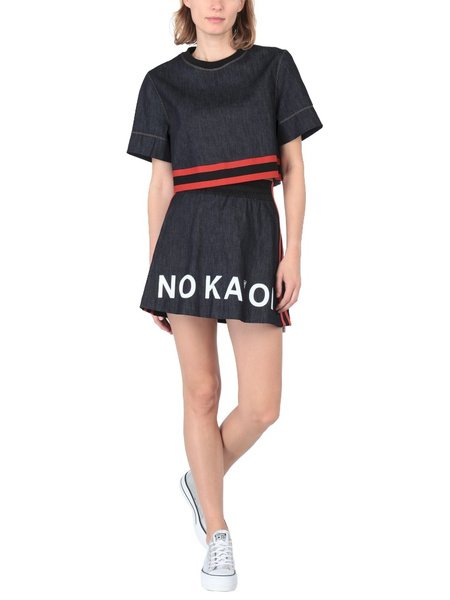}\hfill
  \includegraphics[width=0.185\textwidth]{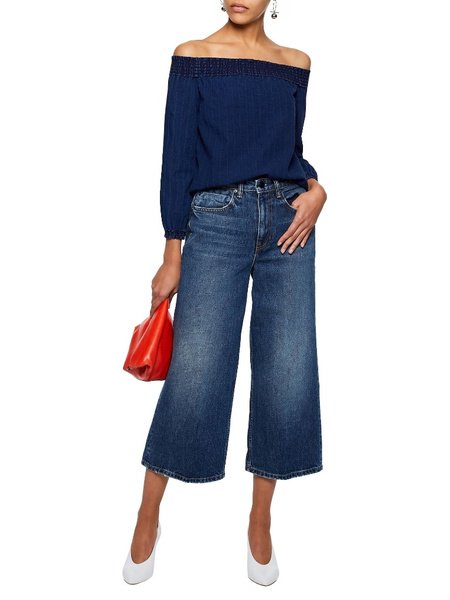}\hfill
  \includegraphics[width=0.185\textwidth]{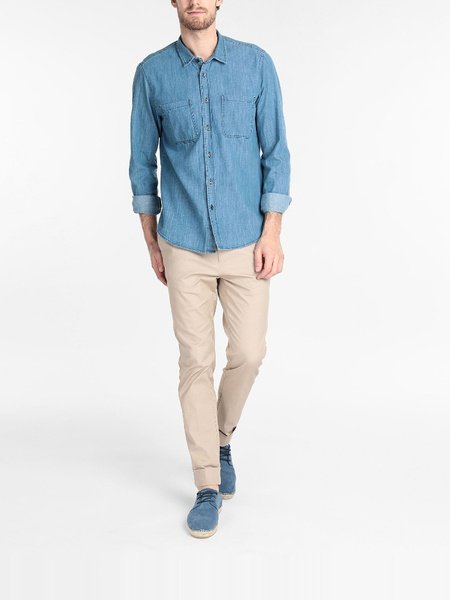}
  \caption{\textbf{DressCode upper-body samples.} Representative upper body person images in the DressCode dataset. Poses are mostly frontal, and backgrounds are uniform. Sampled garments mainly cover Western upper body types.
  }
  \label{fig:dataset_comp_dresscode_ub}
\end{figure*}

\begin{figure*}[!htbp]
  \centering
  \includegraphics[width=0.185\textwidth]{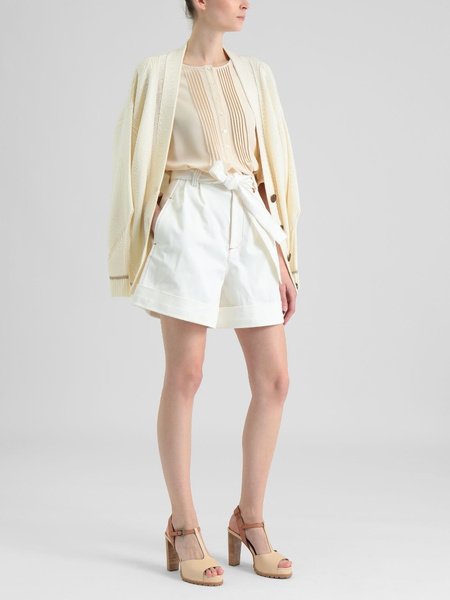}\hfill
  \includegraphics[width=0.185\textwidth]{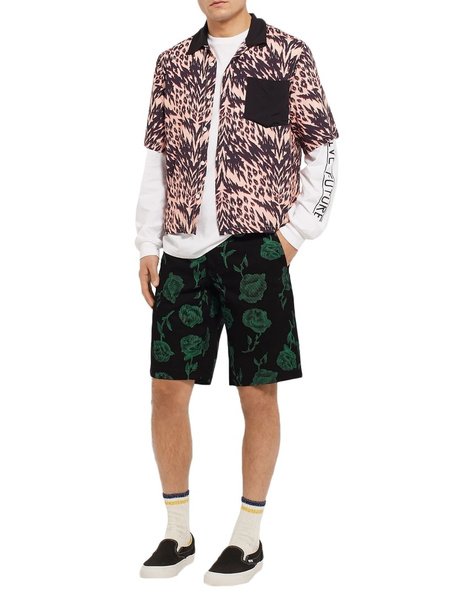}\hfill
  \includegraphics[width=0.185\textwidth]{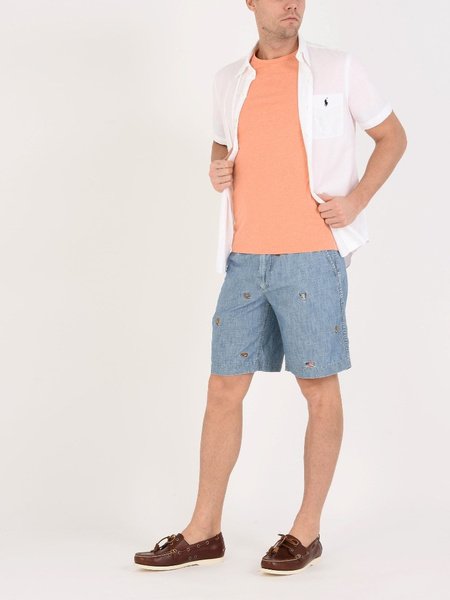}\hfill
  \includegraphics[width=0.185\textwidth]{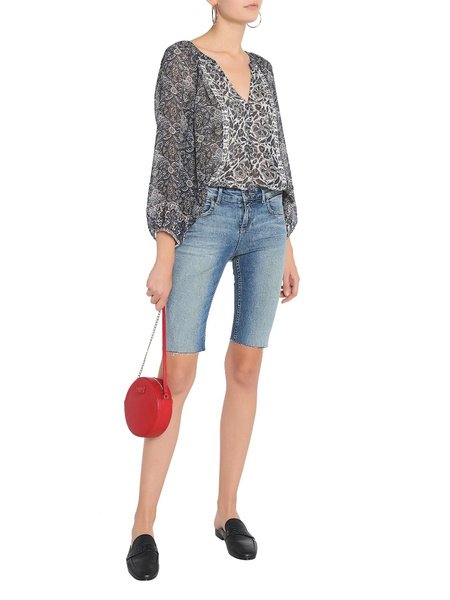}\hfill
  \includegraphics[width=0.185\textwidth]{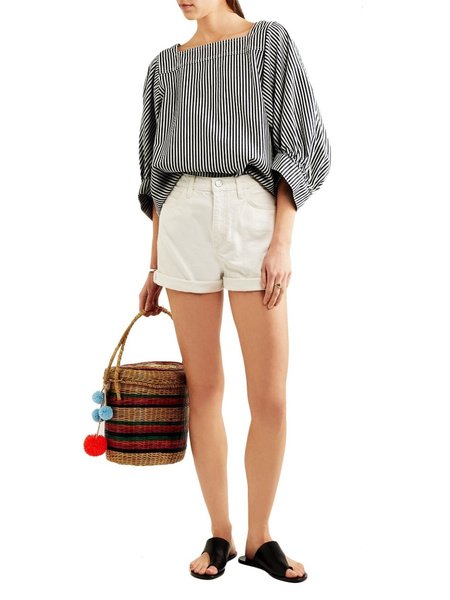}
  \caption{\textbf{ 
  Representative lower body person images in the DressCode dataset. Images include a variety of standard trouser and skirt styles but pose and background variety are limited.}}

  \label{fig:dataset_comp_dresscode_lb}
\end{figure*}

\begin{figure*}[!htbp]
  \centering
  \includegraphics[width=0.185\textwidth]{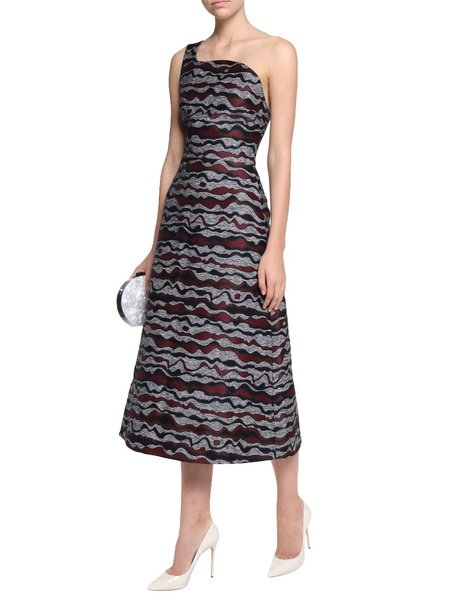}\hfill
  \includegraphics[width=0.185\textwidth]{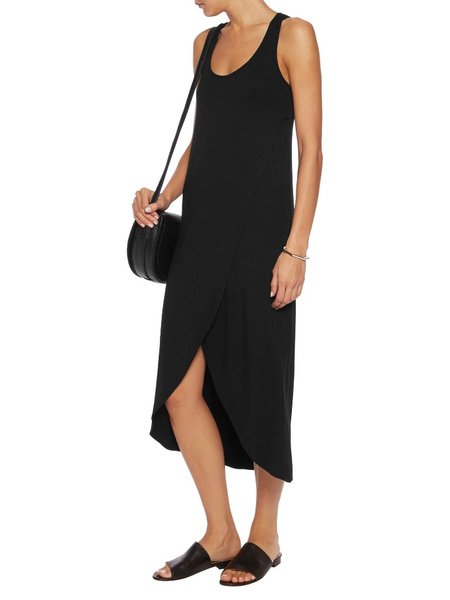}\hfill
  \includegraphics[width=0.185\textwidth]{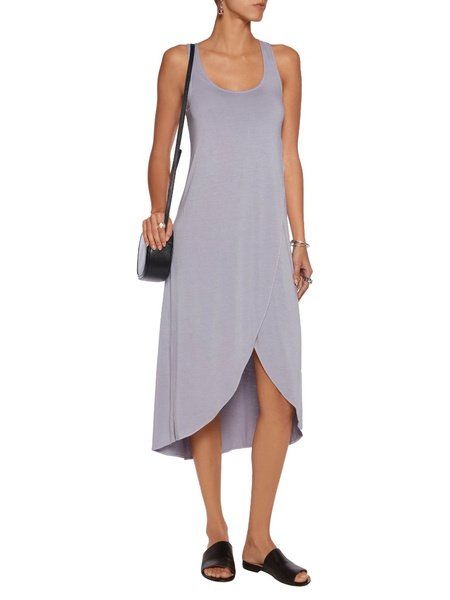}\hfill
  \includegraphics[width=0.185\textwidth]{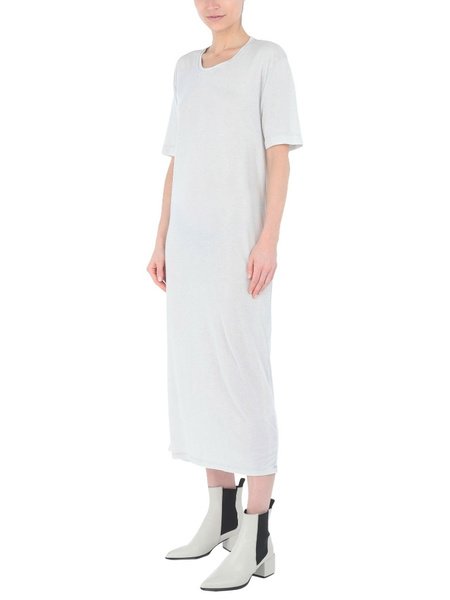}\hfill
  \includegraphics[width=0.185\textwidth]{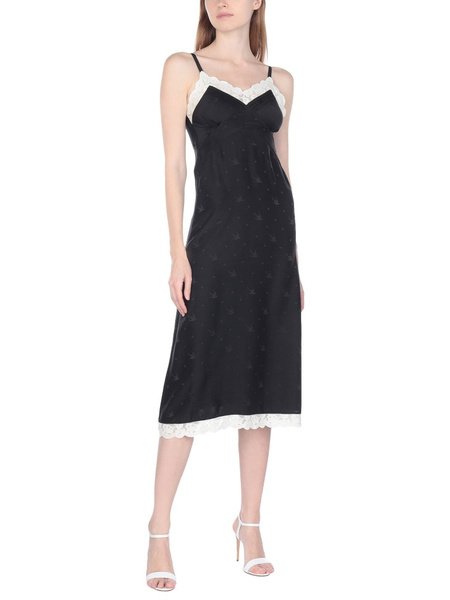}
  \caption{\textbf{DressCode dress samples.} Representative full body dress person images in the DressCode dataset. The coverage is limited to contemporary Western styles without pose variety.}
  \label{fig:dataset_comp_dresscode_dr}
\end{figure*}

\begin{figure*}[!htbp]
  \centering
  \includegraphics[width=0.185\textwidth]{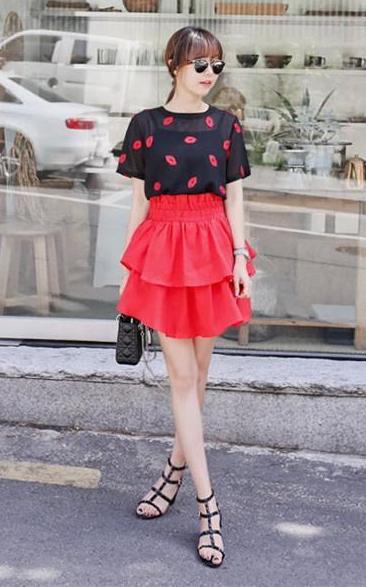}\hfill
  \includegraphics[width=0.185\textwidth]{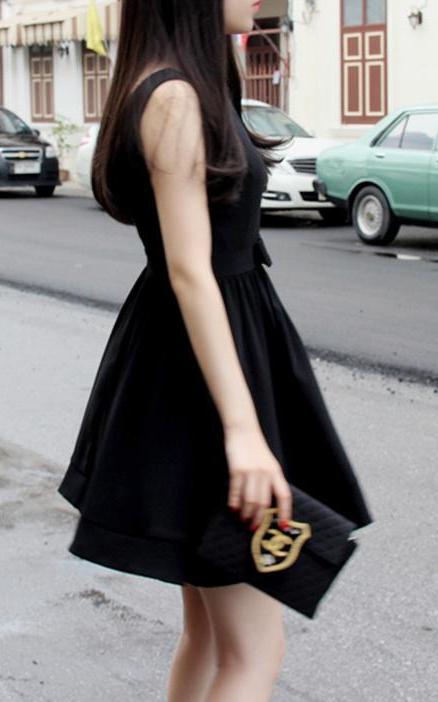}\hfill
  \includegraphics[width=0.185\textwidth]{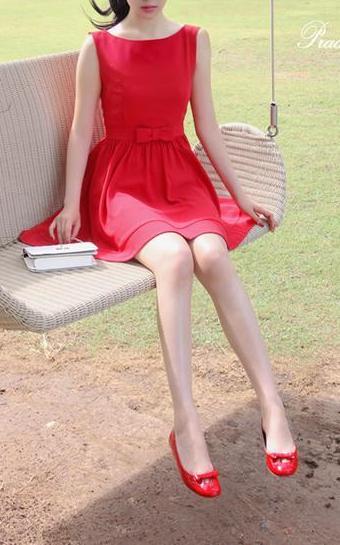}\hfill
  \includegraphics[width=0.185\textwidth]{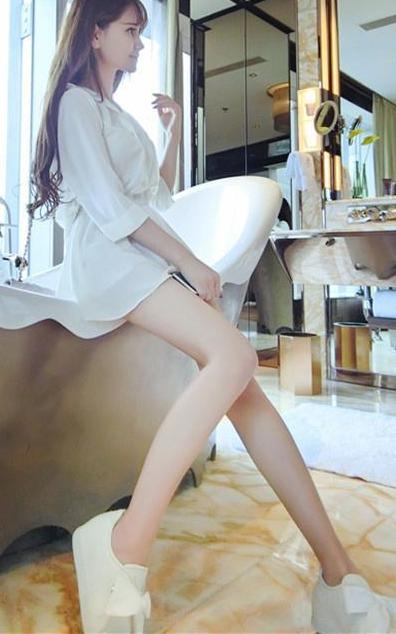}\hfill
  \includegraphics[width=0.185\textwidth]{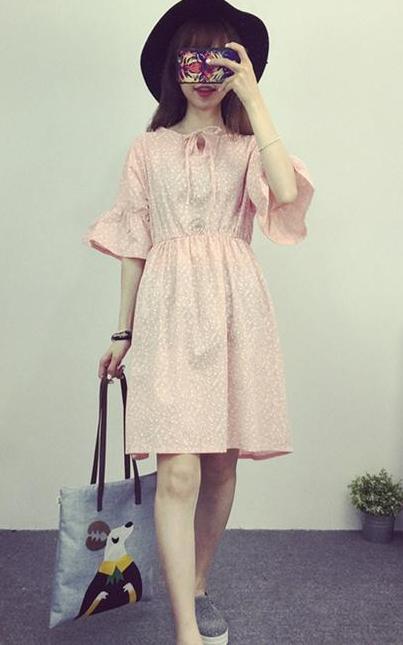}
  \caption{\textbf{StreetTryOn samples.} Representative person images in the StreetTryOn dataset. StreetTryOn offers more varied into-the-wild backgrounds and natural body poses compared to studio based data like VITON-HD. However, garment variety is still limited to Western upper body clothes.}
  \label{fig:dataset_comp_street}
\end{figure*}

\begin{figure*}[!htbp]
  \centering
  \includegraphics[width=\textwidth]{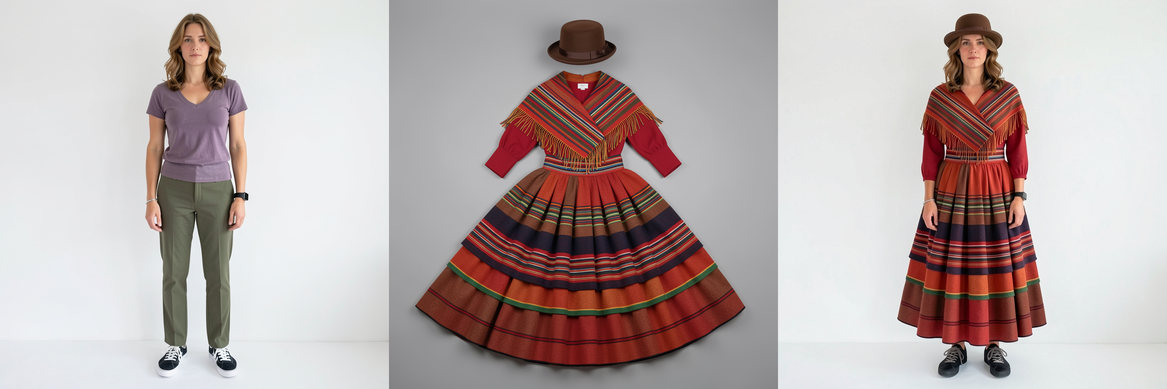}\\[6pt]
  \includegraphics[width=\textwidth]{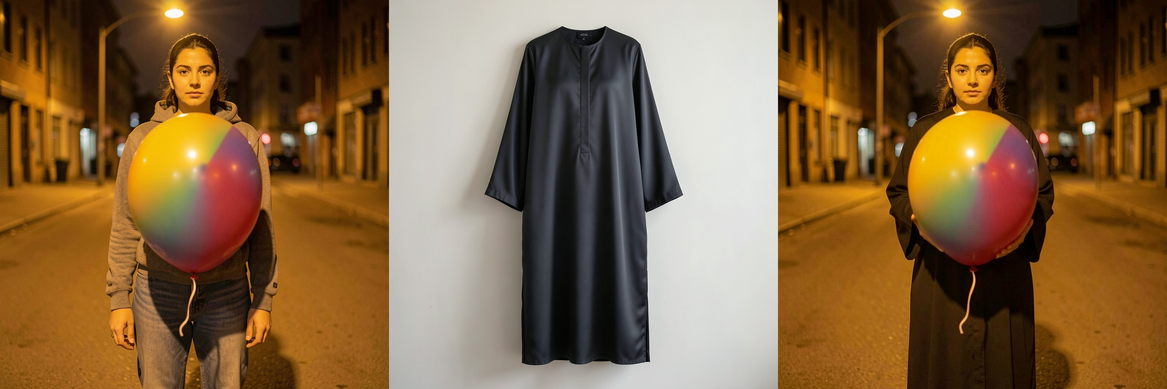}\\[6pt]
  \includegraphics[width=\textwidth]{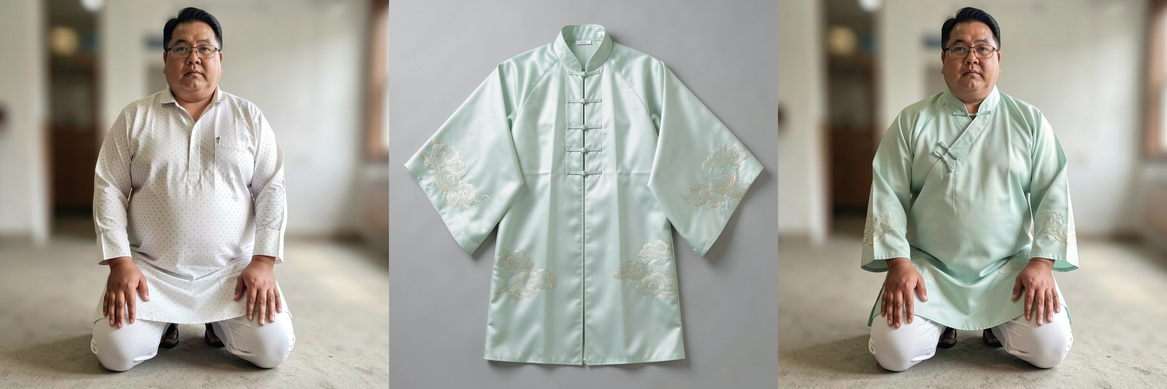}
  \caption{\textbf{CURVTON-205K samples.} Representative try-on triplets in CURVTON-205K. Images show sample source person (left), target garment (middle), and synthesized try-on (right) images. Rows 1--10: Cholita dress, Abaya, Hanfu, Panjabi, Ruqun, lace dress, Dashiki, Berber Amazigh dress, Hakama, blazer. Compared to VITON-HD, DressCode, and StreetTryOn, CURVTON-205K exhibits substantially greater diversity in garment type, cultural origin, pose, and photographic conditions.}
  \label{fig:dataset_comp_curvton}
\end{figure*}
\begin{figure*}[!htbp]
  \centering
  \includegraphics[width=\textwidth]{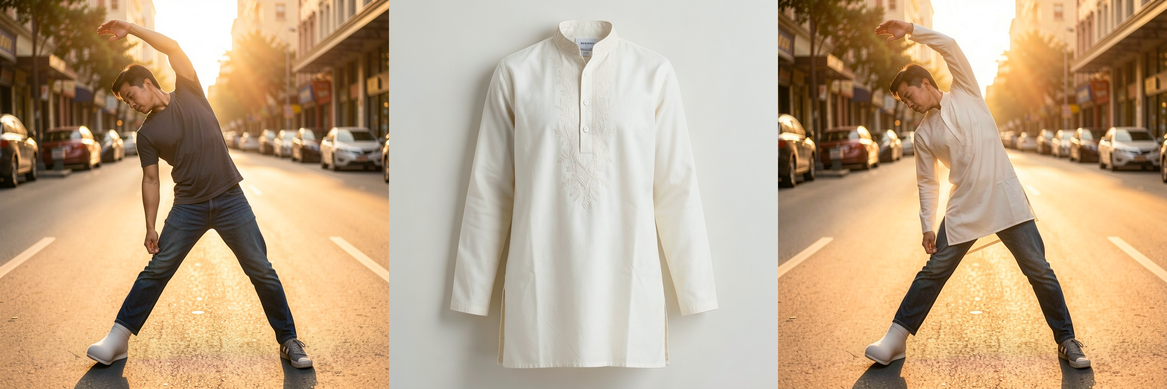}\\[6pt]
  \includegraphics[width=\textwidth]{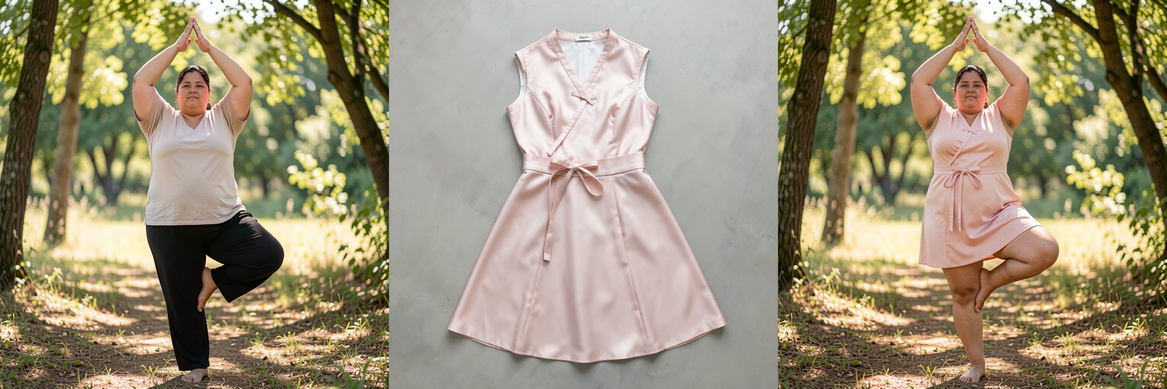}\\[6pt]
  \includegraphics[width=\textwidth]{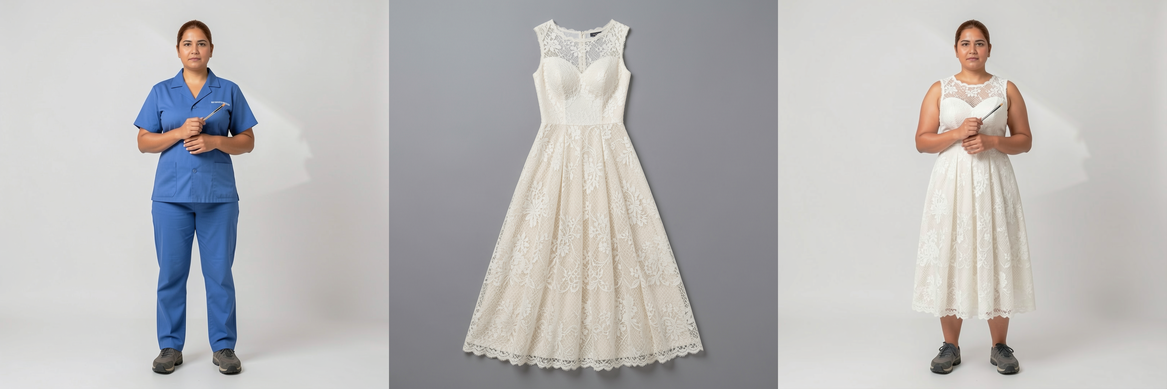}\\[6pt]
  \includegraphics[width=\textwidth]{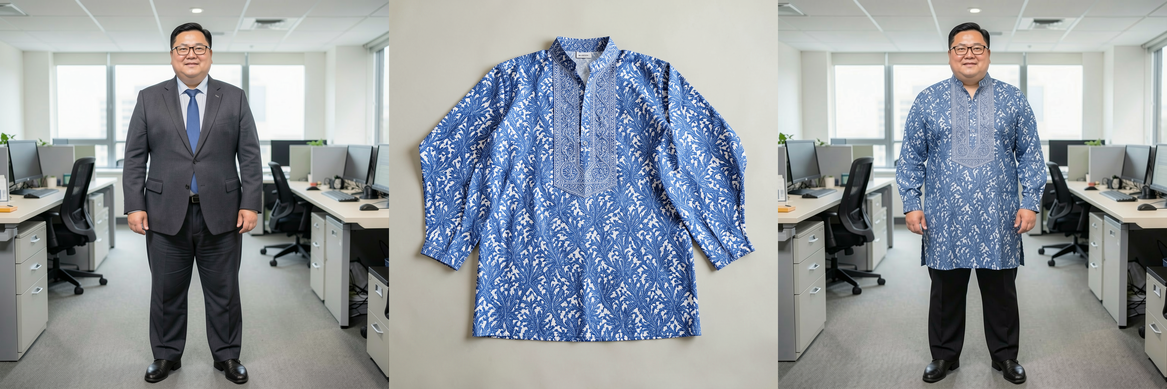}
  \contfig{fig:dataset_comp_curvton}
\end{figure*}
\begin{figure*}[!htbp]
  \centering
  \includegraphics[width=\textwidth]{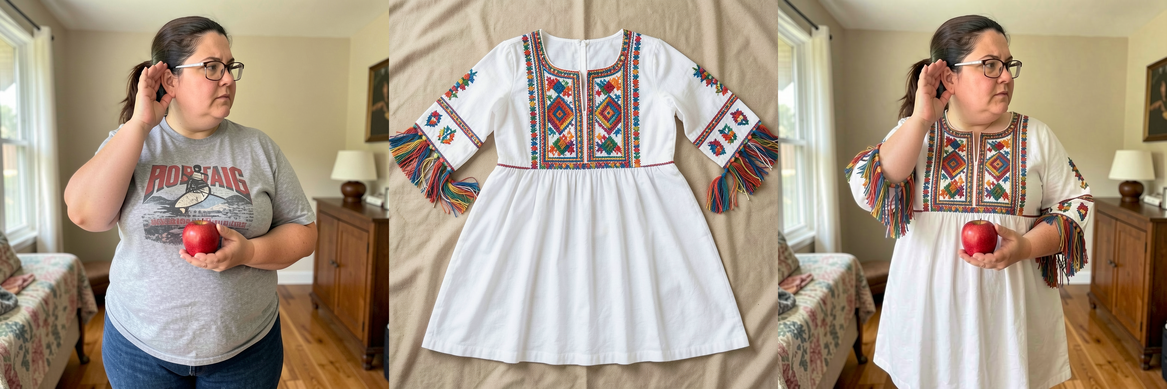}\\[6pt]
  \includegraphics[width=\textwidth]{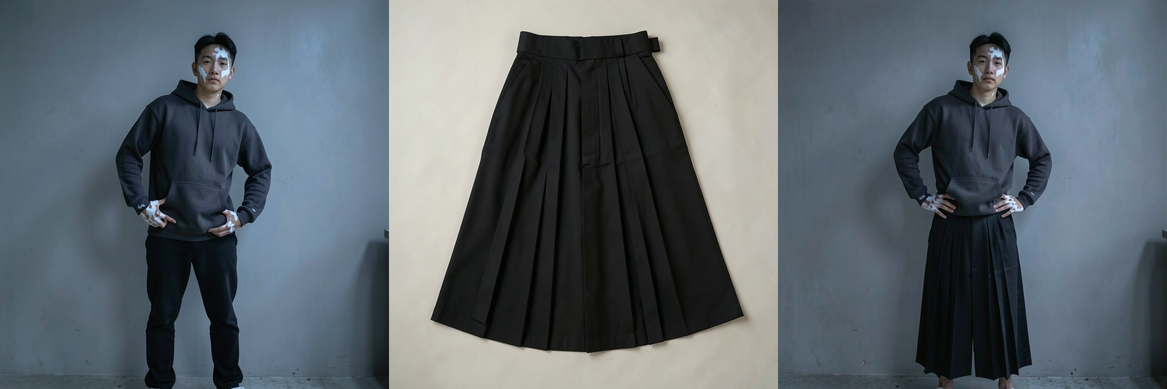}\\[6pt]
  \includegraphics[width=\textwidth]{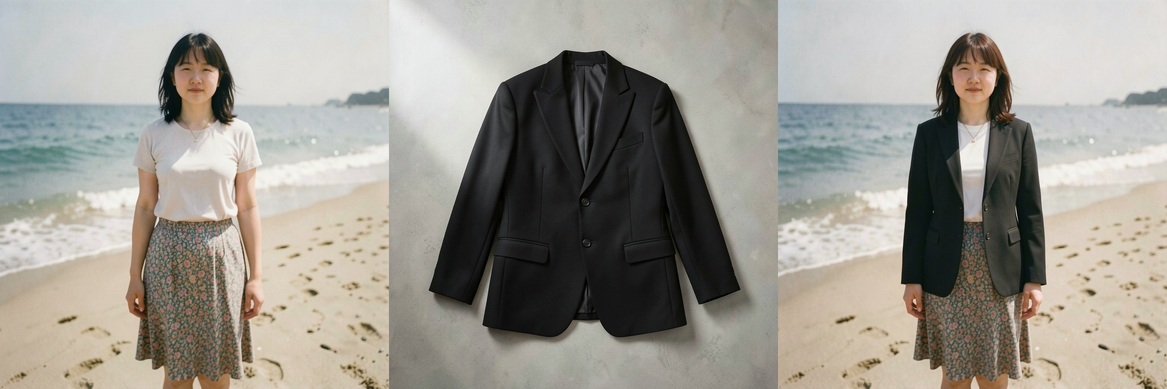}
  \contfig{fig:dataset_comp_curvton}
\end{figure*}

\clearpage
\onecolumn
\begin{multicols}{2}

\section{Diversity Metric Definition}
\label{app:metric_definition}


We provide the complete mathematical formulation of all task-free diversity metrics used to characterize dataset coverage across seven dimensions in Table~\ref{tab:diversity_comparison} of the main paper. For each metric, we describe the feature extraction procedure and the exact formula used for computation.

\subsubsection*{Pose \texorpdfstring{($\sigma^2_{\text{pose}}$)}{(sigma\textasciicircum 2\_pose)}}

The pose metric captures how varied and complex the skeletal articulations are across images in the dataset.
\begin{itemize}
    \item \textbf{Objective.} Measure the spatial variability of human skeleton structure across the population.

    \item \textbf{Feature extraction.} Images are processed using MMPose ViTPose~\cite{mmpose2020, Xu2022ViTPose}. Given input image $I_i$, the pose estimator $f_{\text{pose}}$ extracts 17 COCO keypoints $K_i = f_{\text{pose}}(I_i) \in \mathbb{R}^{17 \times 2}$. For each of $J = 8$ limb triplets $(a, b, c)$ (left/right elbow, knee, shoulder, and torso--hip), we compute the joint bend angle at the vertex joint $b$,
    \begin{equation}
        \theta_{i,j} = \arccos \frac{(K_{i,a} - K_{i,b}) \cdot (K_{i,c} - K_{i,b})}{\lVert K_{i,a} - K_{i,b} \rVert \, \lVert K_{i,c} - K_{i,b} \rVert},
    \end{equation}
    expressed in degrees.

    \item \textbf{Metric.} Pose diversity is the joint bend-angle variance across the population, averaged over the $J$ joints:
    \begin{equation}
        \sigma^2_{\text{pose}} = \frac{1}{J} \sum_{j=1}^{J} \operatorname{Var}_i\bigl(\theta_{i,j}\bigr),
    \end{equation}
    reported in $\text{deg}^2$; for example, $\sigma^2_{\text{pose}} = 182$ corresponds to a typical bend-angle spread of about $13.5^\circ$.
\end{itemize}

\subsubsection*{Occlusion (Occl.\ Cmplx., Person Vis.)}

The occlusion metrics quantify how much of the garment region is blocked by overlapping elements such as hair, arms, or accessories.
\begin{itemize}
    \item \textbf{Objective.} Evaluate the degree of obstruction of the target garment subject and its consistency across the dataset.

    \item \textbf{Feature extraction.} Mask2Former~\cite{Cheng2022Mask2Former} (\texttt{facebook/\allowbreak mask2former-\allowbreak swin-\allowbreak large-\allowbreak coco-\allowbreak panoptic}, 133 panoptic categories) produces per image a binary garment map $\mathbf{G}_i \in \{0,1\}^{H \times W}$ and a union-of-occluders map $\mathbf{O}_{cc,i} \in \{0,1\}^{H \times W}$ combining parsed arms, hair, bags, accessories, and environmental objects.

    \item \textbf{Metric.} The per-image occlusion ratio is
    \begin{equation}
        O_i = \frac{\sum_{(x,y)} \bigl(\mathbf{G}_i(x,y) \wedge \mathbf{O}_{cc,i}(x,y)\bigr)}{\sum_{(x,y)} \mathbf{G}_i(x,y)},
    \end{equation}
    and dataset-level occlusion complexity is $C_{\text{occ}} = \mathbb{E}[O_i] + \mathrm{Var}(O_i)$, where the expectation captures baseline severity and the variance captures dataset-wide variation.
\end{itemize}

\subsubsection*{Background (BGE, BSE, BOD)}

Three complementary scores assess how cluttered and semantically diverse the background regions are.
\begin{itemize}
    \item \textbf{Objective.} Quantify background noise, visual clutter, and semantic density surrounding the subject.

    \item \textbf{Feature extraction.} DeepLabV3-ResNet101~\cite{Chen2017DeepLab} produces a person mask $\mathbf{M}_i$; background pixels are identified as $\neg\mathbf{M}_i$. For BGE, the background region is converted to a 256-bin normalised grayscale histogram. For BOD and BSE, DETR~\cite{Carion2020DETR} (\texttt{facebook/\allowbreak detr-\allowbreak resnet-50}) is evaluated over the masked image to identify background bounding boxes and semantic class scores.

    \item \textbf{Metric.} Background Texture Entropy (BGE) is the mean pixel-level Shannon entropy $\mathbb{E}[H_{\text{bg},i}]$, where $H_{\text{bg},i} = -\sum_{j=0}^{255} p_j \log p_j$. Background Object Density (BOD) is the mean object count $\mathbb{E}[n_i]$ in $\neg\mathbf{M}_i$. Background Semantic Entropy (BSE) is the Shannon entropy over population-level DETR class-score proportions $\mathcal{P}_c$: $\text{BSE} = -\sum_{c} \mathcal{P}_c \log \mathcal{P}_c$.
\end{itemize}

\subsubsection*{Illumination (IGM, LVM)}

The illumination metrics reflect the diversity of lighting conditions and luminance exposures across the dataset.
\begin{itemize}
    \item \textbf{Objective.} Calculate localised lighting disparity, directional shadowing gradients, and macroscopic luminance exposure properties.

    \item \textbf{Feature extraction.} sRGB images are linearised via the standard piecewise gamma correction and converted to Rec.\,709 perceived luminance
    \begin{equation}
        Y_i(x,y) = 0.2126\,R_{\text{linear}} + 0.7152\,G_{\text{linear}} + 0.0722\,B_{\text{linear}}.
    \end{equation}
    Gaussian-blurred luminance maps ($\sigma=1.0$) are then differentiated with Sobel operators to obtain gradient magnitude maps $M_i(x,y) = \sqrt{g_x^2 + g_y^2}$.

    \item \textbf{Metric.} Illumination Gradient Mean (IGM) is the mean normalised gradient-magnitude variance per image:
    \begin{equation}
        \text{IGM} = \mathbb{E}_{i}\!\left[\frac{\mathrm{Var}_{x,y}(M_i)}{\mathbb{E}_{x,y}[M_i] + \epsilon}\right].
    \end{equation}
    Luminance Variance Measure (LVM) is the median absolute deviation of per-image mean luminance $\mu_i$: $\text{LVM} = \mathrm{Median}(|\mu_i - \mathrm{Median}(\mu)|)$.
\end{itemize}

\subsubsection*{Body Shape \texorpdfstring{($\sigma^2_{\text{shape}}$)}{(sigma\textasciicircum 2\_shape)}}

The body shape metric measures anatomical diversity across subjects, independent of pose or articulation.
\begin{itemize}
    \item \textbf{Objective.} Evaluate multi-dimensional body topology deviations, excluding skeletal joint variance.

    \item \textbf{Feature extraction.} HMR2.0 (4D-Humans)~\cite{Goel2023HMR2} fits a SMPL model~\cite{Loper2015SMPL} to each image, returning a 10-dimensional PCA body shape vector $\beta_i = f_{\text{hmr2}}(I_i) \in \mathbb{R}^{10}$ invariant to 3D rotation and joint articulation.

    \item \textbf{Metric.} Shape diversity is the log-determinant of the covariance matrix of shape vectors:
    \begin{equation}
        D_{\text{shape}} = \log \det \bigl(\Sigma_{\beta} + \epsilon I\bigr),
    \end{equation}
    where $\Sigma_{\beta} \in \mathbb{R}^{10 \times 10}$ is the population covariance of $\beta_i$.
\end{itemize}

\subsubsection*{Garment (Gmt Var., Gmt Div.)}

The garment metrics capture both the spread and volume of style variation across all clothing items in the dataset.
\begin{itemize}
    \item \textbf{Objective.} Assess broad stylistic differentiation and textural characteristics across clothing items, decoupled from body geometry.

    \item \textbf{Feature extraction.} Each garment crop is encoded by FashionCLIP~\cite{Chia2022FashionCLIP} (semantic and textual associations) and DINOv2~\cite{Oquab2023DINOv2} (structural patterns), producing a $D$-dimensional feature vector $e_i = f_{\text{encoder}}(I_{\text{cloth}}) \in \mathbb{R}^D$.

    \item \textbf{Metric.} Garment Variance (Gmt Var.) is the trace of the feature covariance matrix, $\mathrm{Var}_{\text{total}} = \mathrm{Tr}(\Sigma_e) = \sum_{j} \lambda_j(\Sigma_e)$. Garment Diversity (Gmt Div.) captures the full style volume via the log-determinant $D_{\text{garment}} = \log \det(\Sigma_e + \epsilon I)$.
\end{itemize}

\subsubsection*{Camera \texorpdfstring{($\hat{\sigma}_\phi,\,\hat{\sigma}_\theta$)}{(sigma\_phi, sigma\_theta)}}

The camera viewpoint metrics summarise how broadly the dataset spans different shooting angles in azimuth and elevation.
\begin{itemize}
    \item \textbf{Objective.} Determine global image perspective dynamics reflecting overall lens-view coverage properties.

    \item \textbf{Feature extraction.} HMR2.0~\cite{Goel2023HMR2} root-orientation estimates are projected onto azimuth $\phi_i$ (lateral angle) and elevation $\theta_i$ (vertical tilt) for each image.

    \item \textbf{Metric.} Distribution bandwidth is captured by the standard deviations over the population:
    \begin{align}
        \hat{\sigma}_\phi &= \sqrt{\frac{1}{N-1}\sum_{i=1}^{N}(\phi_i - \bar{\phi})^2}, \\
        \hat{\sigma}_\theta &= \sqrt{\frac{1}{N-1}\sum_{i=1}^{N}(\theta_i - \bar{\theta})^2}.
    \end{align}
\end{itemize}




\section{Artifact Evaluation Rubric}

\label{app:rubric}

Both human-facing quality judgements in our pipeline are based on a single 10-question rubric, reproduced below. Using the same instrument in both settings keeps the notion of an ``artifact'' consistent across the two points where human judgement enters the construction of CURVTON-205K:

\begin{itemize}
\item \textbf{Annotation for the quality-filter model.} The $5{,}000$ manually labelled triplets used to train the ViT-Small~\cite{Dosovitskiy2021ViT} quality filter $f_{\text{ViT}}$ (Section~\ref{subsec:quality_screening_pipeline} of the main paper) were labelled using this rubric. Annotators answered all ten questions for each triplet and assigned the binary \emph{correct}\,/\,\emph{incorrect} target based on the outcome: a triplet is labelled \emph{positive} only when no question records a severe defect. The rubric thus defines the supervision signal that $f_{\text{ViT}}$ learns to reproduce at scale.

\item \textbf{Human evaluation and test-set verification.} The same ten questions are used as the marking scheme for the human study reported in the main paper and for the manual verification pass over the test split, where triplets containing visible artifacts or semantic inconsistencies are removed.

\end{itemize}

Each question is scored $0$ (none), $1$ (minor), or $2$ (severe), allowing localised defects and overall output quality to be recorded on a common scale. Questions~1--9 contribute to the primary artifact rate; Question~10 records uncategorised issues separately, ensuring that failure modes outside the predefined categories are captured rather than silently discarded. The rubric covers garment placement, background preservation, structural garment details, text and logos, visual artifacts, unintended objects, anatomical consistency, fit and drape, overall physical plausibility, and an open-ended category.

\begin{enumerate}

\renewcommand{\labelenumi}{\textbf{Q\arabic{enumi}.}}

\item \textbf{Object placement.}

Is the garment correctly aligned with and positioned on the body? In particular, are elements such as sleeves, collars, and hemlines placed consistently with the person's pose? 

\item \textbf{Background consistency.}

Does the generated image preserve the background of the reference image without introducing unintended changes, blur, or geometric distortion?

\item \textbf{Structural details.}

Are garment components such as pockets, buttons, zippers, and similar structural elements reproduced correctly and in appropriate locations, without unintended duplication, deformation, or omission?

\item \textbf{Text/logo fidelity.}

When the garment contains text, logos, or graphic prints, are these elements reproduced accurately and legibly without noticeable distortion, warping, or incorrect placement?

\item \textbf{Artifacts and blending.}

Does the generated image contain visual defects that were not present in the reference, including blur, pixel noise, visible seams, texture errors, ghosting, double exposure, or unintended transparency?

\item \textbf{Unwanted object introduction.}

Has the model introduced objects or accessories that were not present in the original reference, such as handbags, jewelry, or other unrequested items?

\item \textbf{Anatomical correctness.}

Does the generated person preserve anatomically consistent limbs, hands, and digits, without extra or missing body parts, malformed fingers, or ghost-limb artifacts?

\item \textbf{Garment fit and drape realism.}

Is the garment worn in a manner that is consistent with its real-world type? Do its folds, fit, and drape respond naturally to gravity and the person's pose rather than appearing unnaturally stiff, flat, or pasted onto the body?

\item \textbf{Overall try-on plausibility.}

Considering garment identity, fit, drape, body interaction, and pose, does the final try-on image appear visually and physically plausible as a whole?

\item \textbf{Open-ended / other.}

Is there any additional visual issue not covered by the previous questions that makes the generated result appear incorrect, unnatural, or inconsistent with either the source person or the target garment? These cases are recorded separately as an uncategorised-artifact rate.

\end{enumerate}


\section{System Prompts Used in the Generation and Evaluation Pipeline}
\label{sec:system_prompts}

This section provides the system prompts used at the principal
LLM/VLM-assisted stages of the CURVTON-205K pipeline. Specifically,
we report the prompt used to convert structured difficulty-editing
commands into FLUX.2 Klein 9B editing instructions, the quality-control
prompt used to screen edited person images, and the VLM prompt used to
evaluate generated virtual try-on triplets and provide regeneration
feedback when necessary.


\subsection{Difficulty-Aware Person Editing Prompt}
\label{sec:person_edit_prompt}

The following system prompt converts a source person image and structured
JSON edit commands into a natural-language editing instruction for
FLUX.2 Klein 9B. The prompt explicitly preserves person identity and
visible physical attributes while resolving pose, occlusion, limb,
and scene-geometry constraints.

\begin{lstlisting}[
    style=systemprompt,
    caption={System prompt used for difficulty-aware person editing.},
    label={lst:person_edit_prompt}
]
SYSTEM PROMPT - Difficulty-Aware Person Editing

You are an expert image-editing prompt engineer. You receive a source
image of a person and a JSON object of edit commands. Translate the
commands into one clear natural-language editing prompt for FLUX.2
Klein 9B distilled.

The JSON edit commands are authoritative. Replace the source pose,
hand positions, held objects, camera view, occlusion, and requested
scene conditions according to the commands. If the source person holds
an object that is not requested, remove it.

Use the source image to preserve the person's identity and appearance,
including:
- body type,
- skin tone,
- face,
- hair,
- facial hair,
- clothing and clothing colors,
- footwear,
- accessories,
- visible disability attributes,
- prosthetic limbs or mobility aids.

Before writing the prompt:
- resolve physically incompatible commands,
- respect the number of available limbs and hands,
- avoid impossible anatomy or duplicated limbs,
- avoid body-object interpenetration,
- keep pose and object interactions physically plausible.

For requested occlusions:
- place the occluding object between the camera and the subject,
- keep the subject behind the occluder,
- state which body regions are covered,
- preserve realistic foreground-background depth.

Describe every visible hand and foot in a natural pose-specific position.
Do not invent held objects.

Preserve unchanged identity, clothing, footwear, and physical attributes
explicitly rather than saying "keep everything else the same."

Write one coherent natural-language editing prompt using direct and
literal instructions.

Return only the final FLUX.2 Klein editing prompt.
Do not return reasoning, JSON, alternatives, or explanations.
\end{lstlisting}


\subsection{Edited-Person Quality-Control Prompt}
\label{sec:edited_person_qc_prompt}

After person editing, the resulting image is screened using the following
quality-control prompt. The evaluator checks visible limb counts,
anatomical consistency, clothing continuity, body-shape consistency,
and generation artifacts. Images scoring below the acceptance threshold
are rejected from the subsequent pipeline.

\begin{lstlisting}[
    style=systemprompt,
    caption={System prompt used for edited-person quality-control evaluation.},
    label={lst:edited_person_qc_prompt}
]
SYSTEM PROMPT - Edited-Person QC Evaluation

You are an expert image quality analyst for AI-edited human portraits.
Your judgment is final.

You receive one AI-edited portrait image. Evaluate it in the following
order.

STEP 1 - DETECT HANDS AND FEET

Detect every visible hand and foot, including partial, blurred,
overlapping, ghosted, or distorted instances.

Return each detected instance with:
- label: "hand" or "foot"
- bounding box: [ymin, xmin, ymax, xmax]

Count:
- Total hands
- Total feet

Rules:
- 3 or more hands -> Score 0.0, REJECT immediately.
- 3 or more feet -> Score 0.0, REJECT immediately.
- Fewer than two visible limbs is acceptable.
- Never penalize a person for disability or missing limbs.
- Do not over-count or under-count visible limbs.

STEP 2 - CHECK IMAGE FAILURES

Evaluate for:

- EXTRA_LIMB:
  More than two hands or feet.

- ANATOMICAL_IMPOSSIBILITY:
  Impossible joints, distorted proportions, or unnatural body twists.

- BODY_DOUBLE:
  Duplicated heads, faces, or torsos.

- CLOTHING_DISCONTINUITY:
  Garments terminate unnaturally or appear unsupported by the body.

- BODY_SHAPE_INCONSISTENCY:
  Abrupt or physically implausible body-shape transitions.

- BACKGROUND_ARTIFACT:
  Background elements merge into the body or severe color bleeding occurs.

- SPECIAL_CASE_ARTIFACT:
  Distortions involving items such as hijabs, turbans, tattoos, or
  related special visual attributes.

- OTHER_ANOMALIES:
  Any other clearly implausible generation artifact.

STEP 3 - SCORE

Use the following discrete scale:

1.0 - Perfect image with no detected issues.

0.8 - Only minor texture, background, or lighting imperfections.
      Anatomy remains correct. ACCEPT.

0.6 - Small clothing or finger issue with correct limb count. REJECT.

0.4 - Incorrect proportions or abrupt garment/body transition. REJECT.

0.2 - Obvious distortion or fused body regions. REJECT.

0.0 - Extra limbs, duplicated body structures, or multiple severe
      failures. REJECT.

Acceptance threshold:
- Score >= 0.8 -> ACCEPT
- Score < 0.8 -> REJECT

Minor background imperfections alone must not reduce the score below 0.8.

OUTPUT

First output the detected hand/foot JSON list.

Then output:

Total hands: [number]
Total feet: [number]
Decision: ACCEPT or REJECT
Score: [number]
Primary failure: [
  EXTRA_LIMB |
  ANATOMICAL_IMPOSSIBILITY |
  BODY_DOUBLE |
  CLOTHING_DISCONTINUITY |
  BODY_SHAPE_INCONSISTENCY |
  BACKGROUND_ARTIFACT |
  SPECIAL_CASE_ARTIFACT |
  OTHER_ANOMALIES |
  NONE
]

Do not provide explanation, reasoning, or additional text.
\end{lstlisting}


\subsection{VLM-Based Try-On Evaluation and Feedback Prompt}
\label{sec:tryon_feedback_prompt}

Following try-on synthesis, the VLM compares the source person, target
garment, and generated try-on result. The evaluator checks person
preservation, garment fidelity, layering, fine-grained garment details,
pose and occlusion preservation, and anatomical validity. Failed but
correctable samples additionally receive an improved generation prompt
for iterative refinement.

\begin{lstlisting}[
    style=systemprompt,
    caption={System prompt used for VLM-based virtual try-on evaluation and feedback.},
    label={lst:tryon_feedback_prompt}
]
SYSTEM PROMPT - Virtual Try-On Triplet Quality Assessment

You are an expert evaluator of Virtual Try-On results.

INPUTS

- Image A: source person and identity reference.
- Image B: target garment reference.
- Image C: generated try-on result.
- Garment metadata: name, dress_type, and description.

Evaluate Image C by comparing it directly with Images A and B.

EVALUATION ORDER

1. Catastrophic validity
2. Person and pose preservation
3. Target-garment fidelity
4. Garment replacement and layering
5. Fine-grained garment-detail preservation
6. Occlusion and scene preservation
7. Anatomical and visual quality

CATASTROPHIC FAILURE

Immediately REJECT Image C if it contains severe failures such as:
- duplicated heads or torsos,
- excessive or impossible limbs,
- severe body fragmentation,
- gross anatomical corruption,
- or an otherwise unusable generation.

PERSON PRESERVATION

Use Image A as the reference for:
- facial appearance,
- skin tone,
- hair,
- body build,
- pose,
- visible limb configuration,
- footwear,
- accessories,
- non-target clothing,
- and foreground occlusions.

Do not penalize source clothing that should be replaced according to
dress_type.

TARGET-GARMENT FIDELITY

Use Image B as the reference for:
- garment category,
- silhouette and cut,
- color,
- pattern and print,
- material appearance,
- neckline,
- sleeve shape and length,
- hem,
- closure,
- and distinctive garment structure.

Also inspect visible fine-grained garment details such as:
- pockets,
- buttons,
- zippers,
- seams and panel boundaries,
- trims,
- embroidery,
- logos or graphic elements,
- small motifs,
- decorative elements,
- and other clearly visible design details.

Preserve these details when they are clearly observable in Image B.

Allow natural changes in folds, wrinkles, perspective, shading, and
fabric draping caused by the person's body shape and pose. Do not
require pixel-identical reconstruction.

A small local detail difference, such as a slightly altered seam,
minor pocket-shape variation, or one imperfect button, should be treated
as minor detail drift rather than a complete garment mismatch.

However, multiple missing details, changed distinctive patterns,
incorrect pocket layouts, substantially altered closures, missing
prominent buttons, logos, embroidery, or other identity-defining
features should be treated as a meaningful garment-fidelity error.

GARMENT SCOPE

Interpret dress_type when deciding which source garments should be
replaced or preserved.

- Full-body garment: replace the corresponding torso and lower-body
  clothing as one coherent garment.
- Lower-body garment: preserve the original upper-body clothing.
- Upper-body inner garment: replace the corresponding inner top.
- Flexible outer garment: preserve visible inner layers when appropriate.
- Accessories such as scarves, shawls, and headwear: add the target
  item while preserving unrelated clothing and footwear.

POSE AND OCCLUSION

Image A determines the person's pose and occlusion structure.
Do not infer pose from Image B.

Preserve important foreground occluders, their approximate coverage,
and their front-to-back relationship with the subject.

FAILURES TO CHECK

- TARGET_MISMATCH:
  Wrong garment category, major color, pattern, silhouette, or structure.

- DETAIL_DRIFT:
  Pockets, buttons, zippers, trims, logos, embroidery, seams, prints,
  or other visible design details differ from Image B.

- LAYERING_ERROR:
  Incorrect garment replacement or layering.

- SOURCE_REMAINS:
  Visible remnants of clothing that should have been replaced.

- IDENTITY_DRIFT:
  Meaningful change to the person's face, hair, skin tone, or body
  appearance.

- POSE_DRIFT:
  Significant alteration of the source pose.

- OCCLUSION_ERROR:
  Important occluders are missing or have an incorrect depth relationship.

- GARMENT_DISTORTION:
  Implausible cut, topology, proportions, or garment structure.

- MATERIAL_LEAKAGE:
  Target fabric, pattern, or texture appears on unrelated body or
  scene regions.

- ANATOMY_ERROR:
  Incorrect limbs, joints, or body geometry.

- STRUCTURAL_HALLUCINATION:
  Duplicated or invented garment components.

SCORING

Use only these scores:

1.0 - Excellent transfer. Person identity and target garment, including
      clearly visible distinctive details, are faithfully preserved.

0.8 - High-quality result. Minor local texture or detail drift is allowed,
      such as a small seam, pocket, button, or shading discrepancy that
      does not change garment identity.

0.6 - Acceptable result. One noticeable but limited garment-detail,
      layering, or visual error is present, while the target garment,
      person identity, and overall structure remain correct.

0.4 - Significant error. Important garment details, structure, layering,
      pose, or identity are noticeably incorrect, although the intended
      try-on remains recognizable.

0.2 - Major try-on failure with severe garment mismatch or structural
      corruption.

0.0 - Catastrophic or unusable generation.

Scores >= 0.6 are SUCCESS.
Scores < 0.6 are NOT_SUCCESS.

OUTPUT

For catastrophic failure:

{
  "result": "REJECT",
  "score": 0.0,
  "failure_code": "ANATOMY_ERROR"
}

For successful results:

{
  "result": "SUCCESS",
  "score": 0.8,
  "failure_code": "NONE"
}

For correctable failures:

{
  "result": "NOT_SUCCESS",
  "score": 0.4,
  "failure_code": "DETAIL_DRIFT",
  "reason": "One concise description of the main failure.",
  "improved_prompt": "One concise natural-language instruction describing
  how to correct the failed try-on while preserving the source person,
  pose, occlusions, garment structure, and visible target-garment details."
}

When DETAIL_DRIFT is the main failure, the improved prompt should
explicitly mention the missing or altered visible details from Image B,
such as pockets, buttons, closures, embroidery, logos, trims, or patterns,
without changing unrelated parts of the garment.

Return valid JSON only.
Do not output markdown, analysis, or additional explanation.
\end{lstlisting}


\section{Generation Dictionaries}
\label{sec:generation_dictionaries}

This section describes the structured dictionaries used by the
CURVTON-205K data-generation pipeline. We summarize the source-person
attribute and garment metadata dictionaries, followed by the
difficulty-tier dictionaries used to introduce controlled variations
in pose, occlusion, scene conditions, and other visual factors.

The source-person dictionary defines the attribute space used for
person generation, including body shape, pose, geographic and cultural
appearance cues, gender-conditioned clothing, accessories, visible
assistive devices, background context, and photographic style. The
garment dictionary defines the garment categories and structural
metadata used during garment generation and virtual try-on synthesis.

The difficulty-editing space is organized into three levels:
Easy, Medium, and Hard. The Medium tier uses a common set of pose,
occlusion, background, camera-view, and age attributes for all source
persons. Only the special-case distribution contains a
gender-conditioned branch, primarily to support pregnancy-related
examples while maintaining normalized sampling probabilities.

The complete source-person attribute and garment metadata dictionaries,
as well as the Easy, Medium, and Hard difficulty-tier editing
dictionaries, are available in the project's GitHub repository:
\codeurl.

\end{multicols}

\end{document}